\documentclass[a4paper,fleqn]{cas-sc}

\usepackage[numbers,longnamesfirst]{natbib}

\def\tsc#1{\csdef{#1}{\textsc{\lowercase{#1}}\xspace}}
\tsc{WGM}
\tsc{QE}

\begin{document}
\let\WriteBookmarks\relax
\def\floatpagepagefraction{1}
\def\textpagefraction{.001}

\shorttitle{Knowledge-Informed Spatio-Temporal Efficient Multi-Branch Graph Neural Network}    

\shortauthors{Zesheng Liu, Maryam Rahnemoonfar}  

\title[mode = title]{K-STEMIT: Knowledge-Informed Spatio-Temporal Efficient Multi-Branch Graph Neural Network for Subsurface Stratigraphy Thickness Estimation from Radar Data}  



%

\author[1]{Zesheng Liu}[orcid=0009-0004-6978-7160]



\ead{zel220@lehigh.edu}


\credit{Formal analysis, Investigation, Methodology, Software, Validation, Visualization, Writing-original draft, Writing-review\&editing}

\affiliation[1]{organization={Department of Computer Science and Engineering, Lehigh University},
            addressline={113 Research Drive}, 
            city={Bethlehem},
            postcode={18015}, 
            state={Pennsylvania},
            country={USA}}



\author[1, 2]{Maryam Rahnemoonfar}[orcid=0000-0001-9358-2836]
\cormark[1]


\ead{maryam@lehigh.edu}


\credit{Conceptualization, Funding acquisition, Project administration, Resources, Supervision, Writing-review\&editing}

\affiliation[2]{organization={Department of Civil and Environmental Engineering, Lehigh University},
            addressline={19 Memorial Drive W.}, 
            city={Bethlehem},
            postcode={18015}, 
            state={Pennsylvania},
            country={USA}}

\cortext[1]{Corresponding author}



\begin{abstract}
Spatio-temporal patterns in subsurface stratigraphy encode key information about accumulation, deformation, and layer formation processes. For polar ice sheets and corresponding subsurface ice layer stratigraphy, variations in layer thickness provide quantitative constraints that support snow mass balance estimation, improved projections of ice sheet change, and reduced uncertainty in climate and engineering models. Radar sensors capture these layered structures as depth-resolved radargrams; however, convolutional neural networks applied directly to radargrams are often sensitive to speckle noise and acquisition artifacts. More broadly, purely data-driven approaches tend to underuse known physical knowledge, which can produce inconsistent or unrealistic thickness estimates when extrapolating across space and time. To address these challenges, we develop K-STEMIT, a novel knowledge-informed, efficient, multi-branch spatio-temporal graph neural network that combines a geometric framework for spatial learning with temporal convolution to capture temporal dynamics, and incorporates physical data synchronized from the Model Atmospheric Regional physical weather model. An adaptive feature fusion strategy is employed to dynamically combine features learned from different branches. Extensive experiments on a standardized snow-radar benchmark are conducted to compare K-STEMIT against current state-of-the-art methods in both knowledge-informed and non-knowledge-informed settings, as well as other existing methods. Results show that K-STEMIT consistently achieves the highest accuracy while maintaining near-optimal efficiency. Most notably, incorporating adaptive feature fusion and physical priors reduces the root mean-squared error by 21.99\% with negligible additional cost compared to its conventional multi-branch variants. Additionally, our proposed K-STEMIT achieves consistently lower per-year relative MAE, supporting reliable, continuous spatiotemporal assessment of snow accumulation variability across large spatial regions.
\end{abstract}




\maketitle

\section{Introduction}
As the global temperature rises, researchers have revealed that the rate of mass loss of polar ice has been accelerated~\cite{Forsberg2017}. If the current trend continues, it is estimated that the Arctic Ocean will be ice-free by the 2030s\cite{DIEBOLD2023105479}, and the global sea level will rise about 1 meter by 2100\cite{Shepherd2020}. Polar ice sheets comprise several internal ice layers formed over different years. A better understanding of the internal ice layers, particularly their thickness and variability, provides valuable insights into snowfall accumulation and melting. This knowledge is vital for monitoring changes in snow mass balance, interpreting hard-to-observe processes, and minimizing uncertainties in climate model predictions.

The traditional method to capture internal ice layers and measure their properties involves extracting onsite ice cores\cite{PATERSON1994378}. Researchers drill holes in specific locations of the polar regions, and large cylindrical ice cores are manually extracted. However, this method has a few limitations. Considering the extreme climate conditions in the polar region and the fact that ice cores need to be manually drilled from ice sheets, the coverage of ice cores is limited to certain parts of the polar region. Even in the areas where ice cores are available, the measurements are discrete and sparse, making it impossible to study the continuous changes of the internal layers. Moreover, onsite ice cores are expensive and time-consuming to obtain, and the drilling process causes destructive damages. 

\begin{figure}
    \centerline
    {
        \includegraphics[width=0.5\textwidth]{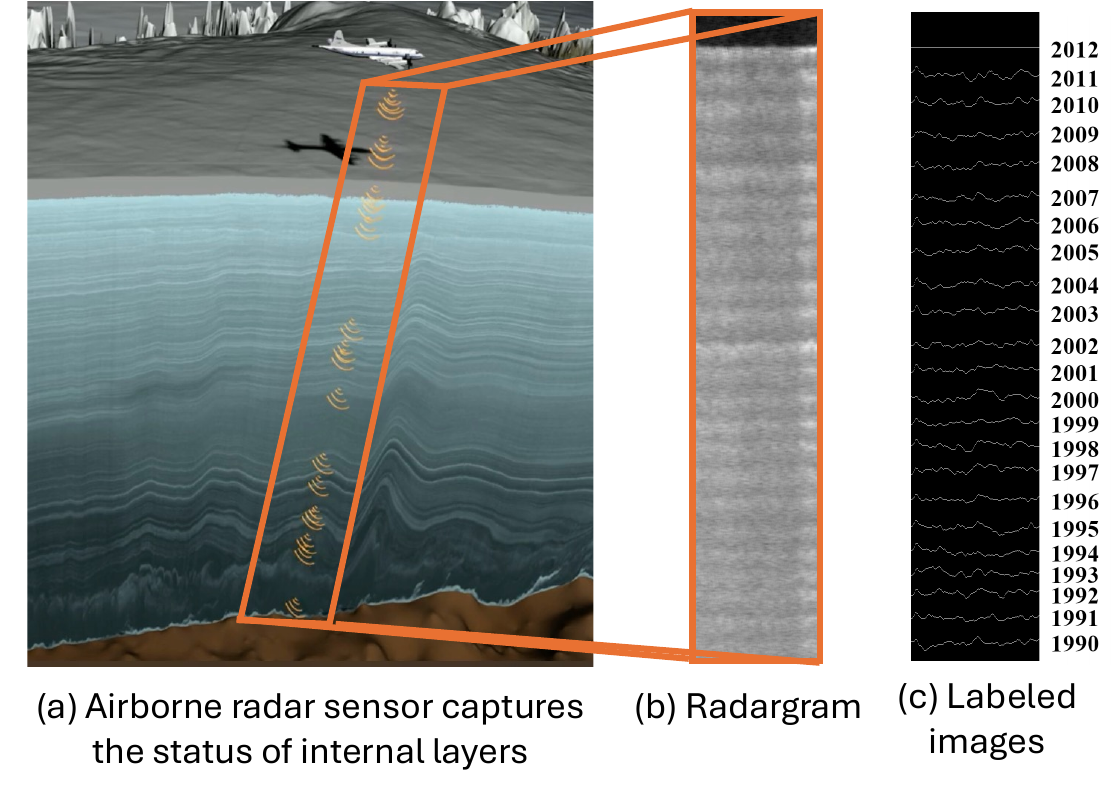}
    }
    \caption{Diagram of capturing radargrams and generating corresponding labels. (\textbf{a}) How airborne radar sensor is used to capture the status of internal ice layers (Image adapted from \cite{diagram-airborne-radar}) (\textbf{b}) Example of a radargram captured by the airborne radar sensor. (\textbf{c}) Corresponding labeled images, where each ice layer is manually annotated. Each layer is formed in a certain year.}
    \label{fig:dataset}
\end{figure}

In recent years, airborne radar sensors\cite{AirborneRadar} have proved to be a more effective tool. Airborne radar scientific campaigns in polar regions are conducted using aircraft equipped with specialized radar systems designed to penetrate ice and snow layers to collect high-resolution data on ice sheet structures and dynamics. These campaigns employ advanced radar systems mounted on aircraft to investigate subsurface features such as internal ice stratigraphy, basal topography, and snow accumulation. One example of airborne radar for polar region measurements is the Snow Radar\cite{snowradar} operated by the Center for Remote Sensing of Ice Sheets (CReSIS), as part of NASA's Operation IceBridge Mission. The Snow Radar can capture the location of snow or firn layers at different depths by measuring the backscatter power of the electromagnetic wave reflections\cite{Arnold_Leuschen_Rodriguez-Morales_Li_Paden_Hale_Keshmiri_2020}, resulting in radargrams that reveal the stratigraphy of shallow and near surface accumulation such as the one shown in Figure~\ref{fig:dataset} (\textbf{a}). Figure~\ref{fig:dataset} (\textbf{b}) shows the radargram captured by the airborne radar sensor. This helps to visualize subsurface features, particularly the accumulation pattern beneath the ice sheet's surface along the flight path. The horizontal axis of the radargram represents the direction of the aircraft flight, often referred to as the along-track axis, while the vertical axis is the fast-time axis, revealing the occurring depth of the snow layers. The value of each pixel is determined by its reflected signal strength, where brighter pixels mean a higher reflection power\cite{Arnold_Leuschen_Rodriguez-Morales_Li_Paden_Hale_Keshmiri_2020}. These radargrams are then annotated to create corresponding training labels (Figure~\ref{fig:dataset}(\textbf{c})), where annotated lines represent the ice layer boundaries formed by different years of snow accumulation\cite{Yari_2021_JSTAR}.

With the development of deep learning techniques, various convolutional-based neural networks \cite{ibikunle_2020,ibikunle_snow_radar_echogram_2023,Rahnemoonfar_Yari_Paden_Koenig_Ibikunle_2021,DeepIceLayerTracking,varshney_2021_regression_networks,DeepLearningOnAirborneRadar,varshney2023skipwavenet,Yari_2021_JSTAR,Yari_2020} have been proposed to detect ice layer boundaries directly from the radargram. However, noise in the radargrams has been proven to be a major obstacle for achieving high-quality results. In recent years, graph-based learning has emerged as an effective paradigm for modeling scientific data with irregular spatial relationships. Graph neural networks and graph transformers have shown strong performance across a wide range of non-Euclidean learning problems in science and engineering, such as modeling traffic patterns~\cite{SSGCRTN, MSTDFGRN, PSTCGCN, TIIDGCN, SDSINet, GDGCRN, MTEGCRN, DMFGCRN, SGASeq}, trajectory recovery~\cite{MReDTrajRec}, molecular property prediction~\cite{Neuro_Molecular} and drug recommendation~\cite{Neuro_Drug}. This broader development motivates representing internal ice layers as graphs rather than treating them purely as image pixels. Zalatan et al.\cite{Zalatan_igarss,Zalatan2023,zalatan_icip} proposed to represent internal ice layers as individual fully-connected graphs and learn the spatio-temporal relationships between shallow and deep ice layers through graph neural networks(GNNs). Liu et al.\cite{liu2025locate} proposed a two-stage ``Locate and Extend'' strategy that significantly reduces the training time while maintaining a competitive prediction accuracy. Liu et al.\cite{GRIT,ST-GRIT} applied two types of graph transformer networks and further improved the accuracy compared with Zalatan et al.\cite{Zalatan2023,Zalatan_igarss,zalatan_icip}. In some recent work, researchers have briefly explored applying neural operators to the same deep ice layer thickness prediction task~\cite{BIRFNO,GNO}. 

Although existing methods have achieved notable success, an important limitation remains in their predominantly data-driven nature. Because these models rely heavily on multi-fidelity observational data, their performance can be sensitive to the availability and quality of training data, and their predictions may generalize poorly under unseen conditions. This can lead to unrealistic or physically inconsistent results, reducing their reliability for scientific analysis. To address this limitation, we adopt a knowledge-informed learning framework that incorporates domain knowledge into the network through physically meaningful input features.

Domain knowledge can be introduced into deep learning models in different forms. As outlined by Karniadakis et al.~\cite{Karniadakis2021}, these include observational biases, which incorporate domain-specific variables into the network input; inductive biases, which are embedded in the model architecture; and learning biases, which influence the training objective. In this work, we incorporate domain knowledge primarily as an observational bias by introducing physical variables simulated from climate models as part of the node features in the graph neural network. These physically grounded features help guide the model in learning the spatio-temporal relationships between more reliably tracked shallow layers and more uncertain deeper internal layers, without explicitly embedding governing equations into the architecture or loss function.

Prior studies have demonstrated the value of incorporating physical information into graph-based models. For example, Liu et al.~\cite{liu2024learningspatiotemporalpatternspolar} and Rahnemoonfar et al.~\cite{PI_GCNLSTM} showed that augmenting node features with physically meaningful variables improves prediction accuracy over earlier fused spatio-temporal GNNs developed by Zalatan et al. Building on these efforts, we propose K-STEMIT, a \textbf{K}nowledge-informed \textbf{S}patio-\textbf{T}emporal \textbf{E}fficient \textbf{M}ulti-branch graph neural network for \textbf{I}ce layer \textbf{T}hickness, designed to further improve both the predictive accuracy and computational efficiency of existing knowledge-informed methods. Specifically, K-STEMIT decouples spatial and temporal representation learning into dedicated branches, allowing the model to capture complementary aspects of the prediction task more effectively. These branch-specific representations are then integrated through an adaptive feature fusion mechanism. Furthermore, the model incorporates node features derived from physical variables synchronized from the Modèle Atmosphérique Régional (MAR) climate model, allowing prediction to be guided by both observed stratigraphic structure and physically meaningful environmental context.

Compared with prior radargram-based CNN methods, our approach does not operate directly on noisy image intensities, but instead models the spatio-temporal relationships among layer geometry and thickness in graph form. More broadly, as in other application domains where GNNs are used to model non-Euclidean relationships, our formulation exploits the relational structure among ice-layer observations instead of treating them as regular image grids. Compared with earlier graph-based and graph-transformer methods for ice-layer prediction, K-STEMIT further incorporates physical node features derived from the Modèle Atmosphérique Régional (MAR) climate model, explicitly decouples spatial and temporal learning through dedicated branches, and uses adaptive feature fusion to integrate complementary representations. In addition, compared with existing knowledge-informed methods for ice-layer prediction, K-STEMIT is designed to improve both predictive accuracy and computational efficiency. Together, these design choices are intended to enhance generalization to unseen data while maintaining practical scalability.

Our major contributions are:
\begin{itemize}
    \item We developed K-STEMIT, a knowledge-informed efficient multi-branch spatio-temporal graph neural network designed to learn from the upper $m$ ice layers to predict the thickness of the underlying $n$ layers.
    
    \item K-STEMIT adopts a multi-branch strategy that explicitly decouples spatial feature extraction and temporal dynamics into dedicated branches, enabling specialized learning and improving both accuracy and efficiency.
    
    \item We apply an adaptive feature fusion strategy that dynamically combines the learned features from different branches, enabling more effective information integration and better generalization to unseen data.

    \item We incorporate physical node features synchronized from the Modèle Atmosphérique Régional (MAR) weather model as prior domain knowledge, which significantly boosts the predictive performance of our network.

    \item We conduct comprehensive experiments to benchmark K-STEMIT and its non-knowledge-informed counterpart, STEMIT, against current state-of-the-art methods and other existing approaches for deep ice layer thickness prediction.
\end{itemize}

This paper is organized as follows: Section~\ref{relatedwork} discusses related work. Section~\ref{dataset} introduces the Snow Radar Echogram Dataset (SRED), describes how physical data is obtained from the Modèle Atmosphérique Régional weather model and synchronized with radargram measurements, and explains how graph data is generated based on the SRED dataset. Section~\ref{method} presents the key designs of our proposed multi-branch network architecture. Section~\ref{experiments} provides the experimental details. Section~\ref{results} and Section~\ref{discussion} evaluate the proposed method from different perspectives, and Section~\ref{conclusion} concludes the paper.

\section{Related Work}\label{relatedwork}

\subsection{Internal Ice Layer Tracking}
Tracking the internal ice layers from radargrams is a complex task. Layers formed long ago may be broken, incomplete, melted, or have fewer contrast variations\cite{LearnSnowLayerThickness}, heavily increases the difficulty of extracting ice layers from radar results. 

Several works utilized traditional non-learning algorithms to process radargram images and detect ice layers\cite{7731235,article,https://doi.org/10.1002/2014JF003215,Koenig_2016}. However, these automatic algorithms cannot be scaled and applied to larger datasets. In recent years, deep learning techniques, especially convolutional neural networks and generative adversarial networks, have also been developed to precisely extract ice layer boundary positions from radargrams\cite{ibikunle_2020,ibikunle_snow_radar_echogram_2023,Rahnemoonfar_Yari_Paden_Koenig_Ibikunle_2021,DeepIceLayerTracking,varshney_2021_regression_networks,DeepLearningOnAirborneRadar,varshney2023skipwavenet,Yari_2021_JSTAR,Yari_2020}. Varshney et al. use different kinds of fully conventional networks\cite{DeepIceLayerTracking,DeepLearningOnAirborneRadar} or multi-output convolution-based regression network\cite{varshney_2021_regression_networks} to estimate internal layer depth directly from radargram images. Rahnemoonfar et al. applied a multi-scale convolutional neural network to segment ice layers from radargram images\cite{Rahnemoonfar_Yari_Paden_Koenig_Ibikunle_2021}. The results of these works show that the major problems are the noise in the radargram images and the limited amount of high-quality datasets and annotations. 

Unlike previous convolutional-based networks, our proposed K-STEMIT represents internal ice layers as individual graphs and uses a knowledge-informed graph neural network to learn the shallow layer thickness patterns, where the graph neural network is shown to be less sensitive to noise and has a more stable performance. 

\subsection{Graph Neural Network for Ice Thickness Prediction}
Deep learning has demonstrated remarkable success with Euclidean data, like the famous convolutional neural network (CNN) and recurrent neural network (RNN) for images and sequences. Building on these successes, researchers have increasingly focused on adapting these concepts to graphs, which are composed of nodes and edges. While CNNs are well-suited for processing data with a regular, grid-like structure, such as images, graph convolutional networks (GCNs) are specifically designed to handle complex graph data where the relationships between elements are irregular and non-Euclidean\cite{TNNLS_GNN_Survey}. 

Graph neural networks have been widely applied to different tasks, including molecular property prediction~\cite{Neuro_Molecular}, drug recommendation~\cite{Neuro_Drug}, and diseases analysis~\cite{Neuro_Disease}. Among them, spatio-temporal graph neural networks (ST-GNNs) have emerged as a powerful paradigm for modeling dynamic graph-structured data, achieving state-of-the-art performance in applications such as motion prediction~\cite{Neuro_Motion}, traffic data imputation~\cite{Neuro_Traffic_Data} and rainfall forecasting~\cite{RINENG_Rainfall}. Recent advances in ST-GNNs have focused on building more powerful networks and have been applied in more domains. Cini et al.~\cite{cini2023scalable} and Tang et al.~\cite{tang2023explainable} focus on the scalability and explainability of spatio-temporal graph neural networks. Wang et al.~\cite{wang2024stgformer} integrated more advanced transformer networks or diffusion-based models into ST-GNNs. In traffic-related spatio-temporal modeling, Yang et al.~\cite{SSGCRTN} proposed SSGCRTN, a space-specific graph convolutional recurrent transformer network for traffic prediction. Yang et al.~\cite{MSTDFGRN} further proposed MSTDFGRN, which learns dynamic spatial dependencies and short- and long-range temporal patterns from multiple views. Yang et al.~\cite{PSTCGCN} introduced PSTCGCN, a principal spatio-temporal causal graph convolutional network that considers data drift and causal temporal dependencies. Yang et al.~\cite{TIIDGCN} proposed TIIDGCN, which incorporates temporal identity embeddings and multiscale interactive learning for traffic forecasting. Yang et al.~\cite{SDSINet} developed SDSINet, a spatiotemporal dual-scale interaction network that combines implicit temporal enhancement with dynamic multiscale spatial modeling. Yang et al.~\cite{GDGCRN} proposed GDGCRN, a general decoupled graph convolutional recurrent network that models multiscale temporal patterns, sensor heterogeneity, and signal decoupling. Yang et al.~\cite{MTEGCRN} introduced MTEGCRN, which combines temporal feature enhancement, node-oriented graph convolution, and global temporal fusion for traffic prediction. Yang et al.~\cite{DMFGCRN} further proposed DMFGCRN, which generates time-varying graph structures and progressively separates steady-state and non-steady-state traffic signals. In trajectory recovery, MReDTrajRec~\cite{MReDTrajRec} proposed a multi-representation data-driven model under road network constraints, integrating fine-grained trajectory representation with road semantic graph modeling. In cellular traffic prediction, Yang et al.~\cite{SGASeq} proposed SGA-Seq, a station-aware graph attention sequence network that combines learnable temporal embeddings, station-aware adaptive modeling, and progressive signal separation. In other application aspect, Verdone et al.\cite{VERDONE_APP1} proposed a novel explainable spatio-temporal graph neural network for energy forecasting by applying a spectral graph convolution network and a state-of-the-art explainer. Li et al.\cite{LI_APP2} proposed a multimodal adaptive spatio-temporal graph neural network for airspace complexity prediction, introducing a novel multimodal adaptive graph convolution module and a dilated causal temporal convolution with multi-step self-attention.

Given the irregular and non-Euclidean characteristics of ice layers, representing them as graphs and leveraging graph neural networks (GNNs) offers a more suitable approach for effectively modeling and processing such complex data structures. Zalatan et al. \cite{Zalatan2023, Zalatan_igarss, zalatan_icip} developed a network that combines an LSTM structure with a graph convolutional network (GCN) to solve the problem of ice layer tracking and ice thickness estimation. They used the fused spatio-temporal graph neural network, GCN-LSTM\cite{seo2016structured}, with an additional Evolve-GCN layer\cite{EGCN} as a prior adaptive layer. The fusion of LSTM and GCN architectures allows this network to learn both the temporal changes of the relationships between nodes in a graph and their spatial relationships at the same time, while the use of an adaptive layer improves the model performance by enabling the model to learn more complicated features and be more robust to noise. In the most recent work, Liu et al. proposed a novel ``Locate and Extend'' strategy\cite{liu2025locate} that reduces the computation time while maintaining a competitive accuracy compared with Zalatan et al.\cite{Zalatan_igarss,Zalatan2023,zalatan_icip}. To further improve the accuracy of Zatalan et al.\cite{Zalatan_igarss,Zalatan2023,zalatan_icip}, Liu et al.\cite{GRIT, ST-GRIT} combine the self-attention mechanism with a geometric deep learning framework and apply graph transformers on the task of deep ice layer thickness prediction.

Compared with Zalatan et al. \cite{Zalatan2023,Zalatan_igarss,zalatan_icip} that uses a fused spatio-temporal graph neural network, our proposed method uses different branches to learn spatial and temporal patterns separately. This multi-branch design enables a decoupled approach to spatial and temporal learning, allowing each branch to focus on extracting its respective features more effectively. Compared with Zalatan et al. \cite{Zalatan2023,Zalatan_igarss,zalatan_icip} and Liu et al.\cite{liu2024learningspatiotemporalpatternspolar, GRIT, ST-GRIT}, our methods is more training-efficient while maintaining competitive accuracy, achieving a better trade-off between model's accuracy and computational cost.

\subsection{Combining Domain Knowledge with Deep Learning}
Although machine learning algorithms have achieved great success in various domains, their pure data-driven nature often limits generalizability and interpretability, which are key requirements for scientific discovery. While machine learning excels at analyzing data, it can yield unrealistic or physically inconsistent predictions when applied to unseen data \cite{Karniadakis2021}. To address these limitations, researchers have explored knowledge-informed learning, where prior domain knowledge, such as insights from scientific models, empirical rules, or expert understanding, is integrated into machine learning workflows. This combination enhances the model’s generalizability, interpretability, and supports more meaningful outcomes in scientific discovery and engineering tasks~\cite{Desai2021}.

Some researchers have already applied the concept of knowledge-informed machine learning to study polar ice\cite{teisberg,DeepHybridWavelet,LearnSnowLayerThickness,PI_GCNLSTM,liu2024learningspatiotemporalpatternspolar}. However, given the scope of this paper, we focus specifically on prior studies that utilize knowledge-informed machine learning for processing radargrams and identifying ice layer boundaries\cite{varshney2021refining,DeepHybridWavelet,LearnSnowLayerThickness,PI_GCNLSTM,liu2024learningspatiotemporalpatternspolar}. Varsheny et al.\cite{varshney2021refining} integrates a physical wavelet transform block as a denoising part to a convolutional neural network designed to detect ice layer boudaries. Kamangir et al.\cite{DeepHybridWavelet} applied wavelet transformation as the pre-processing stage and developed a novel convolutional neural network to detect the layer boundaries more effectively. Varsheny et al.\cite{LearnSnowLayerThickness} highlights that regression-based neural networks can benefit from pretraining on labels simulated from physical models and have better accuracy on the thickness prediction. For the task of deep ice layer thickness prediction, in some recent work, liu et al.\cite{liu2024learningspatiotemporalpatternspolar} proposed a physics-informed GraphSAGE-LSTM network, and Rahnemoonfar et al.\cite{PI_GCNLSTM} applied a physics-informed AGCN-LSTM network that improves the model's accuracy by introducing physical node features synchronized from the physical weather model. 

In this paper, we applied a similar idea that incorporates data from a physical weather model as observational biases. This approach ensures that the model adheres to underlying physical principles as prior domain knowledge, enhances its generalization ability on unseen data, and thereby improves accuracy. Compared with Rahnemoonfar et al.\cite{PI_GCNLSTM} and liu et al.\cite{liu2024learningspatiotemporalpatternspolar}, our method improves both the accuracy and efficiency.

\section{Radargram Datasets and Graph Generation}\label{dataset}

\subsection{Radargram Dataset}
In this work, we use the Snow Radar Echogram Dataset (SRED) dataset\cite{ibikunle2025_Dataset}. It contains radargrams captured over the Greenland region in 2012 via an airborne snow radar sensor, operated by CReSIS\cite{CReSIS_radar} as part of NASA's Operation Ice Bridge project\cite{Leuschen2011SnowRadar}. The Snow Radar radargrams and the labeled images can be accessed at the CReSIS website. To the best of our knowledge, SRED is the first and only comprehensive AI-ready radar echogram dataset derived from airborne Snow Radar airborne measurements.

Ice-penetrating radar transmits electromagnetic waves into the ice sheet, leveraging their ability to propagate through dielectric media and reflect off interfaces where there are changes in dielectric constants. Depending on the radar system's frequency, bandwidth, and processing techniques, the captured reflections can be used to reconstruct the subsurface structure of the mapped topography with precision. 

The primary principle underpinning the radar campaigns is the analysis of the time-delay and amplitude of the reflected radar signals. The time delay represents the travel time of the electromagnetic wave from the transmitter to the reflecting interface and back to the receiver, scaled by the radar wave's velocity in snow or ice. This velocity varies with temperature, density, and the presence of unique subglacial structures such as melted or refrozen water. Using the time-delay and a corresponding density-depth model, accurate depth estimation can be achieved. Similarly, the received signal amplitude provides insight into the snow temporal deposition and material properties at each reflecting interface, dictated by the Fresnel reflection coefficient, which quantifies the magnitude of the dielectric discontinuity. Together, time delay and amplitude enable a detailed reconstruction of subsurface features, ranging from internal stratigraphic layers to basal hydrology.

As shown in Figure \ref{fig:dataset} (\textbf{a}), the Snow Radar transmits electromagnetic waves towards the polar ice sheet's surface as the radar-equipped aircraft flies over the ice sheet region with predefined flight routes. The S-C frequency band and 6 GHz bandwidth of the radar are optimized to penetrate the shallow snow layers while capturing backscatter from distinct internal stratigraphy due to its crisp vertical resolution of about 4~cm in snow. Interface between the late summer and early winter snowfall appears characteristically different to the radar receiver, resulting in annual layering seen in the resulting images. This is attributed to the difference in the seasonal weather patterns, creating a snow permittivity change that causes detectable backscatter towards the receiving radar antenna. Distinct peaks in the processed backscatter are further enhanced with various digital filtering and signal processing methods to create the radargram. This is one of the most effective ways to directly and distinctly measure internal snow layer stratigraphy attributed to annual snow accumulation. 

The layer's stratigraphy orientation and degree of curvature hold significant information about the historical deposition and subsequent metamorphosis of the snow at each location. This varies significantly across different regions of Greenland in response to factors such as topographic and climatic features. Areas with sloped and uneven terrain may have layers that display significant curvature due to gravitational settling and compaction. As such, the layer orientation in another radargram in the dataset may appear very different from the example in Figure \ref{fig:dataset}. By analyzing and tracking these curved contours, inferences about past climatic conditions and accumulation rate can be made. The depth of radargrams in the training set used varies from 1200 pixels to 1700 pixels, with a fixed 256 pixels in width.  Using the radargrams, corresponding labeled images are manually generated by tracking each layer in the radargram, as illustrated in Figure \ref{fig:dataset}(\textbf{c}). Based on the position of ice layer boundaries, layer thickness can be calculated as the difference between its upper and lower boundaries. Additional onboard equipment, such as the Inertial Measurement Units (IMUs), is used to track the aircraft's orientation and motion for precise data alignment, while the Global Navigation Satellite System GPS receiver is also used for real-time geolocation tagging of the aircraft's motion to simultaneously record the current latitude and longitude.

Furthermore, integrating auxiliary data such as climate model outputs with annotated radargrams for deep learning model training holds the potential for enhancing model performance by embedding broader environmental context and knowledge-informed realism. Climate data often capture variability in temperature, precipitation, and other climatic conditions, which provide additional crucial contextual information that helps the model's robustness and generalization across regions and seasons. In this work, we demonstrate the efficacy of knowledge-informed learning by integrating climatic data from a regional climate model.

\subsection{Modèle Atmosphérique Régional climate model}
\label{physical-feature}
The MAR (Modèle Atmosphérique Régional) is a high-resolution atmospheric regional climate model (RCM) specifically designed to simulate fine-scale meteorological and cryospheric processes. Particularly for polar and glaciated regions, including the Greenland ice sheet, MAR excels at modeling the complex interactions between the atmosphere, surface, and subsurface, which is crucial for studying climate dynamics \cite{MAR_2020, MAR2021}. At its core, MAR employs a non-hydrostatic dynamical framework that solves the primitive equations of atmospheric motion, enabling accurate simulation of vertical and horizontal atmospheric processes over Greenland's complex topography. The climate model uses the sigma-coordinate vertical discretization numerical weather prediction model to resolve surface-atmosphere interactions effectively, making it particularly adept at capturing mesoscale phenomena such as localized snow redistribution. 

MAR incorporates advanced physics parameterization to simulate cloud microphysics, radiation transfer, turbulence, and convection. Its coupling with the SISVAT (Soil Ice Snow Vegetation Atmosphere Transfer) scheme enables detailed representation of surface energy and mass exchanges, including snow densification, melt, refreezing, and sublimation. This fidelity is crucial for modeling the surface mass balance (SMB) of ice sheets, a key metric for understanding sea level rise and polar climate changes. MAR is also driven by lateral boundary conditions from global climate models (GCMs) \cite{morel1988introduction, washington2005introduction, mcguffie2014climate} or reanalysis datasets such as ERA-5 \cite{hersbach2020era5, soci2024era5}, allowing it to integrate large-scale atmospheric dynamics while resolving fine-scale local features. 

MAR provides a rich source of physically consistent data, which can serve as knowledge-informed priors to provide observational biases for machine learning models. Its ability to resolve surface-atmosphere interactions at fine resolutions positions it as a good candidate to improve the robustness of data-driven models with knowledge-informed constraints.

\subsection{Synchronizing Physical Data with Radargrams}

In this study, the dataset used contains radargrams whose along-track geolocations were already synchronized with the MAR version 3.10 model data that provides climate data across the entire Greenland ice sheet at 15 km grid resolution. Concretely, the radargram dataset has five key physical features: snow mass balance, surface temperature, meltwater refreezing, height change due to melting, and snowpack heights total provided as yearly measurements synchronized with each column of the radargram data. These features have been identified by researchers \cite{fettweis2013estimating, vernon2013surface,reeh1991parameterization} as critical metrics that provide insights into the dynamics of polar ice sheets and are central to understanding and monitoring the effect of climate change on polar ice sheets. 

Snow mass balance is the net difference between accumulation (snowfall) and ablation (melting, sublimation, and runoff). Consequently, a negative snow mass balance in Greenland signals a contribution to global sea level rise. Similarly, rising surface temperature directly influences snow mass balance by increasing melting rates, reducing snow cover, and lowering the albedo effect, which in turn, amplifies warming in the region to further increase melt rate. Meltwater refreezing, however, can contribute to snowpack mass because it is generated from surface melting that percolates into the snowpack but refreezes to partially offset the loss due to surface melting. The changes in the extent of meltwater refreezing are indicative of surface warming trends, with less freezing suggesting that warming snowpacks are unable to retain the meltwater. Height change due to melting is a measure of the mass loss due to ablation, and it is used to estimate the rate of ice loss and its contribution to sea level rise, while total snowpack height estimates all accumulated snow layers, affected by snowfall, compaction, and melt-refreeze cycles. The changes in the snowpack height reflect both seasonal and long-term changes in precipitation and temperature patterns.

These features are provided in the MAR data as daily outputs, but to synchronize with annual layering seen in radargrams, these outputs are summed using an annual interval that approximately coincides with the chronology of the radargram layers. An annual cycle that spans from September of one year to September of the following year was used to correctly reflect the snow layer captured by the sensor during the summer-to-winter transition.  

Conclusively, using latitude and longitude coordinates for each radargram, the estimated annual MAR data are aligned with the airborne measurements using the 2D Delaunay triangulation interpolation technique. This alignment enables the annual climate data to serve as supplementary information for improved predictions of internal layer thickness. 

\subsection{Graph Data Generation}
Our graph dataset is constructed based on thickness and geographical information extracted from radargram images, along with physical data obtained from the MAR climate reanalysis model. Each ground-truth radargram image is converted into a temporal sequence of $m$ spatial graphs, where each graph represents a shallow internal ice layer formed in a specific year. Each spatial graph contains 256 nodes, corresponding to the width of 256 pixels in the radargram images. We denote this sequence as $\mathbf{X} = \{\mathbf{X}_1, \mathbf{X}_2, \dots, \mathbf{X}_m\}$, where $\mathbf{X}_i \in \mathbb{R}^{256 \times d}$ represents the spatial graph at the $i$-th internal layer from the top, and $d$ is the total number of node features.

Each row in $\mathbf{X}_i$, denoted as $\mathbf{x}_i^{(v)} \in \mathbb{R}^d$, corresponds to the feature vector of node $v$ in the $i$-th spatial graph. In this work, the node feature vector $\mathbf{x}_i^{(v)}$ includes three non-physical base features—the latitude $\phi^{(v)}$, longitude $\lambda^{(v)}$, and ice thickness $h_i^{(v)}$ at year $i$—as well as five physical features from the MAR model: surface mass balance, annual surface temperature, meltwater refreezing-induced height change, melt-induced height change, and snowpack height. For the explanation of these physical features, please refer to Section~\ref{physical-feature}. Thus, the node feature vector is given by $\mathbf{x}_i^{(v)} = [\phi^{(v)}, \lambda^{(v)}, h_i^{(v)}, f_{i,1}^{(v)}, f_{i,2}^{(v)}, f_{i,3}^{(v)}, f_{i,4}^{(v)}, f_{i,5}^{(v)}] \in \mathbb{R}^8.$

All nodes within each spatial graph are fully connected by undirected edges. The edge weight between node $u$ and $v$ is defined as the inverse of their geometric distance, computed using the haversine formula:
\begin{equation}
\begin{split}
w_{u,v} & = \frac{1}{2\arcsin{\big( \mathrm{hav}(\phi^{(v)} - \phi^{(u)}) 
+ \cos{\phi^{(u)}} \cos{\phi^{(v)}} }} \\
& \qquad \mathrm{hav}(\lambda^{(v)} - \lambda^{(u)}) \big)
\end{split}
\end{equation}

where $\mathrm{hav}(\theta) = \sin^2\left(\frac{\theta}{2}\right).$

\section{Knowledge-Informed Spatio-Temporal Efficient Multi-Branch Graph Neural Network} \label{method}
In this paper, we introduce K-STEMIT, a novel knowledge-informed spatio-temporal efficient multi-branch graph neural network for ice thickness prediction. K-STEMIT integrates key ideas from knowledge-informed machine learning and multi-branch network architecture, designed to learn from the top $m$ internal ice layer and predict the thickness of $n$ layers beneath.

Unlike pure data-driven deep learning methods, knowledge-informed machine learning leverages domain knowledge, like physical laws or observations, to achieve physically consistent and reliable predictions. In this work, we introduce physical features for ice sheets synchronized from the MAR physical weather model as prior knowledge, ensuring the predicted ice layer thickness aligns with underlying physical phenomena.

The multi-branch network architecture is the core innovation of our network architecture. It allows each branch to focus on a specific task, enabling more effective and specialized learning. By separating the spatial and temporal learning processes, the model optimizes weight allocation for each branch, avoiding the compromise inherent in joint learning. This separation further enhances the model's ability to independently capture localized spatial relationships and temporal trends, leading to a more comprehensive understanding of the data.

\begin{figure}
    {
        \includegraphics[width=0.99\textwidth]{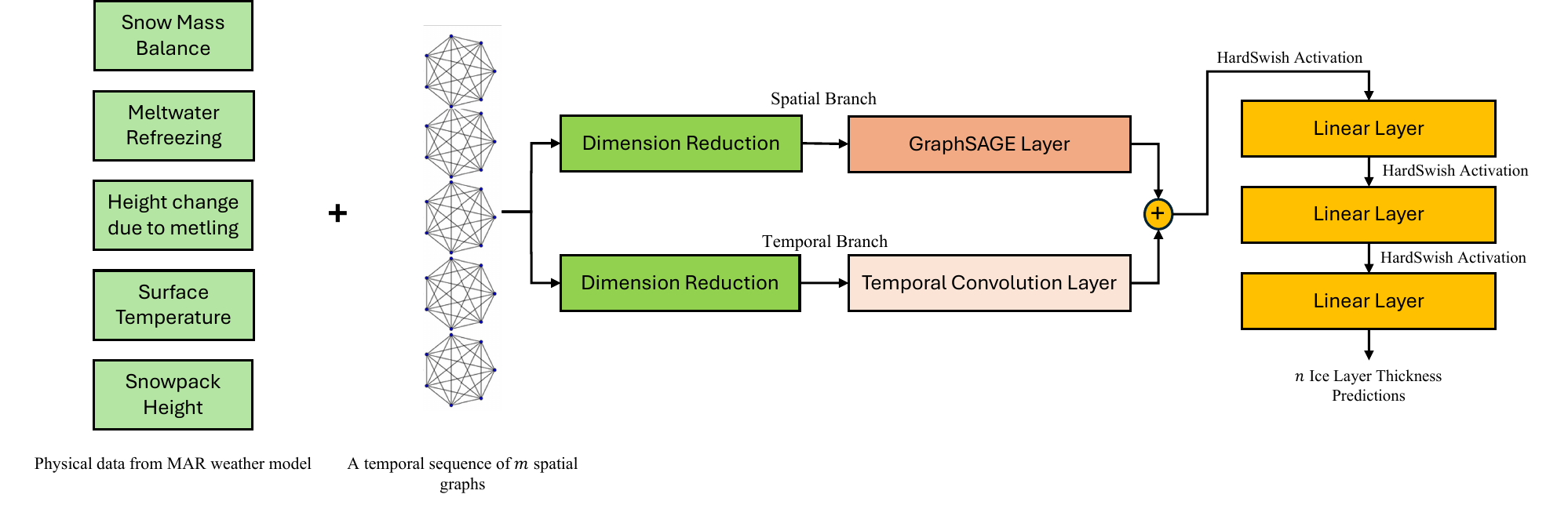}
    }
    \caption{Architecture of our proposed K-STEMIT.\label{fig:arch}}
\end{figure}

Figure~\ref{fig:arch} illustrates an overview of our proposed K-STEMIT. The network takes a temporal sequence $\mathbf{X} = \{\mathbf{X}_1, \mathbf{X}_2, \dots, \mathbf{X}_m\}$ as input, where each $\mathbf{X}_i$ is a spatial graph representing an internal ice layer formed at a specific year. These graphs are processed in parallel through two branches: a GraphSAGE-based spatial branch that captures localized spatial relationships, and a gated temporal convolution branch that extracts temporal trends across the sequence. Dimensionality reduction is applied at the beginning of each branch to eliminate irrelevant or redundant node features, promoting efficient feature extraction. The outputs from both branches are then combined adaptively and passed through a sequence of three fully connected layers with hardswish activation to predict the thickness of $n$ deeper ice layers. 

\subsection{Dimensionality Reduction}
To ensure that each branch focuses on the most relevant features, we introduce dedicated dimensionality reduction modules at the beginning of both the spatial and temporal branches, as illustrated in Figure~\ref{fig:arch}. These dimensionality reduction strategies help disentangle spatial and temporal dependencies, allowing each branch to operate on the most informative feature subsets while minimizing redundancy.

\begin{figure}
    \centerline
    {
        \includegraphics[width=0.5\textwidth]{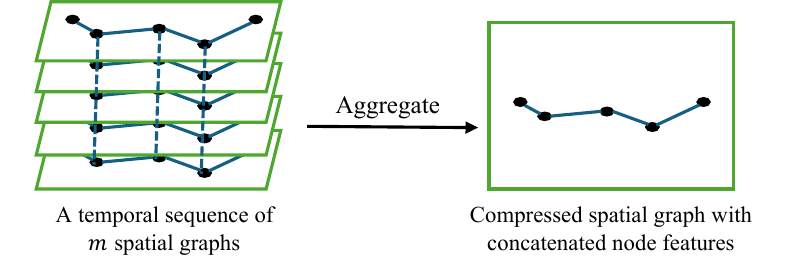}
    }
    \caption{Dimension reduction in the spatial branch, where the graph sequence is compressed into a single graph with concatenated node features.}
    \label{fig:nontemporal}
\end{figure}

For the spatial branch, we leverage the structural characteristics of radargrams: nodes located in the same column across different layers share the same geographic coordinates. In previous section, we denoted the input temporal sequence of spatial graphs as $\mathbf{X} = \{\mathbf{X}_1, \mathbf{X}_2, \dots, \mathbf{X}_m\}$, where each $\mathbf{X}_i \in \mathbb{R}^{256 \times 8}$ represents a spatial graph at the $i$-th internal layer from the top, and each node has 3 base features (latitude, longitude, thickness) and 5 physical features. In the spatial branch, we apply an aggregation operation to construct a compressed spatial graph, as shown in Figure~\ref{fig:nontemporal}. To build this compact spatial representation, we concatenate the node features across the temporal dimension:
\begin{equation}
\tilde{\mathbf{X}}_{\text{spatial}} = \text{Concat}(\mathbf{X}_1, \mathbf{X}_2, \dots, \mathbf{X}_m) \in \mathbb{R}^{256 \times 8m}.
\end{equation}
Since latitude and longitude are identical across all $m$ layers for each node, we retain only a single copy of these two features. The effective input dimensionality to the spatial branch is thus reduced to $8m - 2(m - 1) = 6m + 2$, where $8m$ represents the total number of features in the input temporal sequence of $m$ spatial graphs and $2(m - 1)$ corresponds to the repeated latitude and longitude features. The resulting feature matrix $\tilde{\mathbf{X}}_{\text{spatial}} \in \mathbb{R}^{256 \times (6m + 2)}$, along with the original graph connectivity, is then passed to a GraphSAGE layer to capture localized spatial dependencies.

In contrast, the temporal branch models the evolution of dynamic features over time. To construct its input, we start from the original input sequence \( \mathbf{X} = \{\mathbf{X}_1, \mathbf{X}_2, \dots, \mathbf{X}_m\} \), where each \( \mathbf{X}_i \in \mathbb{R}^{256 \times 8} \) contains 3 base features and 5 physical features. We remove the static geographic components—latitude and longitude—from all \( \mathbf{X}_i \), retaining only the dynamic thickness and physical features. We define the resulting input as:
\begin{equation}
\tilde{\mathbf{X}}_{\text{temporal}} = \{\tilde{\mathbf{X}}_1, \tilde{\mathbf{X}}_2, \dots, \tilde{\mathbf{X}}_m\} \in \mathbb{R}^{m \times 256 \times 6},
\end{equation}
where each $\tilde{\mathbf{X}}_i \in \mathbb{R}^{256 \times 6}$ represents the reduced feature matrix at a certain year, excluding latitude and longitude. This temporal representation is then processed by a gated temporal convolution block to model the temporal dynamics of each node across layers.

\subsection{Multi-Branch Architecture and Adaptive Feature Fusion}
\label{sec:feature-fusion}

After applying dimensionality reduction, the spatial and temporal branches process their respective inputs in parallel to extract complementary structural information. The spatial branch takes the compressed representation $\tilde{\mathbf{X}}_{\text{spatial}} \in \mathbb{R}^{256 \times (6m + 2)}$, together with the original spatial connectivity, and passes it through a GraphSAGE layer to learn localized spatial dependencies among nodes. The temporal branch operates on the reduced input $\tilde{\mathbf{X}}_{\text{temporal}} \in \mathbb{R}^{256 \times m \times 6}$, which preserves the temporal evolution of dynamic features for each node, and processes it using a gated temporal convolution block.

The outputs from the two branches—denoted as $\mathbf{h}_{\text{spatial}} \in \mathbb{R}^{256 \times d'}$ and $\mathbf{h}_{\text{temporal}} \in \mathbb{R}^{256 \times d'}$, respectively—are then fused to form a unified representation. Instead of simple averaging or concatenation, we use a learnable scalar parameter $\alpha \in [0,1]$ to control the relative contribution of the two branches. Importantly, $\alpha$ is a global model-level parameter shared across all nodes and samples, rather than a spatially or temporally varying coefficient. The final fused feature is computed as:
\begin{equation}
\mathbf{h} = \alpha \cdot \mathbf{h}_{\text{spatial}} + (1 - \alpha) \cdot \mathbf{h}_{\text{temporal}}.
\label{equ:adaptive}
\end{equation}

Under this design, a larger learned $\alpha$ indicates that the trained model relies more heavily on the spatial branch overall, whereas a smaller $\alpha$ indicates a greater contribution from the temporal branch. Because $\alpha$ is learned as a single fixed parameter after training, the current K-STEMIT does not model region-specific or depth-specific variations in feature fusion. The fused feature $\mathbf{h}$ is then passed through three fully connected layers with hardswish activation to produce the final prediction for the thickness of the bottom $n$ ice layers. To further improve architectural clarity, Table~\ref{tab:arch_dims} summarizes the input and output tensor shapes of the main components in the proposed K-STEMIT.

\begin{table*}[t]
\centering
\caption{Stage-wise summary of tensor dimensions in K-STEMIT for the experimental setting used in this study. Batch dimension is omitted for clarity. Here, $m$ denotes the number of observed upper ice layers and $n$ denotes the number of target deeper layers. Hidden dimensions are kept symbolic where appropriate to preserve generality.}
\label{tab:arch_dims}
\resizebox{\textwidth}{!}{
\begin{tabular}{lllp{8.3cm}}
\toprule
\textbf{Stage} & \textbf{Input shape} & \textbf{Output shape} & \textbf{Description} \\
\midrule
Input graph sequence 
& $\mathbf{X}\in\mathbb{R}^{m\times 256\times 8}$ 
& $\mathbf{X}\in\mathbb{R}^{m\times 256\times 8}$ 
& Sequence of $m$ observed spatial graphs. Each graph contains 256 nodes, and each node has 8 input features. \\

Spatial branch: dimension reduction 
& $\mathbb{R}^{m\times 256\times 8}$ 
& $\tilde{\mathbf{X}}_{\text{spatial}}\in\mathbb{R}^{256\times (6m+2)}$ 
& Rearranges and compresses the multi-layer node features into a single graph representation for spatial modeling. \\

Spatial branch: GraphSAGE layer 
& $\mathbb{R}^{256\times (6m+2)}$ 
& $\mathbf{h}_{\text{spatial}}\in\mathbb{R}^{256\times d'}$ 
& Learns localized spatial dependencies over the graph topology from the compressed node features. \\

Temporal branch: dimension reduction 
& $\mathbb{R}^{m\times 256\times 8}$ 
& $\tilde{\mathbf{X}}_{\text{temporal}}\in\mathbb{R}^{m\times 256\times 6}$ 
& Projects each observed layer into a reduced node-wise feature representation for temporal modeling. \\

Temporal branch: temporal convolution 
& $\mathbb{R}^{256\times m\times 6}$ 
& $\mathbf{h}_{\text{temporal}}\in\mathbb{R}^{256\times d'}$ 
& Extracts cross-layer temporal dependencies for each node across the $m$ observed layers. \\

Fusion (element-wise addition) 
& $\mathbf{h}_{\text{spatial}},\mathbf{h}_{\text{temporal}}\in\mathbb{R}^{256\times d'}$ 
& $\mathbf{h}\in\mathbb{R}^{256\times d'}$ 
& Fuses the spatial and temporal branch outputs; both branches share the same fusion dimension $d'$. \\

Prediction head: linear block 1 
& $\mathbb{R}^{256\times d'}$ 
& $\mathbb{R}^{256\times d_{h1}}$ 
& First fully connected transformation, followed by HardSwish activation. \\

Prediction head: linear block 2 
& $\mathbb{R}^{256\times d_{h1}}$ 
& $\mathbb{R}^{256\times d_{h2}}$ 
& Second fully connected transformation, followed by HardSwish activation. \\

Prediction head: output layer 
& $\mathbb{R}^{256\times d_{h2}}$ 
& $\hat{\mathbf{Y}}\in\mathbb{R}^{256\times n}$ 
& Produces $n$ predicted ice-layer thickness values for each node. \\
\bottomrule
\end{tabular}}
\end{table*}

Overall, this multi-branch design enables the model to leverage both localized spatial structures and temporal dynamics in a complementary manner, leading to more accurate and physically consistent predictions.

\subsection{GraphSAGE Inductive Framework}
GraphSAGE~\cite{hamilton2018GraphSAGE} is an inductive GNN framework that generates node embeddings by aggregating information from a node's local neighborhood~\cite{ZHOU202057}. In our spatial branch, we apply GraphSAGE to the compressed spatial graph with node feature matrix $\tilde{\mathbf{X}}_{\text{spatial}} \in \mathbb{R}^{256 \times (6m + 2)}$, where each node corresponds to a fixed location across layers and has concatenated feature vectors derived from all top $m$ internal layers.

Let \( \tilde{\mathbf{x}}^{(v)} \in \mathbb{R}^{6m + 2} \) denote the input feature of node \( v \) in \( \tilde{\mathbf{X}}_{\text{spatial}} \), and let \( \mathcal{N}(v) \) denote its neighborhood. GraphSAGE updates the node representation as follows:

\begin{equation}
\mathbf{x}'^{(v)} = \mathbf{W}_1 \tilde{\mathbf{x}}^{(v)} + \mathbf{W}_2 \cdot \text{AGG}_{u \in \mathcal{N}(v)} \tilde{\mathbf{x}}^{(u)},
\label{equation:graphsage}
\end{equation}

where $\mathbf{x}'^{(v)} \in \mathbb{R}^{d'}$ is the updated node embedding, $\mathbf{W}_1$ and $\mathbf{W}_2$ are learnable weight matrices, and $\textit{AGG}$ is an aggregation function such as $\textit{mean}$, $\textit{max}$, or $\textit{LSTM}$. Here we use mean as the aggregation function. This formulation allows the model to capture localized spatial dependencies while enabling inductive generalization to unseen radargram inputs.

Compared to Graph Convolutional Network (GCNs)\cite{GCN}, GraphSAGE offers several advantages that make it a more suitable choice for our deep ice layer thickness prediction task. GCNs operate in the spectral domain, relying on the graph Laplacian and assuming a fixed global graph structure\cite{hamilton2018GraphSAGE,ZHOU202057}. This assumption becomes problematic in our task, where the underlying graph topology may vary significantly between radargrams collected from different regions. In contrast, GraphSAGE is a spatial method that performs feature aggregation directly in the graph domain, without dependence on a fixed global structure. This flexibility allows it to generalize more effectively across diverse ice layer graphs.

Another key distinction between GraphSAGE and GCN is the use of separate weight matrices in GraphSAGE. As shown in Equation~\ref{equation:graphsage}, the transformation applied to a node's own features is decoupled from the transformation applied to its aggregated neighbor features. This separation enables the model to independently learn how a node integrates contextual information while preserving its unique identity. We view the term $\mathbf{W}_1 \tilde{\mathbf{x}}^{(v)}$ as functionally analogous to a residual connection, which is particularly important in our setting where each node encodes geophysically meaningful properties, such as ice thickness and localized physical variables. Preserving these features throughout the network helps maintain node-specific signals and mitigates early-stage over-smoothing before deeper layers like attention-based encoders. As a result, GraphSAGE better supports generalization and robustness in learning from spatially and temporally varying radargram data.

\subsection{Temporal Convolution}
In this paper, we use a gated temporal convolution block\cite{gehring2017convolutional} that extracts temporal patterns from node features via gated 2D convolution and skip connection. 

\begin{figure}
    \centerline
    {
        \includegraphics[width=0.25\textwidth]{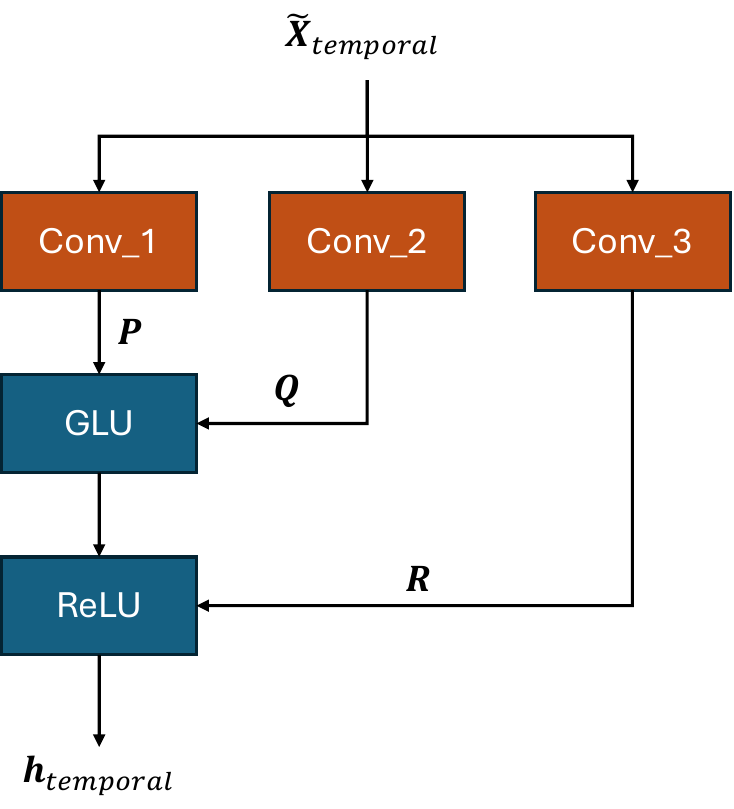}
    }
    \caption{Diagram of Temporal Convolution Block.}
    \label{fig:temporal}
\end{figure}

As shown in Figure \ref{fig:temporal}, the input tensor $\tilde{\mathbf{X}}_{\text{temporal}} \in \mathbb{R}^{256 \times m \times 6}$ is passed into three two-dimensional convolutions $\textit{conv}\_1, \textit{conv}\_2, \textit{conv}\_3$, producing intermediate outputs \( \mathbf{P}, \mathbf{Q}, \mathbf{R} \), respectively. These three outputs are used at different levels with skip connections. $\mathbf{P}$ and $\mathbf{Q}$ are passed into a Gated Linear Unit (GLU) to introduce non-linearity, defined as $ \mathbf{P} \times \sigma(\mathbf{Q})$, where $\sigma$ is the sigmoid function and $\times$ denotes element-wise Hadamard product. The output of GLU is added to the original convolution output $\mathbf{R}$, and a ReLU activation is applied to obtain the learned temporal feature $\mathbf{h}_{\text{temporal}}$. Therefore, we can express the gated temporal convolution as:
\begin{equation}
\mathbf{h}_{\text{temporal}} = \text{ReLU}(\text{GLU}(\mathbf{P}, \mathbf{Q}) + \mathbf{R}),
\end{equation}
where $\text{GLU}(\mathbf{P}, \mathbf{Q}) = \mathbf{P} \times \sigma(\mathbf{Q})$.

\section{Experiment Details}\label{experiments}

\subsection{Data Preprocessing}
We primarily evaluate our proposed K-STEMIT in the main setting with $m=5$ and $n=15$. In this case, the thickness and geometric information from the top 5 ice layers (2007--2011) are used to predict the thickness of the underlying 15 layers (1992--2006). To ensure high data quality and sufficient information content, we exclude radargrams with fewer than 20 complete layers, as well as radargrams containing incomplete traces within the required layers. After preprocessing, 1660 radargrams remain, which are randomly divided into 996 training images, 332 validation images, and 332 testing images.

To further evaluate generalization across different prediction depths, we construct a separate controlled subset by retaining only radargrams with at least 26 complete layers. On this fixed subset, the input is always the top 5 layers, so that $m=5$ remains unchanged, while the prediction depth is varied as $n \in \{5, 8, 10, 13, 15, 18, 21\}$ by using only the first $n$ deeper layers as prediction targets. This protocol ensures that comparisons across different $n$ values are performed on the same set of radargrams, thereby avoiding the confounding effect of depth-dependent dataset filtering.

\subsection{Training Details}\label{training}
To evaluate the effectiveness of our proposed K-STEMIT, we compare it with the non-knowledge-informed version STEMIT, current state-of-the-art methods for deep ice layer thickness prediction, and other existing models such as Graph Attention Network (GAT)~\cite{GAT} and Graph Isomorphism Network(GIN)~\cite{GIN}. All experiments are conducted using 8 NVIDIA A5000 GPUs and an Intel(R) Xeon(R) Gold 6430 CPU. The mean-squared error is used as the loss function, and the Adam optimizer~\cite{kingma2017adam} is adopted for all models. For AGCN-LSTM, GCN-LSTM, and GraphSAGE-LSTM, we set the initial learning rate to 0.01, the weight decay coefficient to 0.0001, and use a step learning rate scheduler that halves the learning rate every 75 epochs. For our proposed multi-branch spatio-temporal graph neural network and its physics-informed version, we use an initial learning rate of $5\times10^{-3}$, a weight decay coefficient of $10^{-5}$, and a cosine annealing learning rate scheduler with $T_{\max}=450$ and $\eta_{\min}=10^{-7}$. All models are trained for 450 epochs, which we found sufficient for convergence.

To reduce the effect of randomness, five different versions of the training, validation, and test sets are generated by applying distinct random permutations to the full set of 1660 radargrams before train-validation-test splitting. All graph neural networks are trained and evaluated on all five versions. Let $\mathcal{D}^{(k)} = \{(\hat{\mathbf{y}}_i^{(k)}, \mathbf{y}_i^{(k)})\}_{i=1}^{N_k}$ denote the test set from the $k$-th permutation, where $\hat{\mathbf{y}}_i^{(k)}, \mathbf{y}_i^{(k)} \in \mathbb{R}^{256 \times n}$ are the predicted and ground-truth ice thicknesses for the bottom $n$ layers across 256 spatial nodes of the $i$-th radargram in the test set, and $N_k$ is the total number of radargrams in the $k$-th test split, for $k = 1, 2, \dots, 5$. In the main $m=5, n=15$ setting, we have $N_k = 332$ for all five test splits. The code of K-STEMIT can be found at \url{https://github.com/BinaLab/K-STEMIT}.

The Root Mean Squared Error (RMSE) and Mean Absolute Error (MAE) for each version are computed as:
\begin{equation}
\mathrm{RMSE}^{(k)} = \sqrt{ \frac{1}{N_k \cdot 256 \cdot n} 
\sum_{i=1}^{N_k} \left\| \hat{\mathbf{y}}_i^{(k)} - \mathbf{y}_i^{(k)} \right\|_F^2 },
\end{equation}
\begin{equation}
\mathrm{MAE}^{(k)} = \frac{1}{N_k \cdot 256 \cdot n} 
\sum_{i=1}^{N_k} \left\| \hat{\mathbf{y}}_i^{(k)} - \mathbf{y}_i^{(k)} \right\|_1.
\end{equation}
where $N_k$ is the total number of radargrams in version $k$ test dataset, 256 refers to the 256 pixels per radargram, and $n$ refers to the number of deep ice layers that we predict.

We report the final performance as the mean and standard deviation of these metrics across the five runs:
\begin{equation}
\begin{split}
\overline{\mathrm{RMSE}} &= \frac{1}{5} \sum_{k=1}^{5} \mathrm{RMSE}^{(k)}, \\
\mathrm{Std}_{\mathrm{RMSE}} &= \sqrt{ \frac{1}{5} \sum_{k=1}^{5} 
\left( \mathrm{RMSE}^{(k)} - \overline{\mathrm{RMSE}} \right)^2 }.
\end{split}
\end{equation}

\begin{equation}
\begin{split}
\overline{\mathrm{MAE}} &= \frac{1}{5} \sum_{k=1}^{5} \mathrm{MAE}^{(k)}, \\
\mathrm{Std}_{\mathrm{MAE}} &= \sqrt{ \frac{1}{5} \sum_{k=1}^{5} 
\left( \mathrm{MAE}^{(k)} - \overline{\mathrm{MAE}} \right)^2 }.
\end{split}
\end{equation}
To stabilize the training process and allow our proposed multi-branch spatio-temporal graph neural network to naturally learn the relative contribution of each branch, we do not apply any constraints to the learnable fusion parameter $\alpha$ in Equation~\ref{equ:adaptive}. After training, we verify that $\alpha$ consistently remains positive across all runs, indicating stable and interpretable adaptive fusion between the spatial and temporal branches.

\subsection{Hyperparameter Selection and Sensitivity}
To improve reproducibility and clarify how the training hyperparameters were chosen, we performed a targeted grid search for K-STEMIT based on prior work and preliminary validation experiments. We varied the initial learning rate, weight decay, and learning-rate scheduler, and used the validation RMSE as the main selection criterion. Specifically, we tested learning rates of $10^{-2}$, $5\times10^{-3}$, and $10^{-3}$, weight decay values of $10^{-4}$, $10^{-5}$, and $10^{-6}$, step-scheduler decay intervals of 50, 75, 100, and 150 epochs, and cosine-annealing minimum learning rates of $10^{-5}$, $10^{-6}$, and $10^{-7}$. The final configuration uses an initial learning rate of $5\times10^{-3}$, weight decay of $10^{-5}$, and a cosine annealing scheduler with $T_{\max}=450$ and $\eta_{\min}=10^{-7}$. Table~\ref{tab:hparam_sensitivity} summarizes the validation RMSE values for all tested configurations and shows that the selected setting achieved the best overall performance among the examined cases.

\begin{table*}[t]
\centering
\small
\caption{Hyperparameter sensitivity study for K-STEMIT. Each case varies one component around the final configuration unless otherwise noted. Validation RMSE is reported for model selection. For the step scheduler, ``Scheduler arg'' denotes the decay interval in epochs; for cosine annealing, it denotes $T_{\max}$. The final selected configuration is highlighted in bold.}
\label{tab:hparam_sensitivity}
\begin{tabular}{cccccc}
\toprule
Learning rate & Weight decay & Scheduler & Scheduler arg & $\eta_{\min}$ & Validation RMSE \\
\midrule
$10^{-2}$        & $10^{-5}$ & Cosine & 450 & $10^{-7}$ & 2.5620 $\pm$ 0.0571 \\
$10^{-3}$        & $10^{-5}$ & Cosine & 450 & $10^{-7}$ & 2.7744 $\pm$ 0.0691 \\
$5\times10^{-3}$ & $10^{-4}$ & Cosine & 450 & $10^{-7}$ & 2.4382 $\pm$ 0.0917 \\
$5\times10^{-3}$ & $10^{-6}$ & Cosine & 450 & $10^{-7}$ & 2.4137 $\pm$ 0.0730 \\
$5\times10^{-3}$ & $10^{-5}$ & Step   & 50  & --         & 2.6098 $\pm$ 0.0902 \\
$5\times10^{-3}$ & $10^{-5}$ & Step   & 75  & --         & 2.5148 $\pm$ 0.0656 \\
$5\times10^{-3}$ & $10^{-5}$ & Step   & 100 & --         & 2.4667 $\pm$ 0.0975 \\
$5\times10^{-3}$ & $10^{-5}$ & Step   & 150 & --         & 2.5043 $\pm$ 0.1200 \\
$5\times10^{-3}$ & $10^{-5}$ & Cosine & 450 & $10^{-5}$  & 2.4294 $\pm$ 0.0951 \\
$5\times10^{-3}$ & $10^{-5}$ & Cosine & 450 & $10^{-6}$  & 2.4264 $\pm$ 0.0930 \\
$\mathbf{5\times10^{-3}}$ & $\mathbf{10^{-5}}$ & \textbf{Cosine} & \textbf{450} & $\mathbf{10^{-7}}$ & \textbf{2.4097 $\pm$ 0.0768} \\
\bottomrule
\end{tabular}
\end{table*}
\section{Result}\label{results}

\subsection{Overall Performance}\label{overall}
We start by evaluating the overall performance of different methods. The mean of the training time and the mean and standard deviations of the prediction error (RMSE, MAE) over the five versions of the training, validation, and testing datasets are reported as the model efficiency and accuracy. Table~\ref{table:OverallResults} summarizes the accuracy and efficiency of each model. In the table, STEMIT refers to K-STEMIT without knowledge-informed learning, Multi-Branch refers to the minimum variant of K-STEMIT that only includes the multi-branch architecture, and Knowledge-Informed Multi-Branch refers to K-STEMIT without adaptive feature fusion strategy.

\begin{figure}
\centering
\includegraphics[width=0.45\textwidth]{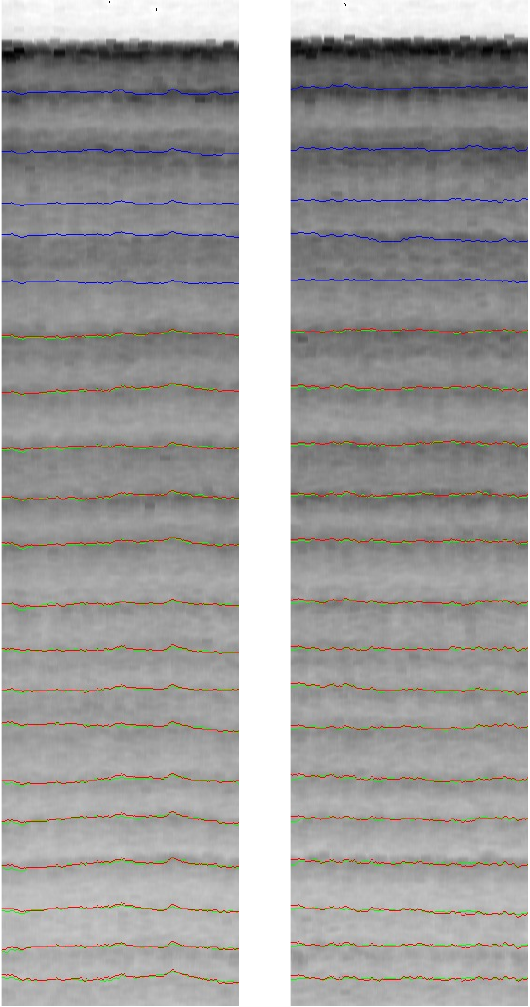}
\caption{Qualitative results of our proposed K-STEMIT. The blue line is used to generate the graphs. The green line is the groundtruth (manually-labeled ice layers) and the red line is the model prediction.\label{fig:qualitative_PIMBSTGNN}}
\end{figure}  

\begin{table*}[!t]
    \caption{Experiment results of K-STEMIT, its different variants, current state-of-the-art models and other baselines. Results are reported as the mean and standard deviation of the RMSE and MAE on the test dataset over five individual trials. Train time is reported as the average train time over five individual trials for 450 epochs.}
    \centering
    \scalebox{0.7}{
    \begin{tabular}{cccccc}
    \toprule
\textbf{Model}              &\textbf{Physical Node Feature?}   & \textbf{RMSE} & \textbf{MAE} & \textbf{Average Training Time (Seconds)} \\ 
\midrule
GCN\cite{kipf2017GCN} & x & 5.0573 $\pm$ 0.0874 & 3.7655 $\pm$ 0.0509 & 1099\\
GIN\cite{GIN} & x & 3.5696 $\pm$ 0.0591 & 2.5416 $\pm$ 0.0521 & 862 \\
AGCN-LSTM\cite{zalatan_icip}  & x            & 3.4808 $\pm$ 0.0397 & 2.4486 $\pm$ 0.0438 & 9404 \\
GCN-LSTM\cite{Zalatan2023}    & x          & 3.1745 $\pm$ 0.1045 & 2.2400 $\pm$ 0.0923 & 7441 \\
GraphSAGE-LSTM\cite{liu2024learningspatiotemporalpatternspolar} & x  & 3.3837 $\pm$ 0.1103 & 2.4380 $\pm$ 0.1171 & 4579 \\
GraphSAGE\cite{hamilton2018GraphSAGE} & x & 3.1873 $\pm$ 0.0588 & 2.1831 $\pm$ 0.0233 & 842 \\
GAT\cite{GAT} & x & 3.1367 $\pm$ 0.0541 & 2.1718 $\pm$ 0.0332 & 1355  \\
Locate and Extend\cite{liu2025locate} & x & 3.1958 $\pm$ 0.0267 & 2.2226 $\pm$ 0.0204 & 993\\
GRIT\cite{GRIT} & x &3.0597 $\pm$ 0.0326 & 2.0713 $\pm$ 0.0200 & 1649 \\
ST-GRIT\cite{ST-GRIT} & x & 2.8866 $\pm$ 0.0569 & 1.9141 $\pm$ 0.0304 & 2459 \\
STEMIT(Ours) & x & 3.1170 $\pm$ 0.0780 & 2.1437 $\pm$ 0.0349  & 1005 \\
Multi-Branch (Ours) & x & 3.0980 $\pm$ 0.0857 & 2.1234 $\pm$ 0.0410 & 990\\

\midrule
Physics-Informed AGCN-LSTM\cite{PI_GCNLSTM}&\checkmark & 3.4441 $\pm$ 0.0513 & 2.4389 $\pm$ 0.0528 & 9554 \\
Knowledge-Informed GIN\cite{GIN} & \checkmark & 3.3438 $\pm$ 0.0991 & 2.3618 $\pm$ 0.0800 & 863\\
Knowledge-Informed GCN\cite{kipf2017GCN} & \checkmark & 3.3298 $\pm$ 0.0491 & 2.3168 $\pm$ 0.0451 & 1127 \\
Physics-Informed GraphSAGE-LSTM\cite{liu2024learningspatiotemporalpatternspolar} & \checkmark & 2.9103 $\pm$ 0.0830 & 2.0341 $\pm$ 0.0660 & 4686 \\
Knowledge-Informed GraphSAGE\cite{hamilton2018GraphSAGE} & \checkmark & 2.6173 $\pm$ 0.0835 & 1.7340 $\pm$ 0.0424 & 926 \\
Knowledge-Informed GAT\cite{GAT} & \checkmark & 2.5865 $\pm$ 0.0648 & 1.7131 $\pm$ 0.0300 & 1361 \\
Knowledge-Informed Multi-Branch (Ours) & \checkmark & 2.4892 $\pm$ 0.0556 & 1.6174 $\pm$ 0.0303 & 1069 \\
\textbf{K-STEMIT(Ours)} & \checkmark & \textbf{2.4168 $\pm$ 0.0553} & \textbf{1.5714 $\pm$ 0.0328} &\textbf{1016}\\
\bottomrule
\end{tabular}
}
    \label{table:OverallResults}
\end{table*}

From Table~\ref{table:OverallResults}, we see that our proposed K-STEMIT outperforms both current state-of-the-art methods with and without physical node features, with the lowest RMSE error and nearly the lowest average computation time. Figure \ref{fig:qualitative_PIMBSTGNN} presents qualitative results of thickness predictions generated by our proposed K-STEMIT network. The results demonstrate the network's ability to effectively learn patterns from shallow ice layers and accurately predict those in deeper layers. 

Compared with the current state-of-the-art non-physical ST-GRIT, K-STEMIT achieves a 16.28\% reduction in RMSE and is about 2.4 times faster than ST-GRIT. These results highlight the effectiveness of knowledge-informed learning. By introducing those physical node features from the MAR physical weather model as prior knowledge, our proposed K-STEMIT can yield better network capacity without any tradeoff between accuracy and computational time. These auxiliary physical features act as domain-informed priors during training, providing weak constraints that guide the model toward more physically plausible predictions and help mitigate unrealistic outputs. Therefore, K-STEMIT can achieve better generalization ability on unseen data and outperform the most advanced network architectures with no knowledge integrated. Compared with the current state-of-the-art knowledge-informed graph neural networks, K-STEMIT gets a 16.96\% reduction in RMSE and is about 4.6 times faster. Compared with the multi-branch network, fundamental variant of K-STEMIT that only applies the multi-branch architecture, our proposed K-STEMIT achieves 21.99\% RMSE reduction by integrating physical node features and applying the adaptive feature fusion strategy.

We attribute the improvement in efficiency to the use of a multi-branch architecture. Compared with those fused spatio-temporal graph neural networks (AGCN-LSTM, GCN-LSTM, GraphSAGE-LSTM) and graph transformers (GRIT, ST-GRIT), the multi-branch architecture used in K-STEMIT dramatically reduces the total number of operations in the network's forward pass. Instead of jointly processing both spatial and temporal features within a single, often over-parameterized fusion module, K-STEMIT decouples the learning process into dedicated branches for spatial and temporal modeling. This allows each branch to use a more specialized network design, reducing redundant computations and avoiding unnecessary entanglement between spatial and temporal representations. We also notice that this decoupled multi-branch network brings about 2.4\% reduction in RMSE compared with previous fused spatio-temporal graph neural networks, GCN-LSTM. Unlike previous methods, where the model learned spatial and temporal features as a mixture, the multi-branch architecture allows each branch to focus on its specific strengths and optimize its performance on either the spatial or temporal domain. Moreover, unlike graph transformers that rely on attention mechanisms with quadratic complexity in the number of nodes or tokens, each branch in K-STEMIT only uses lightweight and specialized layers like GraphSAGE and temporal convolutions with less complexity, resulting in a more efficient training process. 

Having established the strong overall performance of K-STEMIT, we next conduct a series of ablation studies to investigate the impact of specific design choices. We begin by analyzing the impact of physical feature selection, isolating its effect on model performance, and subsequently exploring the contribution of different architecture branches.

\subsection{Ablation Study: Choice of Physical Features}
With the network architecture fixed, we first examine how the inclusion of various physical features influences the model's predictive capability. Specifically, we conduct an ablation study to assess different contributions of the five MAR features in the SRED dataset. As shown in Table \ref{table:ablation_physics}, we find that the combination of average yearly surface temperature, height change due to melting, and snowpack height gives the overall best performance for our proposed K-STEMIT. This highlights the complementary role of physical inputs in enhancing the capability of the network architecture. The varying RMSE observed across different feature combinations suggests that the model benefits most from a balanced set of physical features, where the added variables provide useful context without introducing excessive noise or redundancy.

\begin{table*}[!t]
\caption{Ablation study on the choice of different physical features}
\begin{center}
\resizebox{\textwidth}{!}{
\begin{tabular}{ccccccc}
\toprule
\textbf{Snow Mass Balance} & \textbf{Average Yearly Surface Temperature} & \textbf{Height Change Due To Refreezing} & \textbf{Height Change Due To Melting} & \textbf{Snowpack Height} & \textbf{Average RMSE} & \textbf{Improvement(\%)}\\ \midrule
 & & & & & 3.1170 $\pm$ 0.0780 & 0.00 \\
 & & & &\checkmark & 2.4647 $\pm$ 0.0588 & 20.93 \\
 & & &\checkmark & & 2.8627 $\pm$ 0.0798 & 8.15 \\
 & & &\checkmark &\checkmark & 2.4603 $\pm$ 0.0657 & 21.07 \\
 & & \checkmark & & & 2.8877 $\pm$ 0.0621 & 7.36 \\
 & & \checkmark & &\checkmark & 2.4831 $\pm$ 0.0544 & 20.34 \\
 & & \checkmark &\checkmark & & 2.8288 $\pm$ 0.0612 & 9.24\\
 & & \checkmark &\checkmark &\checkmark & 2.5130 $\pm$ 0.0694 & 19.38\\
 &\checkmark & & & & 3.0149 $\pm$ 0.0727 & 3.27\\
 &\checkmark & & &\checkmark & 2.4647 $\pm$ 0.0653 & 20.92\\
 &\checkmark & &\checkmark & & 2.8231 $\pm$ 0.0880 & 9.43\\
 &\checkmark & &\checkmark &\checkmark & \textbf{2.4168 $\pm$ 0.0553} & \textbf{22.46} \\
 &\checkmark &\checkmark & & &2.8138 $\pm$ 0.0536 & 9.72\\
 &\checkmark &\checkmark & &\checkmark & 2.4629 $\pm$ 0.0498 & 20.98\\
 &\checkmark &\checkmark &\checkmark & & 2.7983 $\pm$ 0.0605 & 10.23\\
 &\checkmark &\checkmark &\checkmark &\checkmark & 2.4644 $\pm$ 0.0587 & 20.94\\
\checkmark & & & & & 2.9362 $\pm$ 0.0670 & 5.80\\
\checkmark & & & &\checkmark & 2.4529 $\pm$ 0.0614 & 21.30\\
\checkmark & & &\checkmark & & 2.7353 $\pm$ 0.0504 & 12.25\\
\checkmark & & &\checkmark &\checkmark & 2.4557 $\pm$ 0.0543 & 21.21\\
\checkmark & &\checkmark & & & 2.7553 $\pm$ 0.0580 & 11.60\\
\checkmark & &\checkmark & &\checkmark & 2.4674 $\pm$ 0.1037 & 20.90\\
\checkmark & &\checkmark & \checkmark &  & 2.7384 $\pm$ 0.0574 & 12.15\\
\checkmark & &\checkmark &\checkmark &\checkmark & 2.4697 $\pm$ 0.0710 & 20.77\\
\checkmark &\checkmark & & & & 2.8963 $\pm$ 0.0567 & 7.08\\
\checkmark &\checkmark & & &\checkmark & 2.4587 $\pm$ 0.0710 & 21.12\\
\checkmark &\checkmark & &\checkmark & & 2.7378 $\pm$ 0.0683 & 12.17\\
\checkmark &\checkmark & &\checkmark &\checkmark & 2.4455 $\pm$ 0.0553 & 21.54\\
\checkmark &\checkmark &\checkmark & & & 2.7662 $\pm$ 0.0683 & 11.25\\
\checkmark &\checkmark &\checkmark & &\checkmark & 2.4624 $\pm$ 0.0586 & 21.00\\
\checkmark &\checkmark &\checkmark &\checkmark & & 2.7416 $\pm$ 0.0455 & 12.04\\
\checkmark &\checkmark &\checkmark &\checkmark &\checkmark & 2.4756 $\pm$ 0.0432 & 20.58 \\
\bottomrule
\end{tabular}
    }
\end{center}
\label{table:ablation_physics}
\end{table*}

\subsection{Ablation Study: Choice of Branches}

Having evaluated the impact of physical feature selection, we now turn to the architectural design of the network itself. In this ablation study, we systematically examine different branch combinations to validate the effectiveness of the multi-branch structure. Specifically, we test all possible combinations of three candidate branches: the GCN branch, the GraphSAGE branch, and the temporal branch. Across both our proposed K-STEMIT and its non-knowledge-informed version STEMIT, the results consistently show that the two-branch spatial-temporal combination—illustrated in Figure \ref{fig:arch}—offers the most effective and balanced design.

For the three-branch case where GCN is also used as a spectral branch, we introduced two learnable scalar parameters, $\alpha$ and $\beta$, to combine the output of each branch as a weighted sum:
\begin{equation}
    \mathbf{h} = \alpha \cdot \mathbf{h}_{spatial} + \beta \cdot \mathbf{h}_{spectral} + (1-\alpha-\beta)\cdot \mathbf{h}_{temporal}
    \label{three-branch}
\end{equation}
, where $\mathbf{h}_{spatial}$ is the learned feature from the GraphSAGE branch, $\mathbf{h}_{spectral}$ is the GCN learned features and $\mathbf{h}_{temporal}$ is the output of the temporal convolution branch. For the other two-branch cases, we will still use the adaptive feature fusion with a single learnable parameter, similar to Equation~\ref{equ:adaptive}.

To stabilize the training process and allow the model to fully explore the solution space, we trained all variants under no constraints on the values of $\alpha$ or $\beta$. After training, we inspected the learned $\alpha$ and $\beta$ values. If any model assigned a negative value to a branch (i.e., $\alpha <0$ or $\beta<0$), or any model assigned a value that is greater than 1, we considered this an indication of potential instability or destructive interference from that branch. In such cases, we retrained the corresponding variant using a constrained fusion strategy, where $\alpha$ and $\beta$ are clamped to 0-1 during training. All experiments were conducted using the same training settings and hyperparameters outlined in Section~\ref{training}.

\begin{table*}[ht]
\centering
    \caption{Experiment results of ablation study on the combination of different branches for STEMIT and K-STEMIT. STEMIT denotes the non-knowledge-informed version of K-STEMIT that removes the MAR physical node features. Results are reported as the mean and standard deviation of the RMSE on the test dataset over five individual trials.}
    \scalebox{0.9}{
    \begin{tabular}{ccc}
    \toprule
\textbf{Model}                 & \textbf{Results for STEMIT} & \textbf{Results for K-STEMIT} \\ 
\midrule
GCN+GraphSAGE+TempConv (Three-branch, No Clamp)      & 3.1204 $\pm$ 0.0801 & 2.4468 $\pm$ 0.0777 \\
GCN+GraphSAGE+TempConv (Three-branch, Clamp) & 3.1089 $\pm$ 0.0569 & 2.4545 $\pm$ 0.0707\\
GCN+TempConv (Spectral + Temporal)        & 3.2461 $\pm$ 0.0529 & 2.5190 $\pm$ 0.0529 \\
GCN+GraphSAGE (Spectral + Spatial) & 3.2195 $\pm$ 0.1183 & 2.8079 $\pm$ 0.0709 \\ 
GCN (Spectral) & 5.0573 $\pm$ 0.0874 & 3.3298 $\pm$ 0.0491 \\
GraphSAGE (Spatial) & 3.1873 $\pm$ 0.0588 & 2.6173 $\pm$ 0.0835\\
TempConv (Temporal) & 3.2961 $\pm$ 0.0491 & 2.4848 $\pm$ 0.0537\\
GraphSAGE+TempConv (Spatial+Temporal, Proposed choice of branch) & \textbf{3.1170 $\pm$ 0.0780} & \textbf{2.4168 $\pm$ 0.0553}\\
\bottomrule
\end{tabular}
}
    \label{table:ablation_arch}
\end{table*}

As shown in Table~\ref{table:ablation_arch}, the multi-branch architectural configuration—consisting of a GraphSAGE spatial branch and a gated temporal convolution branch—outperforms other branch combinations for both STEMIT and K-STEMIT. While the three-branch variant (GCN + GraphSAGE + TempConv) achieves comparable accuracy with only marginal differences, the learned fusion weight for the GCN branch ($\beta$ in Equation~\ref{three-branch}) is consistently around 1\%, indicating that it contributes minimally to the overall performance. We attribute this to the fact that both GCN and GraphSAGE perform graph convolution to extract spatial features from the graph, leading to substantial redundancy between their outputs. In particular, GraphSAGE uses a more flexible neighborhood feature aggregation scheme in the spatial domain, which offers stronger generalization and allows it to capture local structural variations more effectively than the spectral-based GCN. Consequently, the GCN branch provides limited complementary information when fused with the GraphSAGE and TempConv branches. Given the additional computational resource needed for having an additional GCN branch, we consider the two-branch design (GraphSAGE + TempConv) to be the overall optimal choice for both STEMIT and K-STEMIT in terms of both efficiency and accuracy.

\subsection{Ablation Study: Contribution of Each Components in K-STEMIT}

We further conduct a comprehensive analysis to evaluate the contribution of each individual component in our proposed K-STEMIT. Specifically, we isolate the impact of the multi-branch architecture, the adaptive feature fusion strategy, and the integration of physical node features (i.e., knowledge-informed components). Our experiments include the following variants: a multi-branch network (denote as Multi-Branch) that uses the multi-branch architecture but directly concatenates the output features from each branch, representing the simplest form of our framework with only the multi-branch design; STEMIT, which combines the multi-branch architecture with the adaptive feature fusion strategy; the knowledge-informed version of the multi-branch network; and our proposed K-STEMIT, which integrates all components, including physical node features, multi-branch architecture and adaptive feature fusion strategy.

\begin{table*}[ht]
    \centering
    \caption{RMSE values and percentage improvements for different variants of K-STEMIT.}
    \scalebox{0.68}{
    \begin{tabular}{cccccc}
        \toprule
        Model & Multi-Branch Architecture & Adaptive Feature Fusion Strategy & Knowledge Informed Learning & Average RMSE & Improvement (\%) \\ \midrule
        Multi-Branch & \checkmark & & &3.0980 $\pm$ 0.0857 & 0.00 \\ 
        STEMIT& \checkmark & \checkmark & & 3.1170 $\pm$ 0.0780 &   -0.61 \\  
        Knowledge-Informed Multi-Branch & \checkmark & & \checkmark & 2.4892 $\pm$ 0.0556 &  19.65 \\ 
        K-STEMIT& \checkmark & \checkmark & \checkmark & \textbf{2.4168 $\pm$ 0.0553} &  \textbf{21.99} \\ \bottomrule
    \end{tabular}
    }
    \label{table:contribution}
\end{table*}

As shown in Table~\ref{table:contribution}, the multi-branch network serves as the foundational variant of our proposed design, incorporating only the multi-branch design. Adding the adaptive feature fusion strategy in a non-knowledge-informed scenario, as in STEMIT, results in a slight decline in performance, suggesting that feature fusion alone may not consistently yield benefits. We believe this is because, in the absence of domain knowledge, current available features are already limited in expressiveness. As a result, the adaptive feature fusion strategy, which merges outputs from different branches, may amplify redundant features shared across branches while diminishing the influence of unique features captured by each individual branch. This can lead to a less distinctive and less informative representation. Introducing physical node features into the multi-branch network results in a 19.65\% improvement in RMSE, demonstrating the value of domain knowledge. 

Notably, when applying the adaptive feature fusion strategy in the knowledge-informed setting, it provides an additional 2.34\% reduction in RMSE by effectively integrating complementary information across branches, further enhancing prediction accuracy. This can be attributed to the richer, more comprehensive feature space introduced by the physical node features, which allows distinctions to be drawn between learned spatial and temporal representations from each branch. In this setting, our proposed adaptive feature fusion strategy can learn to selectively weigh and combine the most relevant aspects of each branch, preserving each branch's distinct contributions while reducing potential feature redundancy.

\subsection{Ablation Study: Granularity of Adaptive Fusion Coefficients}

\label{sec:ablation-alpha}

As described in Section~\ref{sec:feature-fusion}, the default K-STEMIT model uses a globally shared learnable fusion coefficient $\alpha$ to balance the spatial and temporal branches:
\begin{equation}
    h_i = \alpha h_i^{s} + (1-\alpha)h_i^{t},
\end{equation}
where $h_i^{s}$ and $h_i^{t}$ denote the spatial-branch and temporal-branch hidden representations of graph node $i$, respectively. This design provides a lightweight and interpretable mechanism for learning the overall contribution of the two branches without introducing additional regional or layer-specific fusion parameters.

To investigate whether more fine-grained fusion can further improve performance, we implement three additional variants of K-STEMIT using the best-performing physical-feature combination identified in the ablation study, while keeping all other hyperparameters and experimental settings unchanged. These variants only modify the granularity of the fusion coefficient.

First, we consider a regional adaptive fusion variant, where the globally shared coefficient is replaced by a node-level coefficient $\alpha_i$: 
\begin{equation}
    h_i = \alpha_i h_i^{s} + (1-\alpha_i)h_i^{t}.
\end{equation}

Here, $i$ indexes a graph node or local spatial region. In our implementation, $\alpha_i$ is generated by a small gating network:
\begin{equation}
    \alpha_i = \sigma \left(g_{\theta}\left([h_i^{s}, h_i^{t}, p_i]\right)\right),
\end{equation}

where $p_i$ denotes the coordinate features of node $i$, $g_{\theta}(\cdot)$ is a multilayer perceptron, and $\sigma(\cdot)$ is the sigmoid function. This design allows different spatial locations to assign different weights to the spatial and temporal branches.

Second, we evaluate a layer-wise adaptive fusion variant, where one fusion coefficient is learned for each predicted target ice layer:
\begin{equation}
    \hat{y}_{i,\ell}
=
\alpha_{\ell}\hat{y}^{s}_{i,\ell}
+
(1-\alpha_{\ell})\hat{y}^{t}_{i,\ell},
\end{equation}
where $\ell \in \{1,\dots,n\}$ denotes the index of a predicted target layer. In this variant, the spatial and temporal branch representations are decoded separately, and the final prediction is fused using a learnable vector of layer-wise coefficients. This allows different target depths to use different spatial-temporal fusion weights.

Third, we evaluate a joint regional-layer adaptive fusion variant:
\begin{equation}
    \hat{y}_{i,\ell}
=
\alpha_{i,\ell}\hat{y}^{s}_{i,\ell}
+
(1-\alpha_{i,\ell})\hat{y}^{t}_{i,\ell}.
\end{equation}
In this case, the fusion coefficient varies across both graph nodes and predicted target layers. This is the most flexible variant, since the fusion behavior can depend jointly on spatial location and target-layer depth.

The results are summarized in Table~\ref{tab:response-adaptive-alpha}. Under the current SRED benchmark setting, all three fine-grained fusion variants perform worse than the globally shared $\alpha$ used in the default K-STEMIT model. This suggests that, for the current single-benchmark Greenland dataset, the additional flexibility introduced by regional or layer-wise fusion does not improve prediction accuracy and may increase the risk of overfitting. Therefore, we retain the globally shared learnable coefficient $\alpha$ in the final K-STEMIT model.

\begin{table}[!t]
\centering
\caption{Additional comparison of different fusion granularities under the same experimental setting on the SRED Greenland benchmark. The globally shared $\alpha$ is the fusion strategy used in the final K-STEMIT model.}
\label{tab:response-adaptive-alpha}
\resizebox{\linewidth}{!}{
\begin{tabular}{lcccc}
\toprule
\textbf{Fusion Strategy} & \textbf{Fusion Coefficient} & \textbf{RMSE} & \textbf{MAE} & \textbf{Time} \\
\midrule
Global adaptive fusion, used in K-STEMIT 
& $\alpha$ 
& \textbf{2.4168 $\pm$ 0.0553} 
& \textbf{1.5714 $\pm$ 0.0328} 
& \textbf{1016 sec} \\

Regional adaptive fusion 
& $\alpha_i$ 
& 2.6208 $\pm$ 0.0672 
& 1.7414 $\pm$ 0.0442 
& 1140 Sec \\

Depth-dependent fusion 
& $\alpha_\ell$ 
& 2.4528 $\pm$ 0.0570 
& 1.5883 $\pm$ 0.0307 
& 1068 Sec \\

Joint regional-depth fusion 
& $\alpha_{i,\ell}$ 
& 2.6324 $\pm$ 0.0782 
& 1.7373 $\pm$ 0.0466
& 1090 Sec\\
\bottomrule
\end{tabular}
}
\end{table}

\section{Discussion}\label{discussion}

\subsection{Statistical Significance Analysis}
\label{stat_test}

To further assess whether the performance gains of K-STEMIT reflect a stable effect rather than run-to-run fluctuation, we conducted paired significance tests using the results from five matched runs under the same data split and experimental protocol. We considered four targeted comparisons: (1) K-STEMIT versus Knowledge-Informed Multi-Branch, to isolate the effect of the proposed adaptive fusion within the same physics-conditioned multi-branch design; (2) K-STEMIT versus GAT with physical node features, which is a strong knowledge-informed graph baseline in Table~\ref{table:OverallResults}; (3) K-STEMIT versus Physics-Informed GraphSAGE-LSTM, which is another strong prior physics-informed spatio-temporal graph baseline; and (4) K-STEMIT versus ST-GRIT, which is the current state-of-the-art non-physical model. For each comparison, we evaluated both RMSE and MAE using two-sided paired $t$-tests. In addition, we report two-sided Wilcoxon signed-rank tests~\cite{conover1999practical} as a supplementary nonparametric check.

For each run, the paired difference was defined as $\Delta = \mathrm{Error}_{\mathrm{OtherModel}} - \mathrm{Error}_{\mathrm{K\text{-}STEMIT}}$. Under this definition, a positive difference means that K-STEMIT has lower error than the compared other model. For consistency of interpretation, the paired $t$-tests were conducted in the order (baseline, K-STEMIT), so that a positive $t$-statistic also indicates lower error for K-STEMIT.

The paired $t$-test evaluates whether the mean paired difference is significantly different from zero. Its test statistic is the average paired difference divided by the estimated standard error of that difference. Therefore, a larger positive $t$-value indicates that the improvement of K-STEMIT is not only positive on average, but also large relative to the run-to-run variability across matched trials. The corresponding $p$-value quantifies how likely it would be, under the null hypothesis of no true mean difference, to observe a result at least as extreme as the one obtained. Thus, a small $p$-value provides evidence that the average improvement is unlikely to be explained by random variation alone.

Table~\ref{tab:per_run_stats} reports the per-run RMSE and MAE values used in the paired analysis. As shown, K-STEMIT achieves lower RMSE and MAE than every comparison model in all five matched runs. The mean paired RMSE / MAE reductions are $0.0724/0.0460$ relative to Knowledge-Informed Multi-Branch, $0.1697/0.1416$ relative to physical-feature GAT, $0.4935/0.4627$ relative to Physics-Informed GraphSAGE-LSTM, and $0.4698/0.3427$ relative to ST-GRIT. Because all paired differences are positive, the observed advantage of K-STEMIT is not driven by only one or two favorable runs, but is directionally consistent across the entire set of matched trials.

Table~\ref{tab:paired_stats} summarizes the formal test results. The paired $t$-test yields statistically significant improvements for K-STEMIT in all comparisons at the $0.05$ level. In particular, the gains are highly significant against physical-feature GAT (RMSE: $p=7.74\times10^{-6}$; MAE: $p=3.36\times10^{-5}$), Physics-Informed GraphSAGE-LSTM (RMSE: $p=4.73\times10^{-5}$; MAE: $p=2.29\times10^{-5}$), and ST-GRIT (RMSE: $p=1.70\times10^{-5}$; MAE: $p=9.29\times10^{-6}$). Even against the strongest internal variant, Knowledge-Informed Multi-Branch, the reduction remains statistically significant (RMSE: $p=0.0230$; MAE: $p=3.69\times10^{-4}$), indicating that the adaptive fusion mechanism contributes a measurable and repeatable improvement beyond the non-adaptive fusion design.

The Wilcoxon signed-rank test serves a complementary purpose. Unlike the paired $t$-test, it does not focus on the mean difference; instead, it evaluates whether the paired differences are consistently shifted away from zero based on their signed ranks, and is less sensitive to distributional assumptions. In all comparisons and for both RMSE and MAE, the Wilcoxon test returns $W=0$ and $p=0.0625$. Here, $W=0$ means that all nonzero paired differences have the same sign, i.e., every matched run favors K-STEMIT. The fact that the two-sided Wilcoxon $p$-value is slightly above $0.05$ should be interpreted with caution: with only five paired runs, the exact two-sided Wilcoxon test is highly conservative, and $p=0.0625$ is the smallest attainable two-sided $p$-value when all five paired differences are in the same direction. Therefore, the Wilcoxon results should not be read as contradictory to the paired $t$-test; rather, they confirm perfect directional consistency, while the paired $t$-test shows that the average improvement is large relative to run-to-run variability.

Overall, the two tests provide complementary evidence. The paired $t$-test demonstrates that the magnitude of the average improvement of K-STEMIT is statistically reliable, whereas the Wilcoxon signed-rank test confirms that this improvement is consistently observed across all five matched runs. Taken together, these results provide strong evidence that the gains of K-STEMIT are robust and not an artifact of random initialization or run-to-run noise.

\begin{table*}[t]
\centering
\caption{Per-run RMSE and MAE values used in the paired statistical tests. KI denotes knowledge-informed and PI denotes physics-informed.}
\label{tab:per_run_stats}
\setlength{\tabcolsep}{4pt}
\resizebox{\textwidth}{!}{
\begin{tabular}{c cc cc cc cc cc}
\toprule
\multirow{2}{*}{Run} 
& \multicolumn{2}{c}{K-STEMIT} 
& \multicolumn{2}{c}{KI Multi-Branch} 
& \multicolumn{2}{c}{GAT + Physical Features} 
& \multicolumn{2}{c}{PI-GraphSAGE-LSTM} 
& \multicolumn{2}{c}{ST-GRIT} \\
\cmidrule(lr){2-3}
\cmidrule(lr){4-5}
\cmidrule(lr){6-7}
\cmidrule(lr){8-9}
\cmidrule(lr){10-11}
& RMSE & MAE 
& RMSE & MAE 
& RMSE & MAE 
& RMSE & MAE 
& RMSE & MAE \\
\midrule
1 & 2.4028 & 1.5448 & 2.4430 & 1.5899 & 2.5764 & 1.7099 & 2.7972 & 1.9495 & 2.8252 & 1.9167 \\
2 & 2.4168 & 1.5958 & 2.4924 & 1.6399 & 2.5732 & 1.7177 & 2.9519 & 2.0500 & 2.9153 & 1.9253 \\
3 & 2.5214 & 1.6237 & 2.5700 & 1.6619 & 2.7110 & 1.7663 & 3.0394 & 2.1365 & 2.9683 & 1.9468 \\
4 & 2.3671 & 1.5399 & 2.4151 & 1.5808 & 2.5289 & 1.6775 & 2.8534 & 1.9766 & 2.8190 & 1.8568 \\
5 & 2.3760 & 1.5530 & 2.5254 & 1.6148 & 2.5432 & 1.6940 & 2.9095 & 2.0581 & 2.9054 & 1.9250 \\
\bottomrule
\end{tabular}
}
\end{table*}

\begin{table*}[t]
\centering
\caption{Paired statistical comparison of K-STEMIT against representative baselines over five matched runs. Positive mean difference indicates lower error for K-STEMIT, computed as baseline error minus K-STEMIT error. Because the paired $t$-tests were conducted in the order (baseline, K-STEMIT), a positive $t$-statistic also indicates lower error for K-STEMIT.}
\label{tab:paired_stats}
\setlength{\tabcolsep}{5pt}
\begin{tabular}{lcccccc}
\toprule
Comparison & Metric & Mean Diff. & $t$ & $p$ ($t$-test) & $W$ & $p$ (Wilcoxon) \\
\midrule
K-STEMIT vs KI Multi-Branch & RMSE & 0.0724 & 3.5872 & 0.0230 & 0.0 & 0.0625 \\
K-STEMIT vs KI Multi-Branch & MAE  & 0.0460 & 11.1466 & $3.69\times10^{-4}$ & 0.0 & 0.0625 \\
K-STEMIT vs GAT + physical features & RMSE & 0.1697 & 29.6171 & $7.74\times10^{-6}$ & 0.0 & 0.0625 \\
K-STEMIT vs GAT + physical features & MAE  & 0.1416 & 20.4686 & $3.36\times10^{-5}$ & 0.0 & 0.0625 \\
K-STEMIT vs PI-GraphSAGE-LSTM & RMSE & 0.4935 & 18.7834 & $4.73\times10^{-5}$ & 0.0 & 0.0625 \\
K-STEMIT vs PI-GraphSAGE-LSTM & MAE  & 0.4627 & 22.5506 & $2.29\times10^{-5}$ & 0.0 & 0.0625 \\
K-STEMIT vs ST-GRIT & RMSE & 0.4698 & 24.3143 & $1.70\times10^{-5}$ & 0.0 & 0.0625 \\
K-STEMIT vs ST-GRIT & MAE  & 0.3427 & 28.2870 & $9.29\times10^{-6}$ & 0.0 & 0.0625 \\
\bottomrule
\end{tabular}
\end{table*}

\subsection{Visualization of Optimization and Prediction Characteristics}
\label{app:supp_vis}

\begin{figure}
\centering
\includegraphics[width=0.8\textwidth]{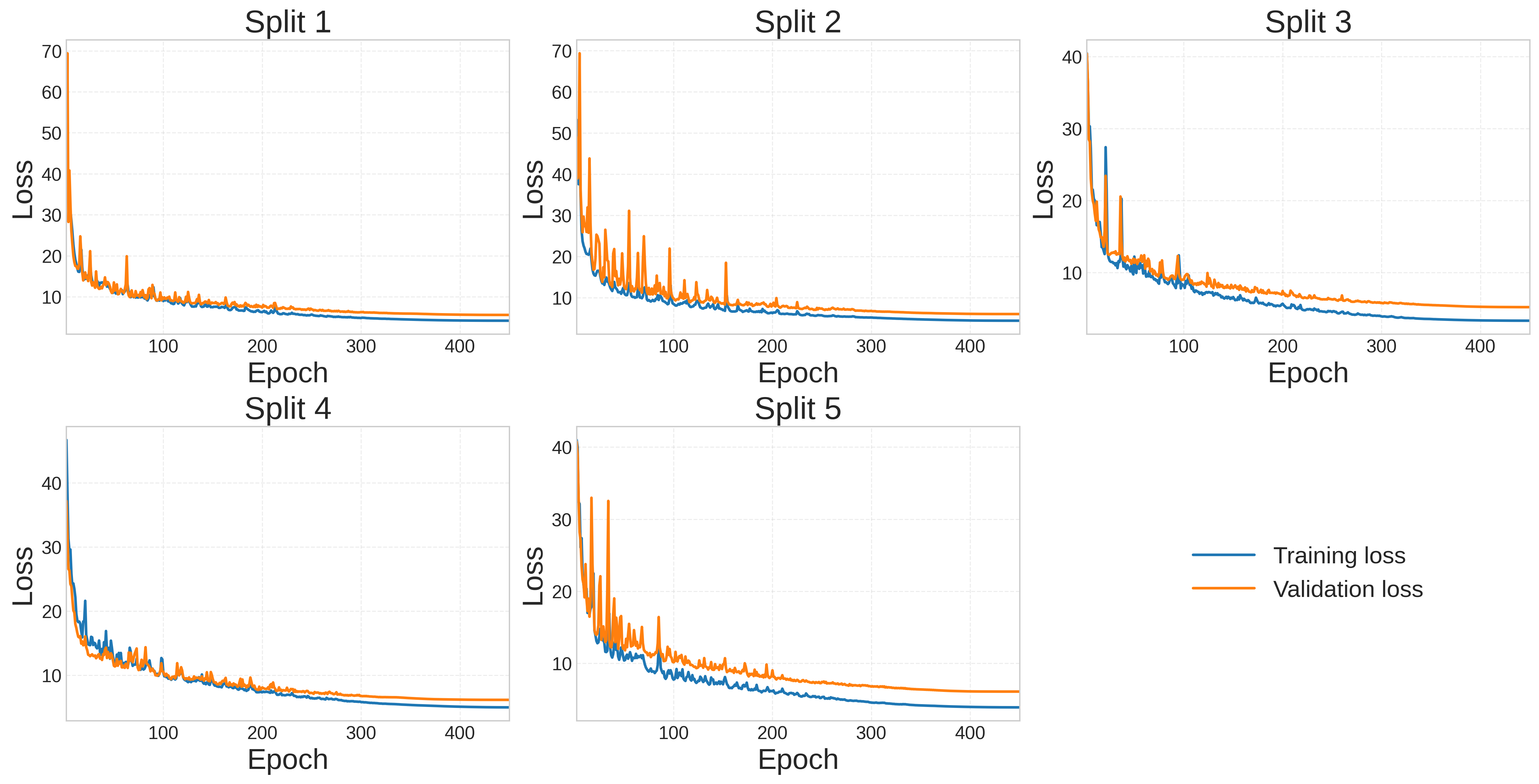}
\caption{Training and validation loss curves of K-STEMIT over all five data splits. Across splits, the losses show overall downward trends with only minor transient fluctuations, indicating stable convergence behavior during optimization.\label{fig:learning_curves}}
\end{figure}  

\begin{figure}
\centering
\includegraphics[width=0.8\textwidth]{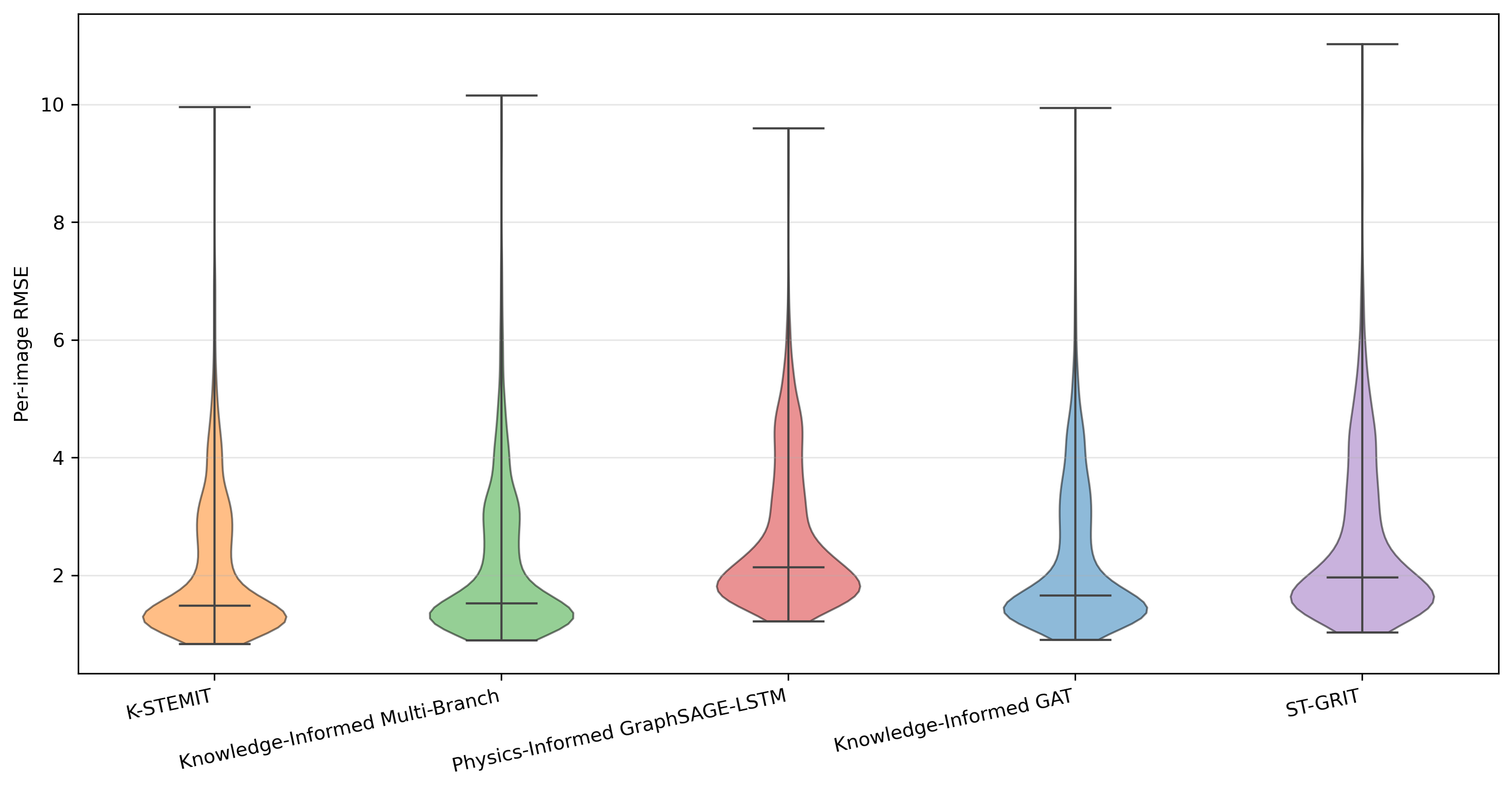}
\caption{Violin plots of per-image RMSE distributions across five runs for K-STEMIT, Multi-Branch Graph Neural Network, and Physics-Informed GraphSAGE-LSTM. The figure provides a distributional view of prediction errors beyond the averaged RMSE values reported in the main text.\label{fig:violin_rmse}}
\end{figure}  

\begin{figure}
\centering
\includegraphics[width=0.8\textwidth]{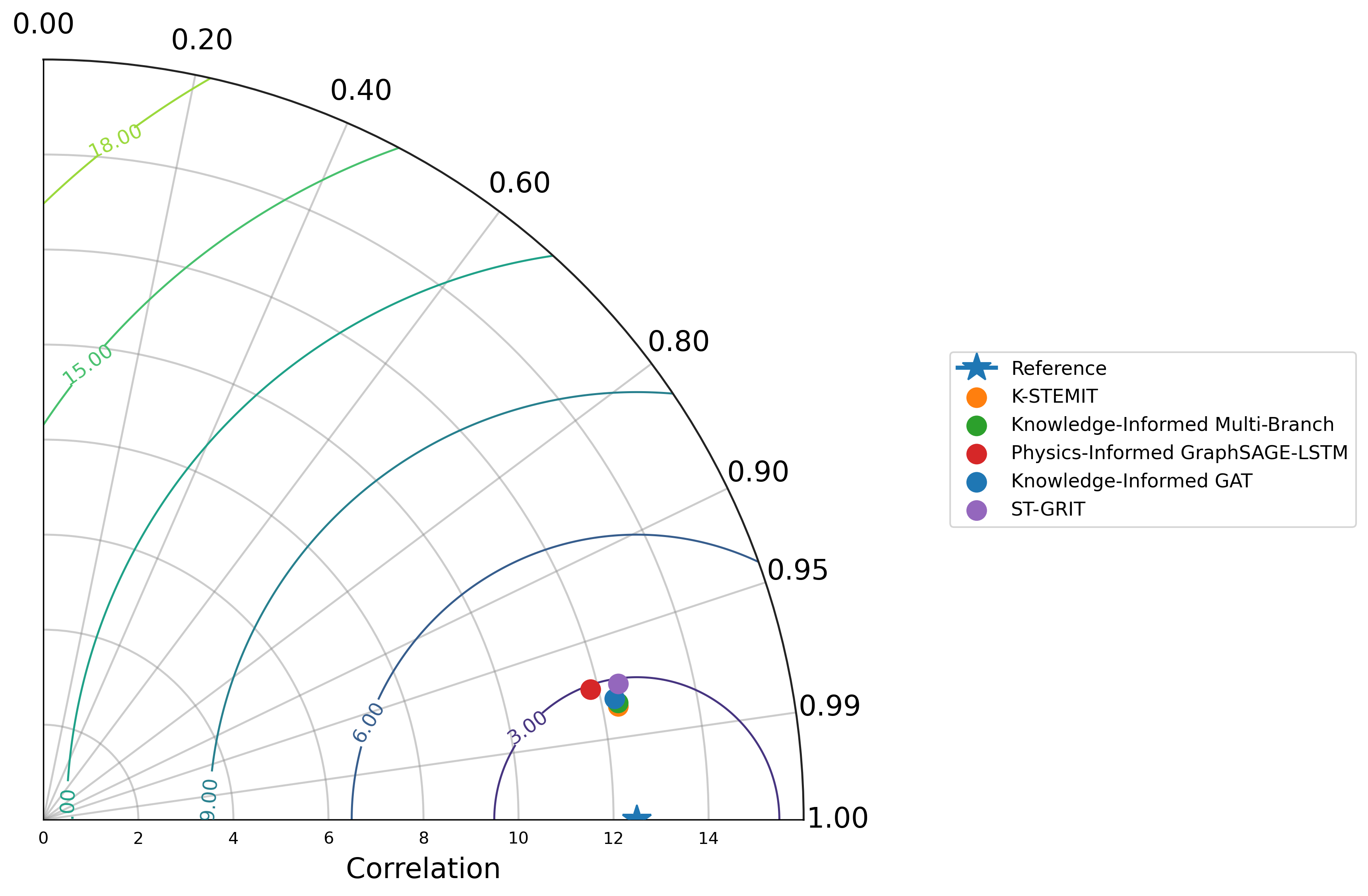}
\caption{Taylor diagram summarizing the agreement between model predictions and ground truth across five runs for K-STEMIT, Multi-Branch Graph Neural Network, and Physics-Informed GraphSAGE-LSTM in terms of correlation, standard deviation, and centered RMSE. Points closer to the reference point indicate better overall agreement.\label{fig:taylor_diagram}}
\end{figure}  

To complement the main quantitative results, we provide additional supplementary visualizations of optimization behavior and prediction characteristics. Specifically, we include training and validation loss curves of K-STEMIT across all five data splits, violin plots of per-image RMSE distributions for K-STEMIT and four competitive baselines, and a Taylor diagram comparing the same set of methods. Together, these figures offer additional diagnostic views of convergence stability, error dispersion, and overall agreement with the ground truth in terms of correlation, standard deviation, and centered RMSE. 

Figure~\ref{fig:learning_curves} presents the training and validation loss curves of K-STEMIT over all five data splits. Across all splits, both training and validation losses decrease substantially as training proceeds, indicating stable convergence behavior. Although several splits exhibit occasional fluctuations or transient spikes, especially in earlier epochs, the overall trends remain steadily downward and no late-stage divergence is observed. This behavior supports the robustness of the optimization process across different data splits.

Figure~\ref{fig:violin_rmse} shows violin plots of per-image RMSE distributions aggregated across the five evaluation splits for K-STEMIT, Knowledge-Informed Multi-Branch, Physics-Informed GraphSAGE-LSTM, Knowledge-Informed GAT, and ST-GRIT. Among these methods, K-STEMIT exhibits the lowest concentration of errors in the central low-RMSE region and one of the most compact overall distributions. In contrast, other four methods generally show higher central error levels and broader upper tails, indicating that larger prediction errors occur more frequently on difficult samples. This distributional view is consistent with the overall quantitative comparison reported before.

Figure~\ref{fig:taylor_diagram} presents a Taylor diagram summarizing the agreement between model predictions and the ground truth across the evaluated methods. The diagram jointly characterizes correlation, standard deviation, and centered RMSE. All methods achieve high correlation with the reference, but they differ in how closely they match the reference variability and how much centered error remains. K-STEMIT is located among the points closest to the reference and within the low-centered-RMSE region, indicating a favorable balance among correlation, variability matching, and centered error. This observation is consistent with the quantitative results reported before.

\subsection{Qualitative Comparison With Other Models} \label{comparison}
\begin{figure}
\centering
\includegraphics[width=0.9\textwidth]{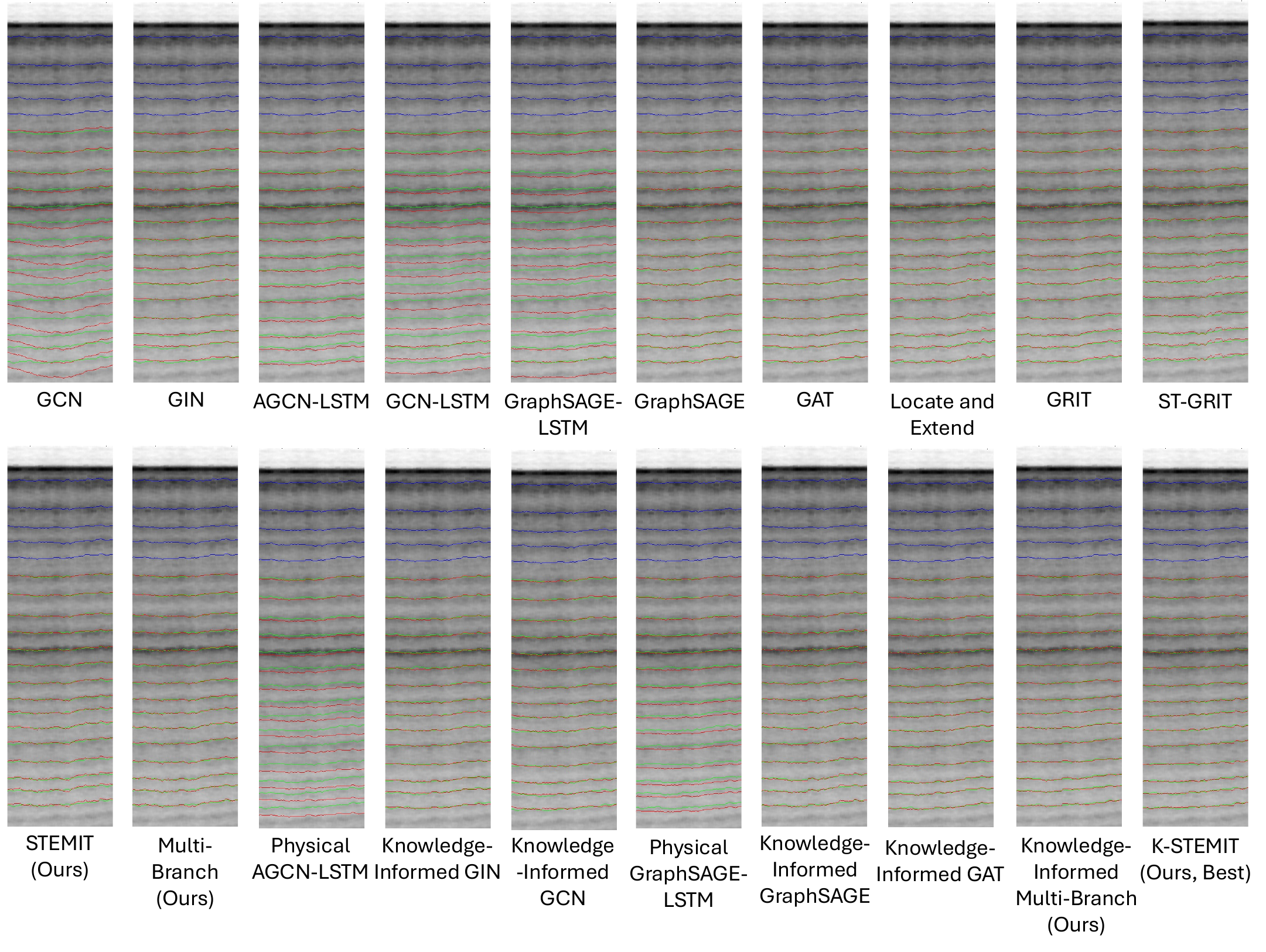}
\caption{Comparison between qualitative results of different graph models. The blue line is used to generate the graphs. The green line is the groundtruth (manually-labeled ice layers) and the red line is the model prediction.\label{fig:qualitative}}
\end{figure}

In order to gain a better understanding of how our proposed K-STEMIT outperforms current state-of-the-art models and other existing methods, we create qualitative samples of each model prediction on the same radargram image, shown in Figure \ref{fig:qualitative}. We observe that while all the networks achieve decent accuracy in predicting ice layer thickness, those non-knowledge-informed fused spatio-temporal graph neural networks exhibit significant error accumulation. This accumulation leads to noticeable shifts in the visualization of deeper ice layers on the radargram, as the coordinates of deeper ice layer boundaries are calculated cumulatively by adding the predicted thickness values from the top to the bottom. The figure illustrates that, in comparison to previous models, the multi-branch architecture significantly improves the predictions for deeper ice layers and the predictions of the boundary regions of the radargram. This results in more precise and reliable visualizations of the ice layers. Additionally, by comparing Figure~\ref{fig:qualitative} (d) and (e), we observe that integrating physical features as domain-specific knowledge further enhances the performance, particularly in the left and right boundary regions of each radargram. This suggests that the inclusion of physical features enables the model to better capture fine-grained, pixel-level local patterns.

\begin{figure}
    {
        \includegraphics[width=\textwidth]{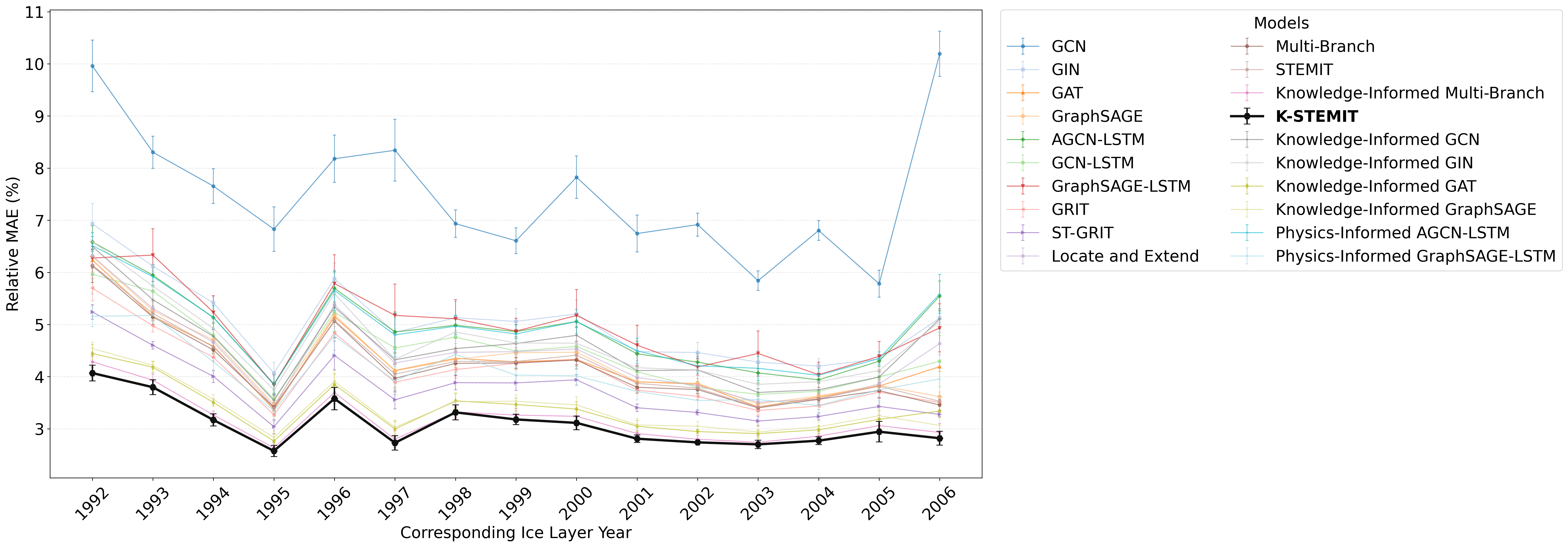}
    }
    \caption{Relative MAE on individual ice layers for each model.}
    \label{fig:relative_mae}
\end{figure}

\begin{figure}
    {
        \includegraphics[width=0.8\textwidth]{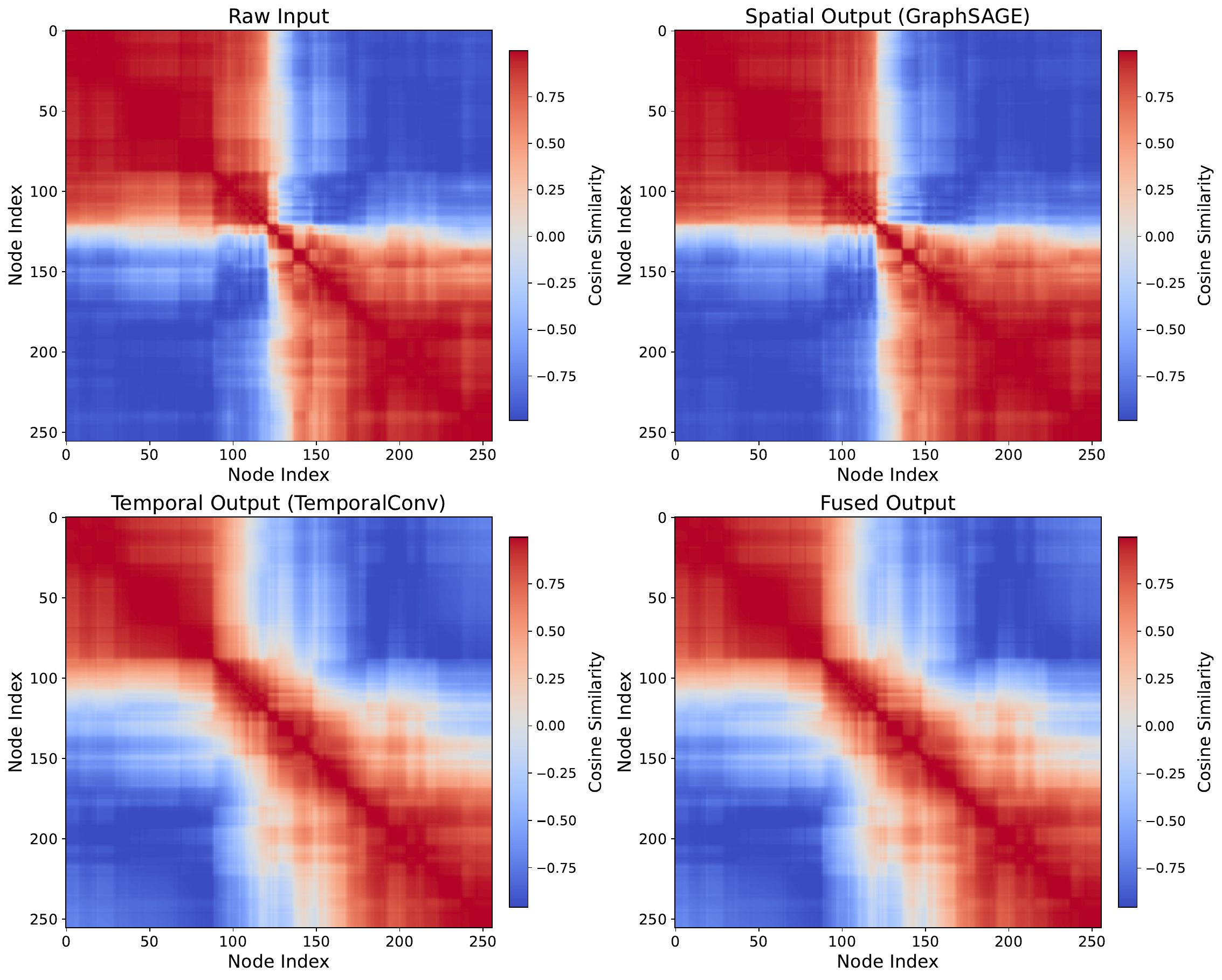}
    }
    \caption{Centered cosine-similarity heatmaps of node representations from the raw input, GraphSAGE spatial branch, temporal convolution branch, and fused representation for a representative held-out sample. Each row/column corresponds to one radargram column node.}
    \label{fig:cosine_similarity}
\end{figure}

\begin{figure}
    {
        \includegraphics[width=0.8\textwidth]{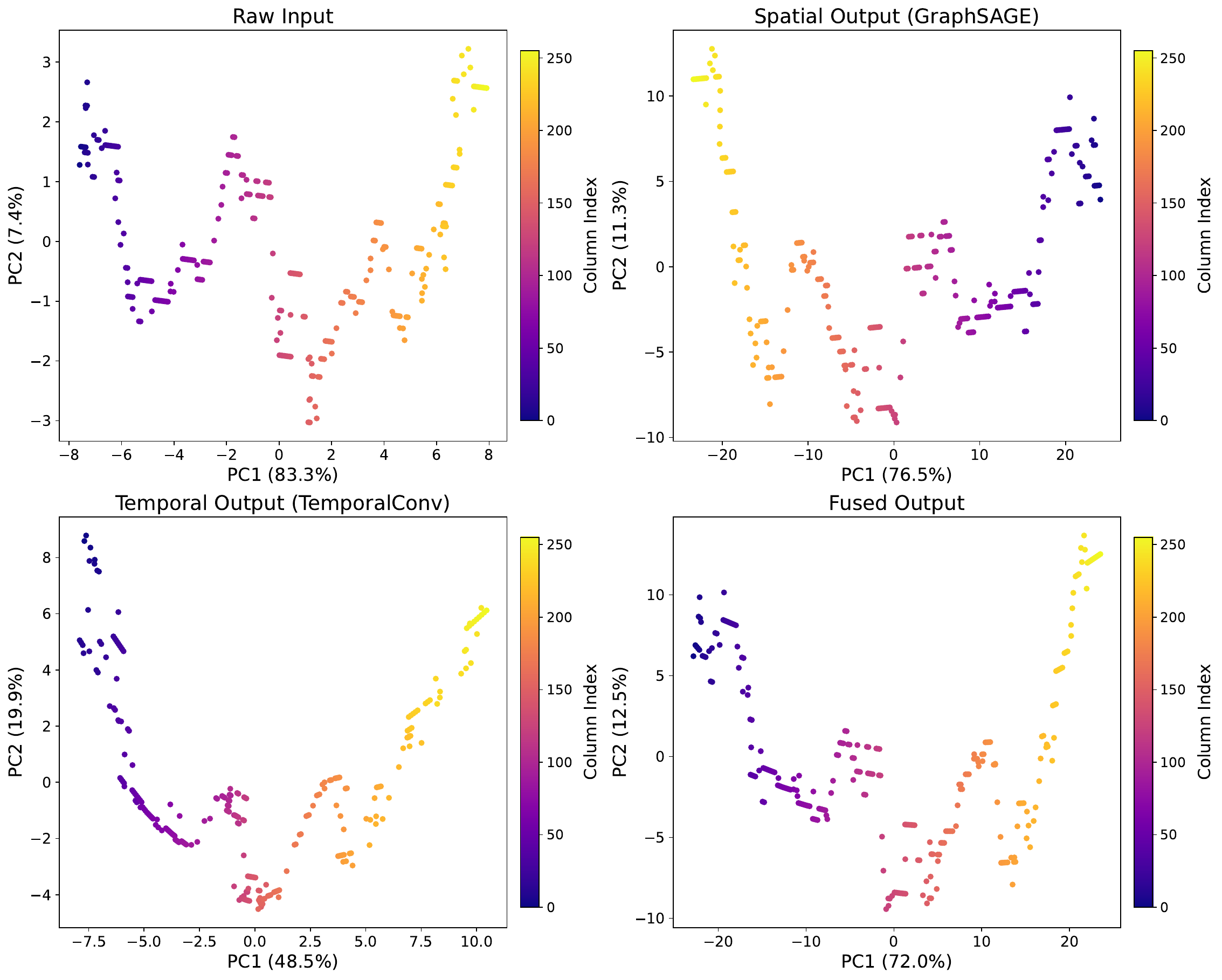}
    }
    \caption{PCA visualization of node representations colored by radargram column index for the same held-out sample. Each point denotes one node, which is one fixed radargram column.}
    \label{fig:pca_column}
\end{figure}

\begin{figure}
    {
        \includegraphics[width=0.8\textwidth]{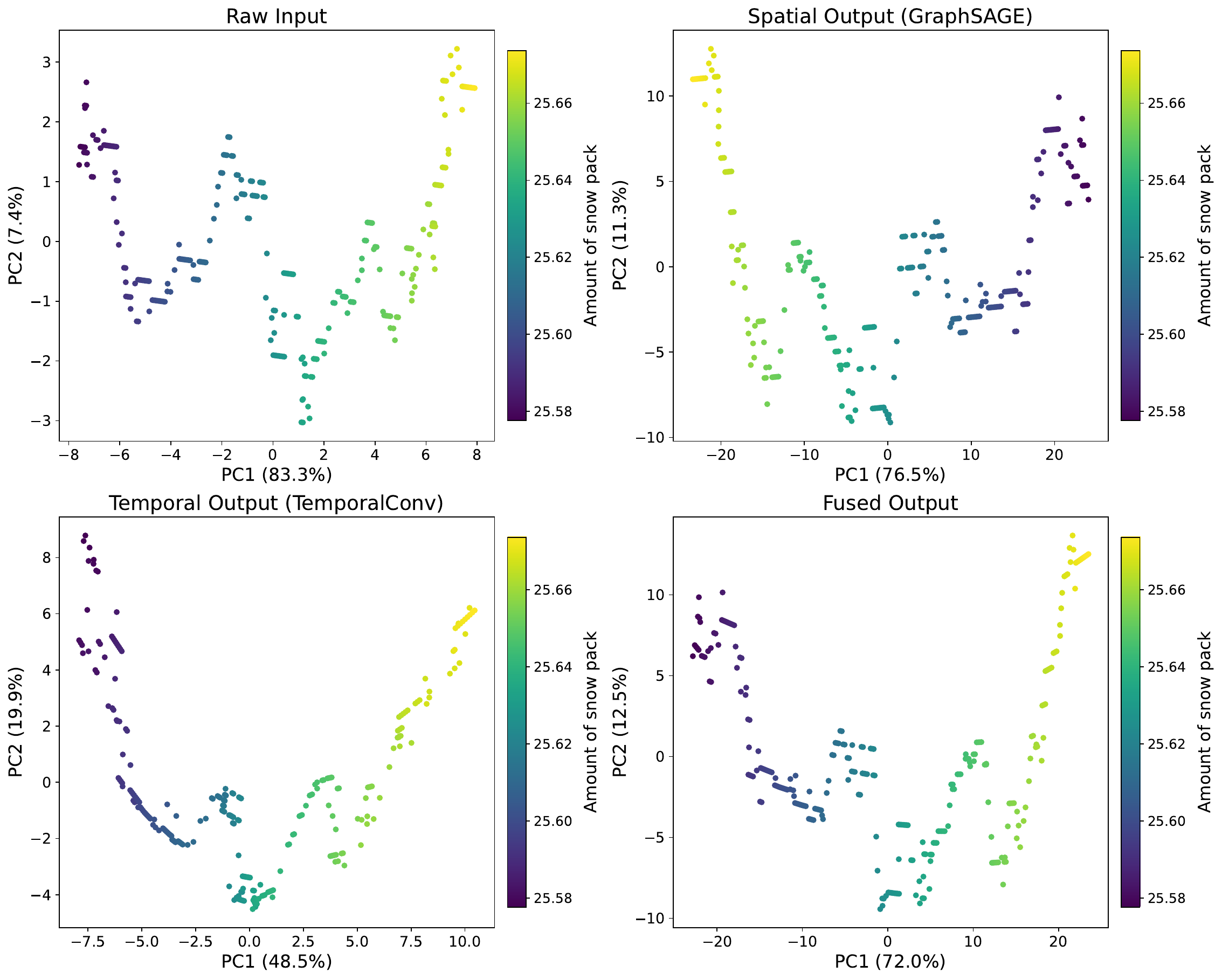}
    }
    \caption{PCA visualization of node representations colored by amount of snow pack for the same held-out sample.}
    \label{fig:pca_snow_pack}
\end{figure}

\begin{figure}
    {
        \includegraphics[width=0.8\textwidth]{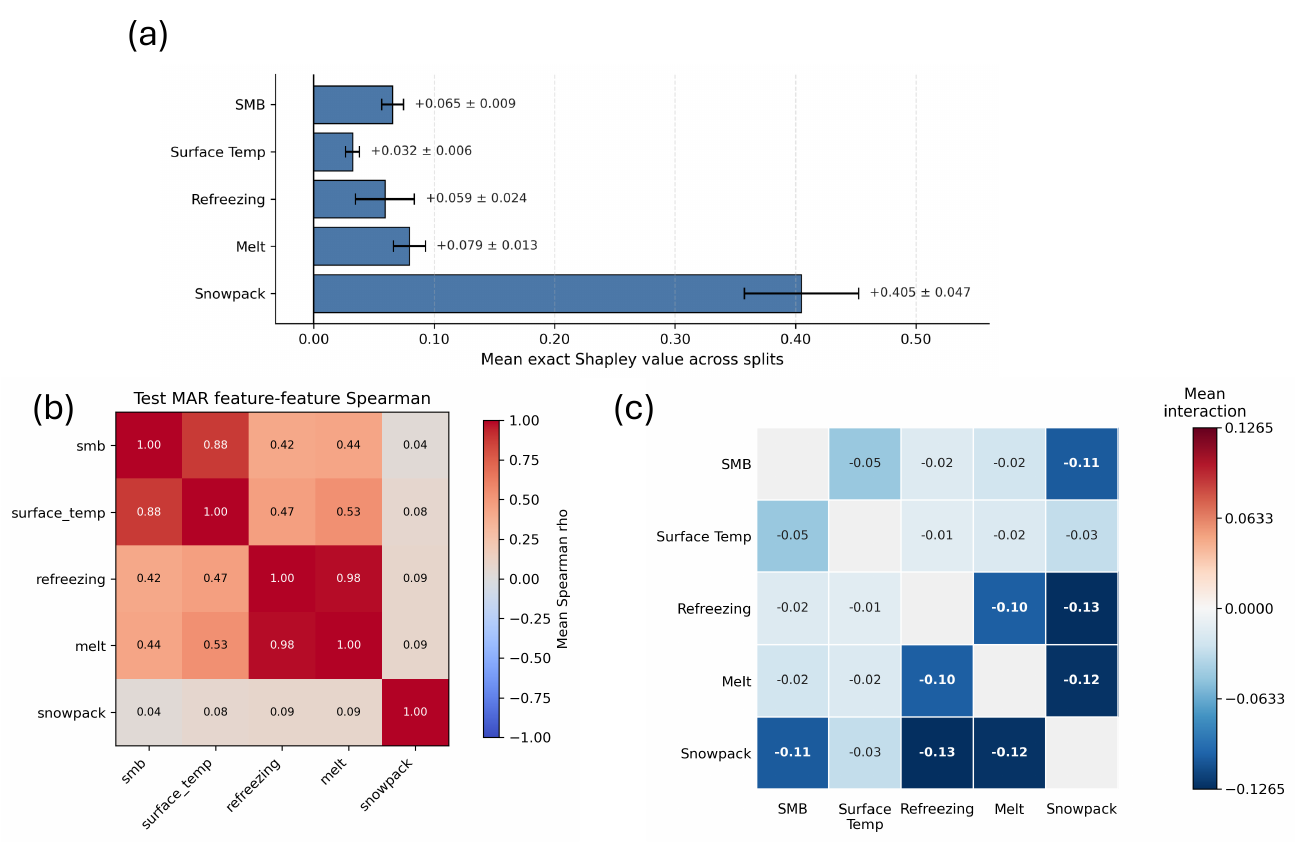}
    }
    \caption{(\textbf{a}) Exact Shapley values computed from the complete 32 ablation table and averaged across the five test splits. Larger positive values indicate greater average reduction in prediction error relative to the empty-feature baseline. (\textbf{b}) Mean test-set feature-feature Spearman correlation matrix among the five MAR variables, showing redundancy in the raw physical inputs. (\textbf{c}) Mean pairwise interaction scores derived from the same coalition-value table used for the Shapley analysis; positive values indicate synergistic interactions and negative values indicate sub-additive or redundant interactions. Taken together, these panels distinguish marginal association, input redundancy, and conditional predictive contribution.}
    \label{fig:physics_results_analysis}
\end{figure}

\subsection{Performance For Each Individual Ice Layer}
To further evaluate the prediction performance of our proposed K-STEMIT, we compute the relative mean absolute error (relative MAE) for each of the $n$ predicted ice layers ($n=15$ for this paper). This metric allows us to assess how the model performs at different depths.

In previous sections, we denote \( \mathbf{y}_i^{(k)} \in \mathbb{R}^{256 \times n} \) and \( \hat{\mathbf{y}}_i^{(k)} \in \mathbb{R}^{256 \times n} \) as the ground truth and predicted outputs for the $i$-th test radargram sample in version $k$ test dataset (where $k$ refers to the $k$-th permutation of the dataset used to generate distinct train, validation, and test sets; see Section~\ref{training} for details). Here, we further define $y_{i,v,j}^{(k)}$ and $\hat{y}_{i,v,j}^{(k)}$ as the ground truth and predicted thickness values at node $v \in \{1, \dots, 256\}$ and layer $j \in \{1, \dots, n\}$, respectively. The relative MAE for the $j$-th predicted layer is computed as:

\begin{equation}
\delta_j^{(k)} = \frac{1}{N_k \cdot 256} \sum_{i=1}^{N_k} \sum_{v=1}^{256}
\left| \frac{ \hat{y}_{i,v,j}^{(k)} - y_{i,v,j}^{(k)} }{ y_{i,v,j}^{(k)} } \right|,
\end{equation}

where $N_k$ is the number of radargrams in version $k$ test dataset, and $\delta_j^{(k)}$ represents the relative error for layer $j$ in that split. Unlike RMSE, where larger errors from outliers disproportionately affect the total error, relative MAE treats all error terms equally by normalizing each residual by its ground truth value. Evaluating $\delta_j^{(k)}$ across layers provides insights into the error consistency of the model, revealing whether deeper layers tend to be harder to predict than shallower ones. To report the final results, we take the mean and standard deviation of $\delta_j^{(k)}$ across the five data splits for each layer $j$, defined as:
\begin{equation}
\bar{\delta}_j = \frac{1}{5} \sum_{k=1}^{5} \delta_j^{(k)}, \quad
\mathrm{Std}_{\delta_j} = \sqrt{ \frac{1}{5} \sum_{k=1}^{5} \left( \delta_j^{(k)} - \bar{\delta}_j \right)^2 }.
\end{equation}

As shown in Figure \ref{fig:relative_mae}, we can find an overall negative slope. This trend indicates that overall, shallow layers' physical structure and thickness are easier to predict. This is a reasonable result, as in the radargram, deep layers are typically less contrasty due to the low signal-to-noise ratio, making them more difficult to distinguish. Compared with current state-of-the-art methods and other existing methods, our proposed K-STEMIT consistently achieves lower, more stable relative MAE across different ice layers. This result supports the conclusion that incorporating domain knowledge enhances both the accuracy and robustness of deep ice layer thickness prediction.

\subsection{Branch-level Interpretability Analysis of Learned Node Representations}

To better understand what the proposed branches capture, we perform a targeted visualization-based interpretability analysis on a representative held-out sample. Since in our graph data, each node corresponds to a fixed radargram column, node embeddings can be interpreted as column-wise latent representations. For the raw input $\textbf{X}$, we construct a node descriptor by concatenating the selected node features across the five observed upper layers, yielding one raw representation per column. We then extract the corresponding node embeddings from the GraphSAGE spatial branch ($\textbf{h}_{\text{spatial}}$), the temporal convolution branch ($\textbf{h}_{\text{temporal}}$), and the fused representation after branch integration ($\textbf{h}$).

We analyze these representations from two complementary perspectives. First, we compute centered cosine-similarity heatmaps to characterize pairwise relationships among nodes. Specifically, for each representation matrix, we subtract the feature-wise mean before computing cosine similarity, which suppresses the influence of a dominant common component and reveals the relative structural organization more clearly. Second, we project the node representations into two dimensions using PCA and visualize the resulting embeddings by coloring nodes according to either radargram column index or a representative physical variable. Here, the amount of snow pack is used as the representative physical variable, enabling us to examine whether the learned embeddings reflect physically meaningful environmental variation.

The resulting visualizations reveal distinct branch behaviors. As shown in Figure~\ref{fig:cosine_similarity}, the raw input and GraphSAGE spatial branch show similar coarse block structure in the similarity maps, indicating that the spatial branch mainly preserves broad structural grouping and localized neighborhood relationships already present in the input. By contrast, the temporal branch produces a more diagonal and locally smooth similarity pattern, suggesting that it captures progression-like cross-layer dependencies along the radargram columns. The fused representation combines these effects. It retains the smoother local organization induced by the temporal branch while preserving part of the broader global structure inherited from the spatial branch.

The PCA visualizations provide a consistent interpretation. As shown in Figure~\ref{fig:pca_column}, when colored by column index, the raw, temporal, and fused representations show a smooth ordering with respect to column index, while the GraphSAGE branch remains structured but more strongly reflects coarse grouping. This indicates that the learned latent spaces preserve the natural column-wise progression of the radargram, with the temporal branch yielding the clearest progression-like geometry. When colored by amount of snow pack, as shown in Figure~\ref{fig:pca_snow_pack}, the latent spaces also show a smooth physical gradient, suggesting that the learned node embeddings are associated with a meaningful environmental property rather than reflecting only geometric position. Taken together, these results provide qualitative evidence that the spatial and temporal branches learn complementary information and that the resulting node embeddings encode both structural and physically relevant cues.

\subsection{Interpretation of MAR-derived Physical Feature Effects} \label{interpretation}
The result in Table \ref{table:OverallResults} confirms that the inclusion of physical features from MAR improves the overall model's performance; however, the ablation study in Table \ref{table:ablation_physics} indicates that the best accuracy is achieved by only using three out of five available features. This may be attributed to the potential redundancy and the indirect nature of relationships between these five radargram internal snow layer features. To further interpret the role of MAR-derived physical variables, we complemented the exhaustive 32 ablation study with four quantitative analyses: feature-thickness correlation, exact Shapley feature attribution~\cite{Shapley,molnar2025_Shapley}, MAR feature-feature correlation, and pairwise interaction analysis~\cite{Grabisch1999}. These analyses address different aspects of the physical features. The feature-thickness correlation quantifies the marginal association between each physical variable and the target thickness. Exact Shapley attribution quantifies the conditional predictive contribution of each variable after averaging over all possible feature-subset contexts~\cite{Shapley,molnar2025_Shapley}. The MAR feature–feature correlation matrix measures redundancy among the physical input variables themselves. Finally, the pairwise interaction analysis evaluates whether pairs of variables provide synergistic or sub-additive predictive effects at the model-performance level~\cite{Grabisch1999}.

For the feature–thickness correlation analysis, we used the aligned radargram-column structure of the dataset. Each radargram contains five observed upper layers, and each layer has 256 graph nodes corresponding to the same horizontal radargram columns. For each MAR physical variable, we averaged its values across the five observed upper layers at each column. This produces one value for each physical feature at each of the 256 columns of each radargram. The target output contains 15 thickness values at each of the same 256 columns; these were summarized into a single node-level target-thickness value. We then computed Spearman correlation~\cite{Spearman_Corr} between each physical feature and the corresponding thickness value using two strategies. First, the pooled Spearman correlation is computed by combining all radargram-column pairs in the test split into one vector before computing the correlation. Second, the within-radargram Spearman correlation is computed separately inside each radargram using only its 256 columns, and the resulting per-radargram correlations are then averaged across radargrams and splits.

\begin{table}[t]
\centering
\caption{Summary of feature-thickness correlation analysis on the five test splits. “Pooled” Spearman correlation is computed over all aligned $(\text{radargram}, \text{node})$ pairs after averaging each MAR feature across the five observed upper layers and summarizing the 15-layer target into one node-level thickness scalar. “Within-radargram” Spearman reports the mean correlation computed separately within each radargram across its 256 nodes.}
\label{tab:mar_feature_summary}
\begin{tabular}{lcc}
\toprule
Feature & Test pooled $\rho$ & Test within-radargram $\rho$ \\
\midrule
Snow Mass balance (SMB)            & $0.9802 \pm 0.0019$ & $0.2492 \pm 0.0379$  \\
Average Yearly Surface Temperature(Surface Temp)  & $0.8963 \pm 0.0031$ & $0.1408 \pm 0.0185$  \\
Height Change Due To Refreezing (Refreezing)     & $0.3915 \pm 0.0382$ & $-0.0354 \pm 0.0335$ \\
Height Change Due To Melting (Melt) & $0.4191 \pm 0.0399$ & $-0.0418 \pm 0.0359$ \\
Snowpack Height (Snowpack)      & $0.0337 \pm 0.0558$ & $-0.0001 \pm 0.0219$ \\
\bottomrule
\end{tabular}
\end{table}

These two correlation measures have different interpretations. The pooled correlation captures global marginal association across the full test set and therefore includes both between-radargram and within-radargram variation. By contrast, the within-radargram correlation focuses on local horizontal variation within individual radargrams. Thus, a feature can have a high pooled correlation if it captures broad regional or climatic differences across radargrams, even if it has only a modest association with column-to-column thickness variation within a single radargram. This is observed in Table~\ref{tab:mar_feature_summary}. Snow mass balance and surface temperature show very high pooled correlations with thickness, but their within-radargram correlations are substantially smaller. This suggests that these variables primarily encode broad-scale thickness-related variation rather than detailed local column-wise structure.

To quantify predictive contribution, we used the complete 32 ablation table (Table~\ref{table:ablation_physics}) to compute the exact Shapley values~\cite{Shapley,molnar2025_Shapley} for the five MAR variables. Let $F$ denote the set of five physical variables and $S\subseteq F$ denote a selected feature subset. For a lower-is-better metric, such as RMSE or MAE, we define the coalition utility as 
\begin{equation}
v(S) = m(\emptyset) - m(S),
\end{equation}
where \(m(S)\) is the model error obtained using feature subset \(S\), and
\(m(\emptyset)\) is the error obtained without any MAR physical feature. With this definition, positive utility indicates improvement relative to the non-physical baseline. The exact Shapley value for feature $i$ is computed as 
\begin{equation}
\phi_i =
\sum_{S\subseteq F\setminus \{i\}}
\frac{|S|!(|F|-|S|-1)!}{|F|!}
\left[
v(S\cup\{i\}) - v(S)
\right].
\end{equation}

Because all 32 feature subsets were evaluated, the Shapley values are exact rather than sampled or approximated. Shapley values measure the average marginal contribution of each feature across all subset contexts and therefore quantify conditional predictive contribution, not simple standalone association. As shown in Figure~\ref{fig:physics_results_analysis}(\textbf{a}), the exact Shapley results show that snowpack has the largest predictive contribution, although its pooled and within-radargram correlations with thickness are near zero. This result highlights the difference between marginal association and conditional predictive utility. A variable may have weak standalone monotonic association with a scalar thickness summary, but still improve the full prediction task when used in combination with other variables. Conversely, a feature may be strongly correlated with thickness but contribute less additional predictive information if its signal overlaps with other inputs.

To assess such redundancy among the physical inputs, we computed a 5×5 MAR feature–feature Spearman correlation matrix. This matrix measures correlations among the MAR variables themselves, not correlations with thickness. Each MAR feature was first averaged across the five observed upper layers at each radargram column, and correlations were then computed between pairs of physical variables across aligned radargram-column samples. As shown in Figure~\ref{fig:physics_results_analysis}(\textbf{b}), the matrix reveals input-space redundancy: high feature–feature correlation indicates that two MAR variables carry overlapping information before being passed to the model. In our results, snow mass balance and average yearly surface temperature are strongly correlated, and height change due to refreezing and melting are also highly correlated, suggesting that these feature pairs encode overlapping physical information. Snowpack height is weakly correlated with the other MAR variables, indicating that its contribution is less likely to be explained by simple redundancy with the remaining features.

To further quantify model-level co-effects, we computed pairwise interaction scores from the same complete ablation utility table used for the Shapley analysis~\cite{Grabisch1999}. For two features $i$ and $j$, the interaction score is computed as 
\begin{equation}
I_{ij} =
\sum_{S\subseteq F\setminus \{i,j\}}
\frac{|S|!(|F|-|S|-2)!}{(|F|-1)!}
\left[
v(S\cup\{i,j\})
- v(S\cup\{i\})
- v(S\cup\{j\})
+ v(S)
\right].
\end{equation}

A positive $I_{ij}$ indicates synergy, meaning that the joint effect of two features is larger than expected from their separate contributions. A negative $I_{ij}$ indicates a sub-additive effect, meaning that the two features provide overlapping or partially redundant predictive information. This interaction formulation follows the Shapley interaction-index framework~\cite{Grabisch1999}. The interaction heatmap is therefore a model-level co-effect analysis, whereas the 5×5 MAR correlation matrix is an input-space redundancy analysis. From Figure~\ref{fig:physics_results_analysis}(\textbf{c}), we get all negative values for the pairwise interaction score, confirming that partially redundancy exists between MAR physical features.

These results explain why the best-performing feature subset does not necessarily contain the features with the largest feature–thickness correlations. Correlation is a one-feature-at-a-time measure of marginal association with a scalarized target. In contrast, ablation performance and exact Shapley values evaluate conditional contribution to the full predictive task in the presence of other variables. Therefore, a highly correlated feature can provide limited incremental value if its information is redundant with other physical variables or with other model inputs. Conversely, a weakly correlated feature can still be useful if it provides complementary information that improves prediction in combination with other variables.

Together, Figure~\ref{fig:physics_results_analysis} and Table~\ref{tab:mar_feature_summary} explain why the best-performing physical-feature subset includes average yearly surface temperature, height change due to melting, and snowpack height. The exact Shapley values show that snowpack height provides the largest conditional predictive contribution. The MAR feature--feature Spearman correlation matrix further shows substantial redundancy between snow mass balance and average yearly surface temperature, and between height change due to refreezing and height change due to melting. Thus, selecting both features from the same highly correlated pair may provide limited incremental benefit.

Although snow mass balance has the strongest pooled correlation with thickness, its information appears partly redundant with average yearly surface temperature. In addition, the pairwise interaction score between snow mass balance and snowpack height is more negative than that between average yearly surface temperature and snowpack height ($-0.11$ versus $-0.03$), suggesting stronger sub-additivity when snow mass balance is paired with snowpack height. This helps explain why average yearly surface temperature is retained in the optimal subset instead of snow mass balance. Similarly, height change due to melting has a slightly larger exact Shapley value and a more favorable interaction pattern than height change due to refreezing, supporting the selection of melting over refreezing.

Overall, the optimal subset should be interpreted as the combination with the highest joint predictive utility after accounting for redundancy and conditional contribution, rather than as the set of individually most correlated variables.

\begin{table}
\centering
\caption{Generalization of K-STEMIT to different number of prediction layers with a fixed input setting of $m=5$ observed layers. All experiments use the same subset of radargrams containing at least $5+21$ complete layers, so that changing the number of predicted layers $n$ does not change the training/test sample population. For smaller $n$, only the uppermost $n$ target layers below the five observed input layers are used. Errors are reported as mean $\pm$ standard deviation over repeated splits/runs.\label{table:generalization}}
\begin{tabular}{ccc}
\toprule
Number of Predicted Layers & All Layer RMSE    & All Layer MAE     \\ \midrule
5                          & 1.7716 $\pm$ 0.1020 & 1.1634 $\pm$ 0.0747 \\
8                          & 1.8359 $\pm$ 0.1048 & 1.2274 $\pm$ 0.0689 \\
10                         & 1.7973 $\pm$ 0.1191 & 1.2240 $\pm$ 0.0828 \\
13                         & 1.8103 $\pm$ 0.1160 & 1.2504 $\pm$ 0.0802 \\
15                         & 1.8724 $\pm$ 0.0949 & 1.2982 $\pm$ 0.0609 \\
18                         & 1.8838 $\pm$ 0.1251 & 1.3143 $\pm$ 0.0843 \\
21                         & 1.8969 $\pm$ 0.0998 & 1.3211 $\pm$ 0.0701 \\ \bottomrule
\end{tabular}
\end{table}

\begin{figure}
    {
        \includegraphics[width=0.8\textwidth]{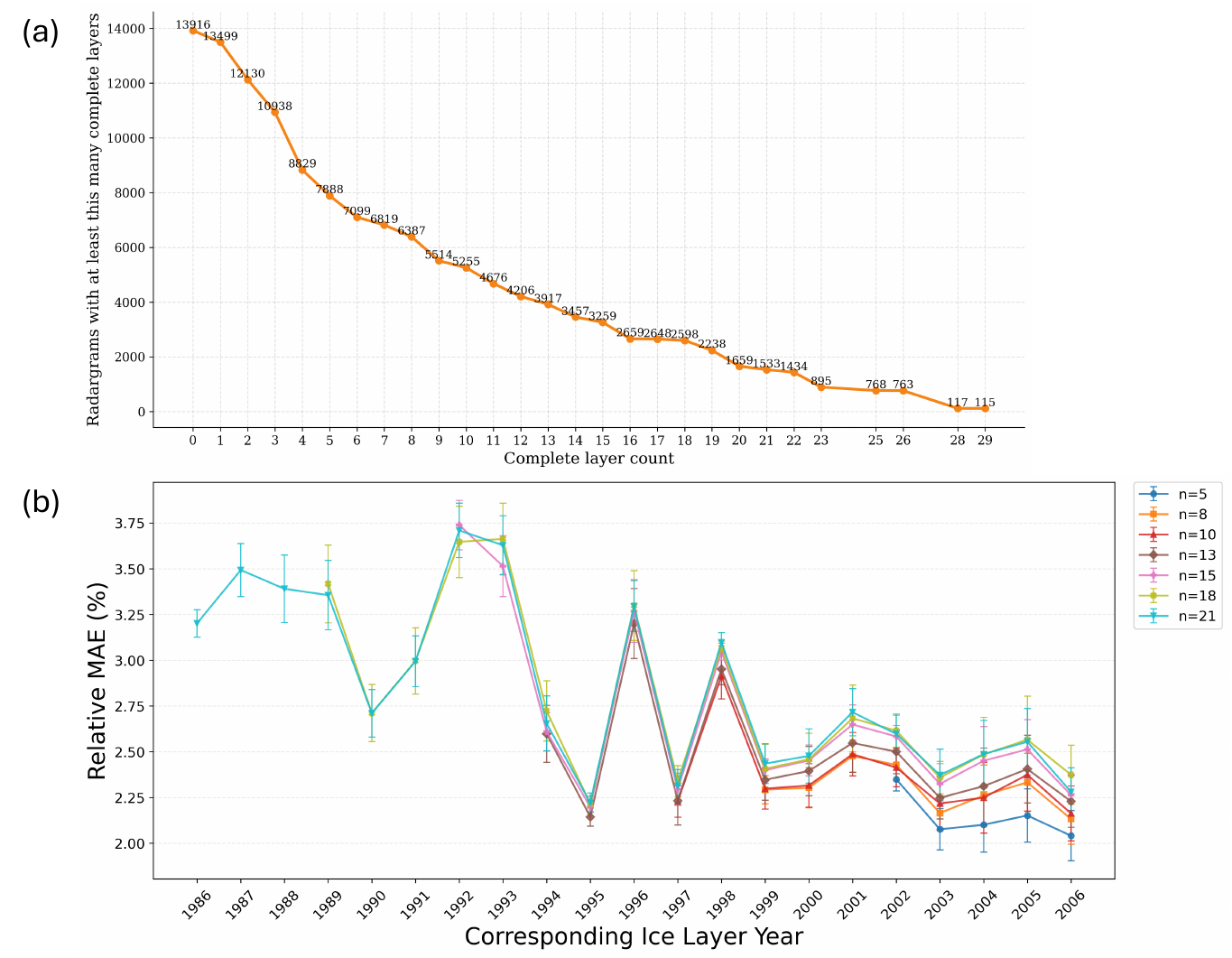}
    }
\caption{Data availability and prediction-horizon generalization for K-STEMIT. (\textbf{a}) Number of radargrams containing at least a given number of complete layers. Because the number of available radargrams decreases sharply for very large complete-layer counts, we set the maximum prediction horizon to $n=21$ with $m=5$ observed input layers, which requires at least $26$ complete layers per radargram. This choice balances deeper-layer evaluation with sufficient training data. (\textbf{b}) Relative MAE for different target-layer years under prediction horizons $n \in \{5,8,10,13,15,18,21\}$. Older years correspond to deeper internal layers, which are generally more challenging because radar signal quality and layer interpretability decrease with depth. }
\label{fig:multi_n_generalization}
\end{figure}

\subsection{Generalization to Deeper Layers}

The main experiments evaluate K-STEMIT using five observed upper layers as input and fifteen deeper layers as output, i.e., $m=5$ and $n=15$. To examine whether the model is tied to this specific input-output layer setting, we further evaluate its generalization ability under different numbers of predicted layers. In these experiments, the number of observed input layers is fixed at $m=5$, while the number of predicted layers is varied as $n \in \{5,8,10,13,15,18,21\}$.

A key consideration is that changing $n$ can also change the set of usable radargrams, since larger values of $n$ require more complete annotated layers. To avoid confounding the number of predicted layers with changes in the training and test sample distributions, we first construct a common subset of radargrams containing at least $5+21$ complete layers. All experiments with different values of $n$ are then conducted on this same subset. For smaller values of $n$, only the uppermost $n$ target layers below the five observed input layers are used. This design ensures that performance differences across $n$ primarily reflect the effect of predicting more target layers rather than differences in data availability.

Figure~\ref{fig:multi_n_generalization}(a) shows the number of radargrams available as a function of the required complete-layer count. Although SRED is, to our knowledge, the largest available snow-radar layer dataset~\cite{ibikunle2025_Dataset}, the number of available radargrams decreases substantially when requiring very large numbers of complete layers. We therefore use $n=21$ as the maximum number of predicted layers in this analysis, corresponding to at least $26$ complete layers per radargram. Using larger values of $n$ would substantially reduce the number of available training samples and increase the risk of under-fitting.

The quantitative results are summarized in Table~\ref{table:generalization}. As $n$ increases from $5$ to $21$, the all-layer RMSE changes from $1.7716 \pm 0.1020$ to $1.8969 \pm 0.0998$, and the all-layer MAE changes from $1.1634 \pm 0.0747$ to $1.3211 \pm 0.0701$. The degradation is moderate despite increasing the number of predicted target layers from 5 to 21. This indicates that K-STEMIT is not restricted to the specific $m=5,n=15$ setting used in the main experiments, but can generalize to different target-layer counts when sufficient complete-layer training data are available.

We further analyze the layer-wise behavior by grouping prediction errors according to the corresponding ice-layer year, as shown in Figure~\ref{fig:multi_n_generalization}(b). Older years correspond to deeper internal layers, which are generally more difficult to extract from snow radar echograms because layer boundaries become less visually distinct and radar signal quality decreases with depth. The relative MAE varies across target years, indicating that deeper-year prediction is not uniformly easy. Nevertheless, the error curves remain within a comparable range across different values of $n$, suggesting that increasing the number of predicted layers does not cause unstable behavior across the shared target layers.

This analysis also provides a practical evaluation of K-STEMIT under more challenging prediction conditions. Since K-STEMIT operates on graph-structured layer representations, the generalization study is performed at the level of derived layer geometry and thickness information, rather than raw radargram intensities. In this setting, deeper and older layers are particularly important because they are more likely to be affected by lower radar signal quality, weaker layer continuity, and greater interpretation uncertainty. The stable performance across target-layer counts up to $n=21$ suggests that K-STEMIT can generalize beyond the standard $m=5,n=15$ setting to deeper target layers when sufficient complete-layer training data are available.

\begin{figure}
    {
        \includegraphics[width=1\textwidth]{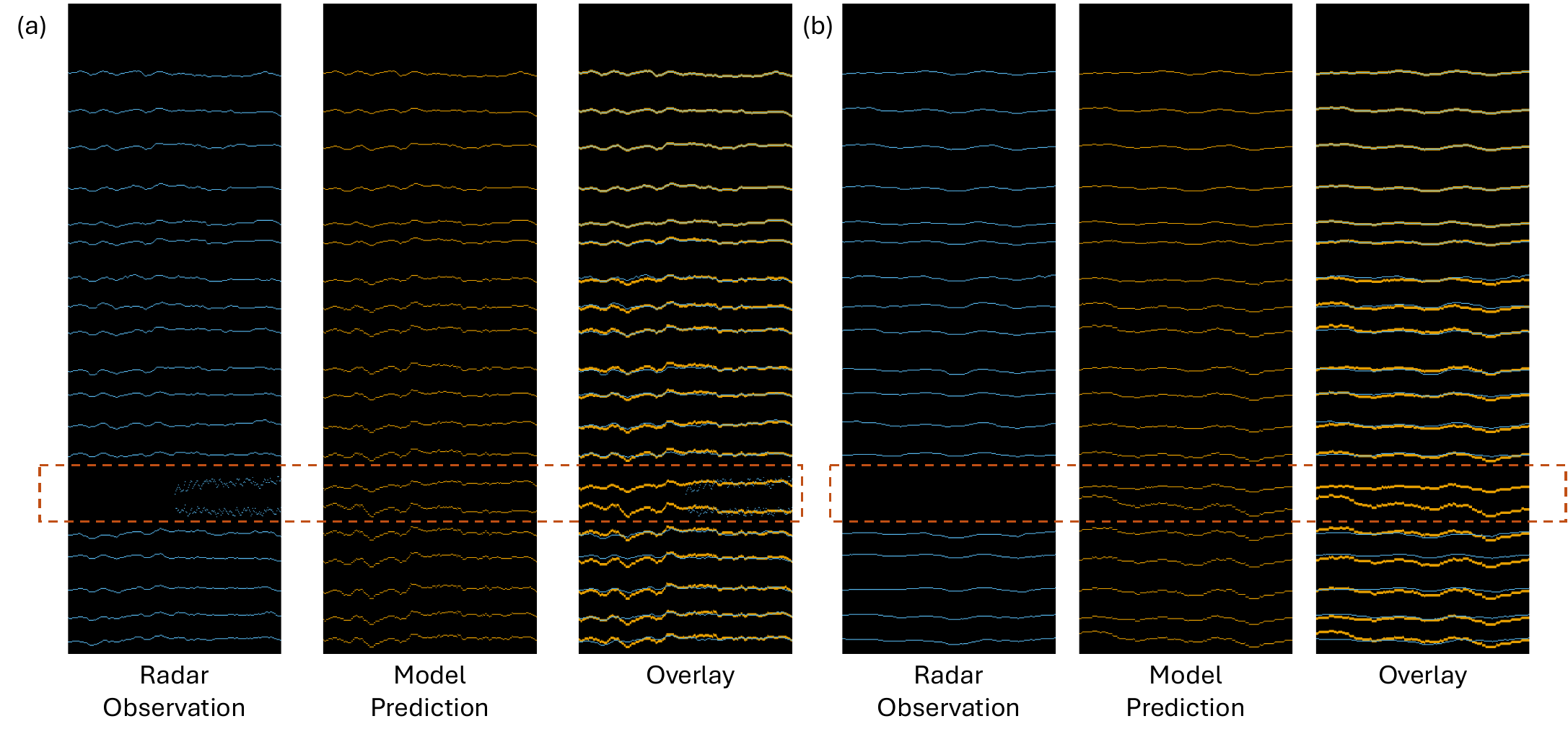}
    }
\caption{Qualitative examples under two types of incomplete deeper target-layer annotation: \textbf{(a)} some deeper target layers are partially missing, and \textbf{(b)} some deeper target layers are entirely missing. In both cases, the five upper observed layers are used as inputs to K-STEMIT, which predicts the deeper layers across the radargram. Where target layer annotations are available, the predictions can be directly compared with the annotated layers; where annotations are missing, the model still produces spatially continuous layer estimates.}
\label{fig:partial_target_qualitative}
\end{figure}

To further illustrate the behavior of K-STEMIT in cases with incomplete deeper annotations, Figure~\ref{fig:partial_target_qualitative} shows qualitative examples where the five observed input layers are complete but some deeper target layers are only partially annotated. K-STEMIT still produces spatially continuous predictions for the deeper layers. For such cases, if needed, quantitative errors can be computed only at target locations with valid annotations, avoiding the introduction of artificial labels in unannotated regions. When the observed input layers themselves are incomplete or fragmented, an upstream layer-completion or imputation step is required before applying K-STEMIT. We refer readers to our recent work on layer completion~\cite{liu2026physicsconditionedsynthesisinternalicelayer}, where layer-completion and synthetic-boundary generation approaches are used to recover or augment incomplete layer annotations. Such generated boundaries may also provide useful pretraining supervision for K-STEMIT.

\subsection{Current Limitations and Future Work}

The present study uses the SRED Greenland dataset~\cite{ibikunle2025_Dataset} as the primary benchmark because, to the best of our knowledge, SRED is the first and currently only comprehensive AI-ready benchmark that provides radar-derived ice-layer annotations together with synchronized physical variables for knowledge-informed ice-layer thickness prediction. This unique data availability makes SRED the most suitable standardized benchmark for systematically evaluating K-STEMIT under a controlled experimental setting. Accordingly, the results in this work demonstrate the effectiveness of the proposed knowledge-informed multi-branch graph neural network framework on the available Greenland snow-radar benchmark.

At the same time, broader cross-regional evaluation remains an important direction for future work. Although airborne radar measurements are available for other polar regions, including Antarctica, many existing data resources do not yet provide the consistent ice-layer annotations and synchronized physical variables required for supervised knowledge-informed learning. This limitation is important because polar regions can differ substantially in accumulation regimes, ice dynamics, radar acquisition characteristics, and subsurface stratigraphic patterns. As a result, evaluating the cross-regional transferability of K-STEMIT will require comparable AI-ready datasets with both annotated layer boundaries and matched physical variables. Once such data resources become available, future work will conduct explicit multi-regional transfer experiments to further assess the generalization ability of K-STEMIT across diverse polar environments.

Another future direction is to further investigate adaptive fusion under broader multi-regional settings. In the current K-STEMIT design, the fusion coefficient $\alpha$ is learned as a globally shared parameter to balance the spatial and temporal branches. As discussed in Section~\ref{sec:ablation-alpha}, we also examined more fine-grained variants based on regional coefficients $\alpha_i$, layer-wise coefficients $\alpha_\ell$, and joint regional-layer coefficients $\alpha_{i,\ell}$. Under the current SRED benchmark setting, these more flexible variants did not improve performance over the globally shared coefficient, suggesting that the additional flexibility may be unnecessary for the available single-benchmark dataset.

Nevertheless, regional or depth-adaptive fusion may become more meaningful when larger and more heterogeneous polar radar datasets become available. For example, multi-regional training data may contain stronger spatial heterogeneity, depth-dependent stratigraphic variation, and acquisition-dependent differences across polar environments. In such settings, regional coefficients could allow the model to adapt fusion behavior across geographic locations, while layer-wise coefficients could capture different spatial-temporal dependencies across shallow and deep target layers. Future work will therefore revisit these fine-grained fusion mechanisms in multi-regional studies, potentially with additional smoothness constraints or regularization to improve robustness and avoid overfitting.
\section{Conclusion}\label{conclusion}
In this work, we developed K-STEMIT, a knowledge-informed spatio-temporal efficient multi-branch graph neural network to learn from the geographical and thickness information of the top $m$ ice layers and predict the thickness of the underlying $n$ layer. Our multi-branch architecture integrates a GraphSAGE-based inductive framework in the spatial branch, gated temporal convolution in the temporal branch, and dimensionality reduction at the start of each branch to eliminate irrelevant features, and physical node features are incorporated from the MAR physical model. We also implement an adaptive feature fusion strategy that combines features from different branches dynamically through learnable parameters.

We evaluated our proposed K-STEMIT on a specific case that uses the information of Greenland ice layers formed from 2007-2011 to predict the thickness of ice layers formed from 1992 to 2006. Notably, within the Greenland SRED benchmark considered in this study, the proposed network is designed to accommodate a variable number of ice layers and radargrams with different spatial sizes. Extensive experiments demonstrate that our proposed K-STEMIT outperforms both knowledge-informed and non-knowledge-informed state-of-the-art methods with the lowest RMSE error and nearly the lowest average computation time. More importantly, we found that when combined with physical node features as prior domain knowledge and applied an adaptive feature fusion strategy, K-STEMIT achieves a 21.99\% reduction in RMSE with negligible additional cost compared with the non-knowledge-informed multi-branch network.

Through ablation studies, we identified the optimal combination of physical features—snow mass balance, meltwater refreezing, height change due to melting, and snowpack height—which achieves the lowest RMSE. Additionally, we evaluate the MAE error across layers formed in different years. Lastly, we observe that our K-STEMIT yields the lowest relative MAE compared to current state-of-the-art methods and other existing methods, demonstrating a stable error distribution across ice layers and a consistent performance across different years. 

Finally, this study focuses on the SRED Greenland benchmark, which is, to the best of our knowledge, the first and currently only comprehensive AI-ready dataset that provides radar-derived ice-layer annotations together with synchronized physical variables for knowledge-informed ice-layer thickness prediction. Although airborne radar measurements are available for other polar regions, including Antarctica, many existing resources are not yet accompanied by consistent layer annotations and matched physical variables required for supervised knowledge-informed learning. As broader AI-ready polar radar datasets become available, future work will extend K-STEMIT to multi-regional transfer evaluation and further investigate fine-grained fusion strategies, such as regional, layer-wise, and joint regional-layer adaptive coefficients.












\printcredits

\section*{Acknowledgement}
This work is supported by NSF BIGDATA awards (IIS-1838230, IIS-2308649), NSF Leadership Class Computing awards (OAC-2139536), NSF PFI awards (2423211). We acknowledge data and data products from CReSIS generated with support from the University of Kansas and NASA Operation Ice-Bridge. We also acknowledge Oluwanisola Ibikunle for providing necessary dataset descriptions and physical analysis of the experiment results.

\bibliographystyle{cas-model2-names}

\bibliography{cas-refs}

@misc{diagram-airborne-radar,
  author = {{NASA Scientific Visualization Studio}},
  title = {Greenland Ice Sheet Stratigraphy},
  year = {2015},
  howpublished = {NASA Scientific Visualization Studio},
  note = {Available online}
}

@article{Arnold_Leuschen_Rodriguez-Morales_Li_Paden_Hale_Keshmiri_2020, title={CReSIS airborne radars and platforms for ice and snow sounding}, volume={61}, DOI={10.1017/aog.2019.37}, number={81}, journal={Annals of Glaciology}, author={Arnold, Emily and Leuschen, Carl and Rodriguez-Morales, Fernando and Li, Jilu and Paden, John and Hale, Richard and Keshmiri, Shawn}, year={2020}, pages={58–67}}

@misc{Leuschen2011SnowRadar,
  author = {Leuschen, Carl and Panzer, Ben and Gogineni, Prasad and Rodriguez, Fernando and Paden, John and Li, Jilu},
  title = {IceBridge Snow Radar L1B Geolocated Radar Echo Strength Profiles},
  year = {2011/2024},  
  howpublished = {Boulder, Colorado USA: National Snow and Ice Data Center. Digital media.},
  note = {Accessed on 2024}  
}

@ARTICLE{7731235,
  author={Carrer, Leonardo and Bruzzone, Lorenzo},
  journal={IEEE Trans. Geosci. Remote Sens.}, 
  title={Automatic Enhancement and Detection of Layering in Radar Sounder Data Based on a Local Scale Hidden Markov Model and the Viterbi Algorithm}, 
  year={2017},
  volume={55},
  number={2},
  pages={962-977},
  doi={10.1109/TGRS.2016.2616949}}

@article{https://doi.org/10.1002/2014JF003215,
author = {MacGregor, Joseph A. and Fahnestock, Mark A. and Catania, Ginny A. and Paden, John D. and Prasad Gogineni, S. and Young, S. Keith and Rybarski, Susan C. and Mabrey, Alexandria N. and Wagman, Benjamin M. and Morlighem, Mathieu},
title = {Radiostratigraphy and age structure of the Greenland Ice Sheet},
journal = {Journal of Geophysical Research: Earth Surface},
volume = {120},
number = {2},
pages = {212-241},
year = {2015},
doi = {https://doi.org/10.1002/2014JF003215},
url = {https://agupubs.onlinelibrary.wiley.com/doi/abs/10.1002/2014JF003215},
eprint = {https://agupubs.onlinelibrary.wiley.com/doi/pdf/10.1002/2014JF003215},
}

@article{article,
author = {Panton, Christian},
year = {2013},
month = {11},
pages = {},
title = {Automated mapping of local layer slope and tracing of internal layers in radio echograms},
volume = {55},
journal = {Annals of Glaciology},
doi = {10.3189/2014AoG67A048}
}

@INPROCEEDINGS{DeepIceLayerTracking,
  author={Varshney, Debvrat and Rahnemoonfar, Maryam and Yari, Masoud and Paden, John},
  booktitle={2020 IEEE International Conference on Big Data (Big Data)}, 
  title={Deep Ice Layer Tracking and Thickness Estimation using Fully Convolutional Networks}, 
  year={2020},
  volume={},
  number={},
  pages={3943-3952},
  doi={10.1109/BigData50022.2020.9378070}}

@Article{DeepLearningOnAirborneRadar,
AUTHOR = {Varshney, Debvrat and Rahnemoonfar, Maryam and Yari, Masoud and Paden, John and Ibikunle, Oluwanisola and Li, Jilu},
TITLE = {Deep Learning on Airborne Radar Echograms for Tracing Snow Accumulation Layers of the Greenland Ice Sheet},
JOURNAL = {Remote Sensing},
VOLUME = {13},
YEAR = {2021},
NUMBER = {14},
ARTICLE-NUMBER = {2707},
URL = {https://www.mdpi.com/2072-4292/13/14/2707},
ISSN = {2072-4292},
DOI = {10.3390/rs13142707}
}

@article{Rahnemoonfar_Yari_Paden_Koenig_Ibikunle_2021, title={Deep multi-scale learning for automatic tracking of internal layers of ice in radar data}, volume={67}, DOI={10.1017/jog.2020.80}, number={261}, journal={Journal of Glaciology}, author={Rahnemoonfar, Maryam and Yari, Masoud and Paden, John and Koenig, Lora and Ibikunle, Oluwanisola}, year={2021}, pages={39–48}}

@INPROCEEDINGS{Yari_2020,
  author={Yari, Masoud and Rahnemoonfar, Maryam and Paden, John},
  booktitle={IGARSS 2020 - 2020 IEEE International Geoscience and Remote Sensing Symposium}, 
  title={Multi-Scale and Temporal Transfer Learning for Automatic Tracking of Internal Ice Layers}, 
  year={2020},
  volume={},
  number={},
  pages={6934-6937},
  doi={10.1109/IGARSS39084.2020.9323758}}

@INPROCEEDINGS{LearnSnowLayerThickness,
  author={Varshney, Debvrat and Ibikunle, Oluwanisola and Paden, John and Rahnemoonfar, Maryam},
  booktitle={IGARSS 2022 - 2022 IEEE International Geoscience and Remote Sensing Symposium}, 
  title={Learning Snow Layer Thickness Through Physics Defined Labels}, 
  year={2022},
  volume={},
  number={},
  pages={1233-1236},
  doi={10.1109/IGARSS46834.2022.9884370}}

@INPROCEEDINGS{DeepHybridWavelet,
  author={Kamangir, Hamid and Rahnemoonfar, Maryam and Dobbs, Dugan and Paden, John and Fox, Geoffrey},
  booktitle={IGARSS 2018 - 2018 IEEE International Geoscience and Remote Sensing Symposium}, 
  title={Deep Hybrid Wavelet Network for Ice Boundary Detection in Radra Imagery}, 
  year={2018},
  volume={},
  number={},
  pages={3449-3452},
  doi={10.1109/IGARSS.2018.8518617}}

@inproceedings{varshney2021refining,
  title={Refining Ice Layer Tracking through Wavelet combined Neural Networks},
  author={Varshney, Debvrat and Yari, Masoud and Chowdhury, Tashnim and Rahnemoonfar, Maryam},
  booktitle={ICML 2021 Workshop on Tackling Climate Change with Machine Learning},
  url={https://www.climatechange.ai/papers/icml2021/49},
  year={2021}
}

@ARTICLE{Yari_2021_JSTAR,
  author={Yari, Masoud and Ibikunle, Oluwanisola and Varshney, Debvrat and Chowdhury, Tashnim and Sarkar, Argho and Paden, John and Li, Jilu and Rahnemoonfar, Maryam},
  journal={IEEE J. Sel. Topics Appl. Earth Observ. Remote Sens.}, 
  title={Airborne Snow Radar Data Simulation With Deep Learning and Physics-Driven Methods}, 
  year={2021},
  volume={14},
  number={},
  pages={12035-12047},
  doi={10.1109/JSTARS.2021.3126547}}

@article{ZHOU202057,
title = {Graph neural networks: A review of methods and applications},
journal = {AI Open},
volume = {1},
pages = {57-81},
year = {2020},
issn = {2666-6510},
doi = {https://doi.org/10.1016/j.aiopen.2021.01.001},
url = {https://www.sciencedirect.com/science/article/pii/S2666651021000012},
author = {Jie Zhou and Ganqu Cui and Shengding Hu and Zhengyan Zhang and Cheng Yang and Zhiyuan Liu and Lifeng Wang and Changcheng Li and Maosong Sun}
}

@INPROCEEDINGS{Zalatan_igarss,
  author={Zalatan, Benjamin and Rahnemoonfar, Maryam},
  booktitle={IGARSS 2023 - 2023 IEEE International Geoscience and Remote Sensing Symposium}, 
  title={Prediction of Annual Snow Accumulation Using a Recurrent Graph Convolutional Approach}, 
  year={2023},
  volume={},
  number={},
  pages={5344-5347},
  doi={10.1109/IGARSS52108.2023.10283236}}

@INPROCEEDINGS{zalatan_icip,
  author={Zalatan, Benjamin and Rahnemoonfar, Maryam},
  booktitle={2023 IEEE International Conference on Image Processing (ICIP)}, 
  title={Prediction of Deep Ice Layer Thickness Using Adaptive Recurrent Graph Neural Networks}, 
  year={2023},
  volume={},
  number={},
  pages={2835-2839},
  doi={10.1109/ICIP49359.2023.10222391}}

@INPROCEEDINGS{Zalatan2023,
  author={Zalatan, Benjamin and Rahnemoonfar, Maryam},
  booktitle={2023 IEEE Radar Conference (RadarConf23)}, 
  title={Recurrent Graph Convolutional Networks for Spatiotemporal Prediction of Snow Accumulation Using Airborne Radar}, 
  year={2023},
  volume={},
  number={},
  pages={1-6},
  doi={10.1109/RadarConf2351548.2023.10149562}}

@misc{seo2016structured,
      title={Structured Sequence Modeling with Graph Convolutional Recurrent Networks}, 
      author={Youngjoo Seo and Michaël Defferrard and Pierre Vandergheynst and Xavier Bresson},
      year={2016},
      eprint={1612.07659},
      archivePrefix={arXiv},
      primaryClass={stat.ML}
}

@article{EGCN,
  author       = {Aldo Pareja and
                  Giacomo Domeniconi and
                  Jie Chen and
                  Tengfei Ma and
                  Toyotaro Suzumura and
                  Hiroki Kanezashi and
                  Tim Kaler and
                  Charles E. Leiserson},
  title        = {EvolveGCN: Evolving Graph Convolutional Networks for Dynamic Graphs},
  journal      = {CoRR},
  volume       = {abs/1902.10191},
  year         = {2019},
  url          = {http://arxiv.org/abs/1902.10191},
  eprinttype    = {arXiv},
  eprint       = {1902.10191},
  bibsource    = {dblp computer science bibliography, https://dblp.org}
}

@Inbook{Forsberg2017,
author="Forsberg, Rene
and S{\o}rensen, Louise
and Simonsen, Sebastian",
title="Greenland and Antarctica Ice Sheet Mass Changes and Effects on Global Sea Level",
bookTitle="Integrative Study of the Mean Sea Level and Its Components",
year="2017",
publisher="Springer International Publishing",
address="Cham",
pages="91--106",
isbn="978-3-319-56490-6",
doi="10.1007/978-3-319-56490-6_5",
url="https://doi.org/10.1007/978-3-319-56490-6_5"
}

@article{DIEBOLD2023105479,
title = {When will Arctic sea ice disappear? Projections of area, extent, thickness, and volume},
journal = {Journal of Econometrics},
volume = {236},
number = {2},
pages = {105479},
year = {2023},
issn = {0304-4076},
doi = {https://doi.org/10.1016/j.jeconom.2023.105479},
url = {https://www.sciencedirect.com/science/article/pii/S0304407623001951},
author = {Francis X. Diebold and Glenn D. Rudebusch and Maximilian Göbel and Philippe {Goulet Coulombe} and Boyuan Zhang}
}

@Article{Shepherd2020,
author={Shepherd, Andrew
and Ivins, Erik
and Rignot, Eric
and Smith, Ben
and van den Broeke, Michiel
and Velicogna, Isabella
and Whitehouse, Pippa
and Briggs, Kate
and Joughin, Ian
and Krinner, Gerhard
and Nowicki, Sophie
and Payne, Tony
and Scambos, Ted
and Schlegel, Nicole
and A, Geruo
and Agosta, C{\'e}cile
and Ahlstr{\o}m, Andreas
and Babonis, Greg
and Barletta, Valentina R.
and Bj{\o}rk, Anders A.
and Blazquez, Alejandro
and Bonin, Jennifer
and Colgan, William
and Csatho, Beata
and Cullather, Richard
and Engdahl, Marcus E.
and Felikson, Denis
and Fettweis, Xavier
and Forsberg, Rene
and Hogg, Anna E.
and Gallee, Hubert
and Gardner, Alex
and Gilbert, Lin
and Gourmelen, Noel
and Groh, Andreas
and Gunter, Brian
and Hanna, Edward
and Harig, Christopher
and Helm, Veit
and Horvath, Alexander
and Horwath, Martin
and Khan, Shfaqat
and Kjeldsen, Kristian K.
and Konrad, Hannes
and Langen, Peter L.
and Lecavalier, Benoit
and Loomis, Bryant
and Luthcke, Scott
and McMillan, Malcolm
and Melini, Daniele
and Mernild, Sebastian
and Mohajerani, Yara
and Moore, Philip
and Mottram, Ruth
and Mouginot, Jeremie
and Moyano, Gorka
and Muir, Alan
and Nagler, Thomas
and Nield, Grace
and Nilsson, Johan
and No{\"e}l, Brice
and Otosaka, Ines
and Pattle, Mark E.
and Peltier, W. Richard
and Pie, Nad{\`e}ge
and Rietbroek, Roelof
and Rott, Helmut
and Sandberg S{\o}rensen, Louise
and Sasgen, Ingo
and Save, Himanshu
and Scheuchl, Bernd
and Schrama, Ernst
and Schr{\"o}der, Ludwig
and Seo, Ki-Weon
and Simonsen, Sebastian B.
and Slater, Thomas
and Spada, Giorgio
and Sutterley, Tyler
and Talpe, Matthieu
and Tarasov, Lev
and van de Berg, Willem Jan
and van der Wal, Wouter
and van Wessem, Melchior
and Vishwakarma, Bramha Dutt
and Wiese, David
and Wilton, David
and Wagner, Thomas
and Wouters, Bert
and Wuite, Jan
and Team, The IMBIE},
title={Mass balance of the Greenland Ice Sheet from 1992 to 2018},
journal={Nature},
year={2020},
month={Mar},
day={01},
volume={579},
number={7798},
pages={233-239},
issn={1476-4687},
doi={10.1038/s41586-019-1855-2},
url={https://doi.org/10.1038/s41586-019-1855-2}
}

@incollection{PATERSON1994378,
title = {15 - Ice Core Studies},
editor = {W.S.B. PATERSON},
booktitle = {The Physics of Glaciers (Third Edition)},
publisher = {Pergamon},
edition = {Third Edition},
address = {Amsterdam},
pages = {378-409},
year = {1994},
isbn = {978-0-08-037944-9},
doi = {https://doi.org/10.1016/B978-0-08-037944-9.50021-2},
url = {https://www.sciencedirect.com/science/article/pii/B9780080379449500212},
author = {W.S.B. PATERSON}
}

@INPROCEEDINGS{snowradar,
  author={Gogineni, S. and Yan, J. B. and Gomez, D. and Rodriguez-Morales, F. and Paden, J. and Leuschen, C.},
  booktitle={IEEE MTT-S International Microwave and RF Conference}, 
  title={Ultra-wideband radars for remote sensing of snow and ice}, 
  year={2013},
  volume={},
  number={},
  pages={1-4},
  doi={10.1109/IMaRC.2013.6777743}}

@incollection{AirborneRadar,
title = {Chapter 12 - Glaciers and Ice Sheets},
editor = {Harry M. Jol},
booktitle = {Ground Penetrating Radar Theory and Applications},
publisher = {Elsevier},
address = {Amsterdam},
pages = {361-392},
year = {2009},
isbn = {978-0-444-53348-7},
doi = {https://doi.org/10.1016/B978-0-444-53348-7.00012-0},
url = {https://www.sciencedirect.com/science/article/pii/B9780444533487000120},
author = {Steven A. Arcone}
}

@misc{kipf2017GCN,
      title={Semi-Supervised Classification with Graph Convolutional Networks}, 
      author={Thomas N. Kipf and Max Welling},
      year={2017},
      eprint={1609.02907},
      archivePrefix={arXiv},
      primaryClass={cs.LG}
}

@misc{hamilton2018GraphSAGE,
      title={Inductive Representation Learning on Large Graphs}, 
      author={William L. Hamilton and Rex Ying and Jure Leskovec},
      year={2018},
      eprint={1706.02216},
      archivePrefix={arXiv},
      primaryClass={cs.SI}
}

@misc{liu2024learningspatiotemporalpatternspolar,
      title={Learning Spatio-Temporal Patterns of Polar Ice Layers With Physics-Informed Graph Neural Network}, 
      author={Zesheng Liu and Maryam Rahnemoonfar},
      year={2024},
      eprint={2406.15299},
      archivePrefix={arXiv},
      primaryClass={cs.LG},
      url={https://arxiv.org/abs/2406.15299}, 
}

@Article{Koenig_2016,
AUTHOR = {Koenig, L. S. and Ivanoff, A. and Alexander, P. M. and MacGregor, J. A. and Fettweis, X. and Panzer, B. and Paden, J. D. and Forster, R. R. and Das, I. and McConnell, J. R. and Tedesco, M. and Leuschen, C. and Gogineni, P.},
TITLE = {Annual Greenland accumulation rates (2009--2012) from airborne snow radar},
JOURNAL = {The Cryosphere},
VOLUME = {10},
YEAR = {2016},
NUMBER = {4},
PAGES = {1739--1752},
URL = {https://tc.copernicus.org/articles/10/1739/2016/},
DOI = {10.5194/tc-10-1739-2016}
}

@misc{gehring2017convolutional,
      title={Convolutional Sequence to Sequence Learning}, 
      author={Jonas Gehring and Michael Auli and David Grangier and Denis Yarats and Yann N. Dauphin},
      year={2017},
      eprint={1705.03122},
      archivePrefix={arXiv},
      primaryClass={cs.CL}
}

@misc{kingma2017adam,
      title={Adam: A Method for Stochastic Optimization}, 
      author={Diederik P. Kingma and Jimmy Ba},
      year={2017},
      eprint={1412.6980},
      archivePrefix={arXiv},
      primaryClass={cs.LG}
}

@article{GCN,
  author       = {Thomas N. Kipf and
                  Max Welling},
  title        = {Semi-Supervised Classification with Graph Convolutional Networks},
  journal      = {CoRR},
  volume       = {abs/1609.02907},
  year         = {2016},
  url          = {http://arxiv.org/abs/1609.02907},
  eprinttype    = {arXiv},
  eprint       = {1609.02907},
  bibsource    = {dblp computer science bibliography, https://dblp.org}
}

@INPROCEEDINGS{CReSIS_radar,
  author={Gogineni, S. and Yan, J. B. and Gomez, D. and Rodriguez-Morales, F. and Paden, J. and Leuschen, C.},
  booktitle={IEEE MTT-S International Microwave and RF Conference}, 
  title={Ultra-wideband radars for remote sensing of snow and ice}, 
  year={2013},
  volume={},
  number={},
  pages={1-4},
  doi={10.1109/IMaRC.2013.6777743}}

@article{LSTM,
    author = {Hochreiter, Sepp and Schmidhuber, Jürgen},
    title = "{Long Short-Term Memory}",
    journal = {Neural Computation},
    volume = {9},
    number = {8},
    pages = {1735-1780},
    year = {1997},
    month = {11},
    issn = {0899-7667},
    doi = {10.1162/neco.1997.9.8.1735},
    url = {https://doi.org/10.1162/neco.1997.9.8.1735},
    eprint = {https://direct.mit.edu/neco/article-pdf/9/8/1735/813796/neco.1997.9.8.1735.pdf},
}

@Article{MAR_2020,
AUTHOR = {Delhasse, A. and Kittel, C. and Amory, C. and Hofer, S. and van As, D. and S. Fausto, R. and Fettweis, X.},
TITLE = {Brief communication: Evaluation of the near-surface climate in ERA5 over the Greenland Ice Sheet},
JOURNAL = {The Cryosphere},
VOLUME = {14},
YEAR = {2020},
NUMBER = {3},
PAGES = {957--965},
URL = {https://tc.copernicus.org/articles/14/957/2020/},
DOI = {10.5194/tc-14-957-2020}
}

@Article{MAR2021,
AUTHOR = {Fettweis, X. and Hofer, S. and S\'ef\'erian, R. and Amory, C. and Delhasse, A. and Doutreloup, S. and Kittel, C. and Lang, C. and Van Bever, J. and Veillon, F. and Irvine, P.},
TITLE = {Brief communication: Reduction in the future
Greenland ice sheet surface melt with the help of solar geoengineering},
JOURNAL = {The Cryosphere},
VOLUME = {15},
YEAR = {2021},
NUMBER = {6},
PAGES = {3013--3019},
URL = {https://tc.copernicus.org/articles/15/3013/2021/},
DOI = {10.5194/tc-15-3013-2021}
}

@INPROCEEDINGS{PI_GCNLSTM,
  author={Rahnemoonfar, Maryam and Zalatan, Benjamin},
  booktitle={IGARSS 2024 - 2024 IEEE International Geoscience and Remote Sensing Symposium}, 
  title={Physics-informed Machine Learning for Deep Ice Layer Tracing in SAR images}, 
  year={2024},
  volume={},
  number={},
  pages={6938-6942},
  doi={10.1109/IGARSS53475.2024.10641831}}

@Article{Karniadakis2021,
author={Karniadakis, George Em
and Kevrekidis, Ioannis G.
and Lu, Lu
and Perdikaris, Paris
and Wang, Sifan
and Yang, Liu},
title={Physics-informed machine learning},
journal={Nature Reviews Physics},
year={2021},
month={Jun},
day={01},
volume={3},
number={6},
pages={422-440},
issn={2522-5820},
doi={10.1038/s42254-021-00314-5},
url={https://doi.org/10.1038/s42254-021-00314-5}
}

@Article{Desai2021,
author={Desai, Saaketh
and Strachan, Alejandro},
title={Parsimonious neural networks learn interpretable physical laws},
journal={Scientific Reports},
year={2021},
month={Jun},
day={17},
volume={11},
number={1},
pages={12761},
issn={2045-2322},
doi={10.1038/s41598-021-92278-w},
url={https://doi.org/10.1038/s41598-021-92278-w}
}

@INPROCEEDINGS{varshney_2021_regression_networks,
  author={Varshney, Debvrat and Rahnemoonfar, Maryam and Yari, Masoud and Paden, John},
  booktitle={2021 IEEE International Geoscience and Remote Sensing Symposium IGARSS}, 
  title={Regression Networks for Calculating Englacial Layer Thickness}, 
  year={2021},
  volume={},
  number={},
  pages={2393-2396},
  doi={10.1109/IGARSS47720.2021.9553596}}

@INPROCEEDINGS{ibikunle_2020,
  author={Ibikunle, Oluwanisola and Paden, John and Rahnemoonfar, Maryam and Crandall, David and Yari, Masoud},
  booktitle={IGARSS 2020 - 2020 IEEE International Geoscience and Remote Sensing Symposium}, 
  title={Snow Radar Layer Tracking Using Iterative Neural Network Approach}, 
  year={2020},
  volume={},
  number={},
  pages={2960-2963},
doi={10.1109/IGARSS39084.2020.9323957}}

@INPROCEEDINGS{ibikunle_snow_radar_echogram_2023,
  author={Ibikunle, Oluwanisola and Talasila, Hara Madhav and Varshney, Debvrat and Paden, John D. and Li, Jilu and Rahnemoonfar, Maryam},
  booktitle={2023 IEEE Radar Conference (RadarConf23)}, 
  title={Snow Radar Echogram Layer Tracker: Deep Neural Networks for radar data from NASA Operation IceBridge}, 
  year={2023},
  volume={},
  number={},
  pages={1-6},
  doi={10.1109/RadarConf2351548.2023.10149734}}

@article{varshney2023skipwavenet, title={Skip-WaveNet: a wavelet based multi-scale architecture to trace snow layers in radar echograms}, volume={3}, DOI={10.1017/eds.2024.25}, journal={Environmental Data Science}, author={Varshney, Debvrat and Yari, Masoud and Ibikunle, Oluwanisola and Li, Jilu and Paden, John and Gangopadhyay, Aryya and Rahnemoonfar, Maryam}, year={2024}, pages={e39}}

@misc{morel1988introduction,
  title={An Introduction to Three-Dimensional Climate Modeling},
  author={Morel, Pierre},
  year={1988},
  publisher={Wiley Online Library}
}

@book{washington2005introduction,
  title={Introduction to three-dimensional climate modeling},
  author={Washington, Warren M and Parkinson, Claire},
  year={2005},
  publisher={University science books}
}

@book{mcguffie2014climate,
  title={The climate modelling primer},
  author={McGuffie, Kendal and Henderson-Sellers, Ann},
  year={2014},
  publisher={John Wiley \& Sons}
}

@article{hersbach2020era5,
  title={The ERA5 global reanalysis},
  author={Hersbach, Hans and Bell, Bill and Berrisford, Paul and Hirahara, Shoji and Hor{\'a}nyi, Andr{\'a}s and Mu{\~n}oz-Sabater, Joaqu{\'\i}n and Nicolas, Julien and Peubey, Carole and Radu, Raluca and Schepers, Dinand and others},
  journal={Quarterly Journal of the Royal Meteorological Society},
  volume={146},
  number={730},
  pages={1999--2049},
  year={2020},
  publisher={Wiley Online Library}
}

@article{soci2024era5,
  title={The ERA5 global reanalysis from 1940 to 2022},
  author={Soci, Cornel and Hersbach, Hans and Simmons, Adrian and Poli, Paul and Bell, Bill and Berrisford, Paul and Hor{\'a}nyi, Andr{\'a}s and Mu{\~n}oz-Sabater, Joaqu{\'\i}n and Nicolas, Julien and Radu, Raluca and others},
  journal={Quarterly Journal of the Royal Meteorological Society},
  volume={150},
  number={764},
  pages={4014--4048},
  year={2024},
  publisher={Wiley Online Library}
}

@article{fettweis2013estimating,
  title={Estimating the Greenland ice sheet surface mass balance contribution to future sea level rise using the regional atmospheric climate model MAR},
  author={Fettweis, Xavier and Franco, Bruno and Tedesco, Marco and Van Angelen, JH and Lenaerts, Jan TM and van den Broeke, Michiel R and Gall{\'e}e, H},
  journal={The Cryosphere},
  volume={7},
  number={2},
  pages={469--489},
  year={2013},
  publisher={Copernicus GmbH}
}

@article{vernon2013surface,
  title={Surface mass balance model intercomparison for the Greenland ice sheet},
  author={Vernon, Christopher L and Bamber, JL and Box, JE and Van den Broeke, MR and Fettweis, Xavier and Hanna, Edward and Huybrechts, Phillipe},
  journal={The Cryosphere},
  volume={7},
  number={2},
  pages={599--614},
  year={2013},
  publisher={Copernicus Publications G{\"o}ttingen, Germany}
}

@article{reeh1991parameterization,
  title={Parameterization of melt rate and surface temperature in the Greenland ice sheet},
  author={Reeh, Niels},
  journal={Polarforschung},
  volume={59},
  number={3},
  pages={113--128},
  year={1991},
  publisher={Alfred Wegener Institute for Polar and Marine Research \& German Society of~…}
}

@INPROCEEDINGS{teisberg,
  author={Teisberg, Thomas O. and Schroeder, Dustin M. and MacKie, Emma J.},
  booktitle={2021 IEEE International Geoscience and Remote Sensing Symposium IGARSS}, 
  title={A Machine Learning Approach to Mass-Conserving Ice Thickness Interpolation}, 
  year={2021},
  volume={},
  number={},
  pages={8664-8667},
  doi={10.1109/IGARSS47720.2021.9555002}}

@ARTICLE{TNNLS_GNN_Survey,
  author={Wu, Zonghan and Pan, Shirui and Chen, Fengwen and Long, Guodong and Zhang, Chengqi and Yu, Philip S.},
  journal={IEEE Trans. Neural Netw. Learn. Syst.}, 
  title={A Comprehensive Survey on Graph Neural Networks}, 
  year={2021},
  volume={32},
  number={1},
  pages={4-24},
  doi={10.1109/TNNLS.2020.2978386}}

@misc{ibikunle2025_Dataset,
      title={AI-ready Snow Radar Echogram Dataset (SRED) for climate change monitoring}, 
      author={Oluwanisola Ibikunle and Hara Talasila and Debvrat Varshney and Jilu Li and John Paden and Maryam Rahnemoonfar},
      year={2025},
      eprint={2505.00786},
      archivePrefix={arXiv},
      primaryClass={cs.CV},
      url={https://arxiv.org/abs/2505.00786}, 
}

@inproceedings{liu2025locate,
  title={Locate and extend: a geometric deep learning strategy for predicting polar ice layer structures using graph neural networks},
  author={Liu, Zesheng and Rahnemoonfar, Maryam},
  booktitle={Pattern Recognition and Prediction XXXVI},
  volume={13464},
  pages={1346402},
  year={2025},
  organization={SPIE}
}

@inproceedings{cini2023scalable,
  title={Scalable spatiotemporal graph neural networks},
  author={Cini, Andrea and Marisca, Ivan and Bianchi, Filippo Maria and Alippi, Cesare},
  booktitle={Proceedings of the AAAI conference on artificial intelligence},
  volume={37},
  number={6},
  pages={7218--7226},
  year={2023}
}

@inproceedings{tang2023explainable,
  title={Explainable spatio-temporal graph neural networks},
  author={Tang, Jiabin and Xia, Lianghao and Huang, Chao},
  booktitle={Proceedings of the 32nd ACM International Conference on Information and Knowledge Management},
  pages={2432--2441},
  year={2023}
}

@misc{wang2024stgformer,
      title={STGformer: Efficient Spatiotemporal Graph Transformer for Traffic Forecasting}, 
      author={Hongjun Wang and Jiyuan Chen and Tong Pan and Zheng Dong and Lingyu Zhang and Renhe Jiang and Xuan Song},
      year={2024},
      eprint={2410.00385},
      archivePrefix={arXiv},
      primaryClass={cs.LG},
      url={https://arxiv.org/abs/2410.00385}, 
}

@article{VERDONE_APP1,
title = {Explainable Spatio-Temporal Graph Neural Networks for multi-site photovoltaic energy production},
journal = {Applied Energy},
volume = {353},
pages = {122151},
year = {2024},
issn = {0306-2619},
doi = {https://doi.org/10.1016/j.apenergy.2023.122151},
url = {https://www.sciencedirect.com/science/article/pii/S0306261923015155},
author = {Alessio Verdone and Simone Scardapane and Massimo Panella}
}

@article{LI_APP2,
title = {MAST-GNN: A multimodal adaptive spatio-temporal graph neural network for airspace complexity prediction},
journal = {Transportation Research Part C: Emerging Technologies},
volume = {160},
pages = {104521},
year = {2024},
issn = {0968-090X},
doi = {https://doi.org/10.1016/j.trc.2024.104521},
url = {https://www.sciencedirect.com/science/article/pii/S0968090X24000421},
author = {Biyue Li and Zhishuai Li and Jun Chen and Yongjie Yan and Yisheng Lv and Wenbo Du}
}

@INPROCEEDINGS{GRIT,
  author={Liu, Zesheng and Rahnemoonfar, Maryam},
  booktitle={IGARSS 2025 - 2025 IEEE International Geoscience and Remote Sensing Symposium}, 
  title={GRIT: Graph Transformer For Internal Ice Layer Thickness Prediction}, 
  year={2025},
  volume={},
  number={},
  pages={1-5},
  doi={10.1109/IGARSS55030.2025.11243115}}

@INPROCEEDINGS{ST-GRIT,
  author={Liu, Zesheng and Rahnemoonfar, Maryam},
  booktitle={2025 IEEE International Conference on Image Processing (ICIP)}, 
  title={ST-GRIT: Spatio-Temporal Graph Transformer For Internal Ice Layer Thickness Prediction}, 
  year={2025},
  volume={},
  number={},
  pages={1109-1114},
  doi={10.1109/ICIP55913.2025.11084445}}

@article{RINENG_Rainfall,
title = {Novel deep learning framework for rainfall forecasting integrating generative adversarial and spatiotemporal graph neural networks},
journal = {Results in Engineering},
volume = {29},
pages = {108616},
year = {2026},
issn = {2590-1230},
doi = {https://doi.org/10.1016/j.rineng.2025.108616},
url = {https://www.sciencedirect.com/science/article/pii/S2590123025046602},
author = {Usa Wannasingha Humphries and Muhammad Waqas and Shakeel Ahmad}
}

@article{Neuro_Molecular,
title = {Feature-enhanced graph neural network with multiple attention for molecular property prediction},
journal = {Neurocomputing},
volume = {669},
pages = {132426},
year = {2026},
issn = {0925-2312},
doi = {https://doi.org/10.1016/j.neucom.2025.132426},
url = {https://www.sciencedirect.com/science/article/pii/S092523122503098X},
author = {Bo Qin and Xu Zhu and Chen-Yang Fan and Xin Xue and Meng-Meng Wang and Hao-Yang Tang}
}

@article{Neuro_Motion,
title = {STGNet: A spatio-temporal graph neural network for motion prediction in autonomous driving},
journal = {Neurocomputing},
volume = {676},
pages = {133063},
year = {2026},
issn = {0925-2312},
doi = {https://doi.org/10.1016/j.neucom.2026.133063},
url = {https://www.sciencedirect.com/science/article/pii/S0925231226004601},
author = {Xinyu Wang and Li Liu}
}

@article{Neuro_Traffic_Data,
title = {Bidirectional spatial–temporal traffic data imputation via graph attention recurrent neural network},
journal = {Neurocomputing},
volume = {531},
pages = {151-162},
year = {2023},
issn = {0925-2312},
doi = {https://doi.org/10.1016/j.neucom.2023.02.017},
url = {https://www.sciencedirect.com/science/article/pii/S0925231223001558},
author = {Guojiang Shen and Wenfeng Zhou and Wenyi Zhang and Nali Liu and Zhi Liu and Xiangjie Kong}
}

@article{Neuro_Drug,
title = {Hierarchy-aware graph neural network and inverse-variance reinforcement learning for drug recommendation},
journal = {Neurocomputing},
volume = {676},
pages = {132989},
year = {2026},
issn = {0925-2312},
doi = {https://doi.org/10.1016/j.neucom.2026.132989},
url = {https://www.sciencedirect.com/science/article/pii/S0925231226003863},
author = {Wei Hou and Xianxing Liu and Linxiao Li and Chunling Fu}
}

@article{Neuro_Disease,
title = {A review of graph neural networks for brain diseases analysis},
journal = {Neurocomputing},
pages = {133174},
year = {2026},
issn = {0925-2312},
doi = {https://doi.org/10.1016/j.neucom.2026.133174},
url = {https://www.sciencedirect.com/science/article/pii/S0925231226005710},
author = {Hong Yang and Ruiwen Huang and Shanshan Ye and Peng Zhang and Yuhuai Guo and Shirui Pan and Yanchun Zhang}
}

@book{conover1999practical,
  title={Practical nonparametric statistics},
  author={Conover, William Jay},
  year={1999},
  publisher={john wiley \& sons}
}

@Article{SSGCRTN,
author={Yang, Shiyu
and Wu, Qunyong
and Wang, Yuhang
and Lin, Tingyu},
title={SSGCRTN: a space-specific graph convolutional recurrent transformer network for traffic prediction},
journal={Applied Intelligence},
year={2024},
month={Nov},
day={01},
volume={54},
number={22},
pages={11978-11994},
issn={1573-7497},
doi={10.1007/s10489-024-05815-1},
url={https://doi.org/10.1007/s10489-024-05815-1}
}

@article{MSTDFGRN,
title = {MSTDFGRN: A Multi-view Spatio-Temporal Dynamic Fusion Graph Recurrent Network for traffic flow prediction},
journal = {Computers and Electrical Engineering},
volume = {123},
pages = {110046},
year = {2025},
issn = {0045-7906},
doi = {https://doi.org/10.1016/j.compeleceng.2024.110046},
url = {https://www.sciencedirect.com/science/article/pii/S0045790624009716},
author = {Shiyu Yang and Qunyong Wu and Yuhang Wang and Zhan Zhou}
}

@Article{PSTCGCN,
author={Yang, Shiyu
and Wu, Qunyong
and Li, Ziwei
and Wang, Keyue},
title={PSTCGCN: Principal spatio-temporal causal graph convolutional network for traffic flow prediction},
journal={Neural Computing and Applications},
year={2025},
month={Jul},
day={01},
volume={37},
number={20},
pages={14751-14764},
issn={1433-3058},
doi={10.1007/s00521-024-10591-7},
url={https://doi.org/10.1007/s00521-024-10591-7}
}

@ARTICLE{TIIDGCN,
  author={Yang, Shiyu and Wu, Qunyong and Li, Mengmeng and Sun, Yu},
  journal={IEEE Internet of Things Journal}, 
  title={Temporal Identity Interaction Dynamic Graph Convolutional Network for Traffic Forecasting}, 
  year={2025},
  volume={12},
  number={11},
  pages={15057-15072},
  doi={10.1109/JIOT.2025.3529761}}

@article{SDSINet,
title = {SDSINet: A spatiotemporal dual-scale interaction network for traffic prediction},
journal = {Applied Soft Computing},
volume = {173},
pages = {112892},
year = {2025},
issn = {1568-4946},
doi = {https://doi.org/10.1016/j.asoc.2025.112892},
url = {https://www.sciencedirect.com/science/article/pii/S1568494625002030},
author = {Shiyu Yang and Qunyong Wu}
}

@ARTICLE{GDGCRN,
  author={Yang, Shiyu and Huang, Zhanchao and Wu, Qunyong and Zhuo, Zihao},
  journal={IEEE Sensors Journal}, 
  title={General Decoupled Graph Convolutional Recurrent Network for Traffic Prediction}, 
  year={2025},
  volume={25},
  number={18},
  pages={35460-35478},
  doi={10.1109/JSEN.2025.3580440}}

@article{MTEGCRN,
title = {MTEGCRN: Multi-scale temporal enhanced graph convolutional recurrent network for traffic prediction},
journal = {Neurocomputing},
volume = {653},
pages = {131064},
year = {2025},
issn = {0925-2312},
doi = {https://doi.org/10.1016/j.neucom.2025.131064},
url = {https://www.sciencedirect.com/science/article/pii/S0925231225017369},
author = {Shiyu Yang and Qunyong Wu}
}

@article{DMFGCRN,
title = {Decoupled multi-spatio-temporal fusion graph convolutional recurrent network for traffic prediction},
journal = {Engineering Applications of Artificial Intelligence},
volume = {163},
pages = {112956},
year = {2026},
issn = {0952-1976},
doi = {https://doi.org/10.1016/j.engappai.2025.112956},
url = {https://www.sciencedirect.com/science/article/pii/S0952197625029872},
author = {Shiyu Yang and Qunyong Wu and Mengmeng Li}
}

@article{MReDTrajRec,
author = {Zhan Zhou and Qunyong Wu and Shiyu Yang and Jiahuan Luo},
title = {MReDTrajRec: a multi-representation data-driven model for trajectory recovery under road network constraints},
journal = {Transportmetrica B: Transport Dynamics},
volume = {13},
number = {1},
pages = {2596312},
year = {2025},
publisher = {Taylor \& Francis},
doi = {10.1080/21680566.2025.2596312},


URL = { 
    
        https://doi.org/10.1080/21680566.2025.2596312
    
    

},
eprint = { 
    
        https://doi.org/10.1080/21680566.2025.2596312
    
    

}

}

@ARTICLE{SGASeq,
  author={Yang, Shiyu and Wu, Qunyong and Huang, Zhanchao and Zhuo, Zihao},
  journal={IEEE Transactions on Network and Service Management}, 
  title={SGA-Seq: Station-Aware Graph Attention Sequence Network for Cellular Traffic Prediction}, 
  year={2026},
  volume={23},
  number={},
  pages={2652-2665},
  doi={10.1109/TNSM.2026.3664401}}

@book{molnar2025_Shapley,
  title={Interpretable Machine Learning},
  subtitle={A Guide for Making Black Box Models Explainable},
  author={Christoph Molnar},
  year={2025},
  edition={3},
  isbn={978-3-911578-03-5},
  url={https://christophm.github.io/interpretable-ml-book}
}

@inbook{Shapley,
url = {https://doi.org/10.1515/9781400881970-018},
title = {17. A Value for n-Person Games},
booktitle = {Contributions to the Theory of Games, Volume II},
author = {L. S. Shapley},
editor = {Harold William Kuhn and Albert William Tucker},
publisher = {Princeton University Press},
address = {Princeton},
pages = {307--318},
doi = {doi:10.1515/9781400881970-018},
isbn = {9781400881970},
year = {2016},
lastchecked = {2026-04-20}
}

@Article{Grabisch1999,
author={Grabisch, Michel
and Roubens, Marc},
title={An axiomatic approach to the concept of interaction among players in cooperative games},
journal={International Journal of Game Theory},
year={1999},
month={Nov},
day={01},
volume={28},
number={4},
pages={547-565},
issn={1432-1270},
doi={10.1007/s001820050125},
url={https://doi.org/10.1007/s001820050125}
}

@inproceedings{GAT,
title={Graph Attention Networks},
author={Petar Veličković and Guillem Cucurull and Arantxa Casanova and Adriana Romero and Pietro Liò and Yoshua Bengio},
booktitle={International Conference on Learning Representations},
year={2018},
url={https://openreview.net/forum?id=rJXMpikCZ},
}

@inproceedings{GIN,
title={How Powerful are Graph Neural Networks?},
author={Keyulu Xu and Weihua Hu and Jure Leskovec and Stefanie Jegelka},
booktitle={International Conference on Learning Representations},
year={2019},
url={https://openreview.net/forum?id=ryGs6iA5Km},
}

@INPROCEEDINGS{BIRFNO,
  author={Wang, Heling and Rahnemoonfar, Maryam},
  booktitle={2025 IEEE International Radar Conference (RADAR)}, 
  title={Predicting Firn Layer Thickness With Birfno: A Novel Fourier Neural Operator for Snow Radar Imaging}, 
  year={2025},
  volume={},
  number={},
  pages={1-6},
  doi={10.1109/RADAR52380.2025.11031618}}

@INPROCEEDINGS{GNO,
  author={Wang, Heling and Yari, Masoud and Rahnemoonfar, Maryam},
  booktitle={IGARSS 2025 - 2025 IEEE International Geoscience and Remote Sensing Symposium}, 
  title={Graph Neural Operators for Accurate Ice Layer Thickness Prediction From Radargram}, 
  year={2025},
  volume={},
  number={},
  pages={9253-9257},
  doi={10.1109/IGARSS55030.2025.11313880}}

@article{Spearman_Corr,
 ISSN = {00029556},
 URL = {http://www.jstor.org/stable/1412159},
 author = {C. Spearman},
 journal = {The American Journal of Psychology},
 number = {1},
 pages = {72--101},
 publisher = {University of Illinois Press},
 title = {The Proof and Measurement of Association between Two Things},
 urldate = {2026-04-21},
 volume = {15},
 year = {1904}
}

@misc{liu2026physicsconditionedsynthesisinternalicelayer,
      title={Physics-Conditioned Synthesis of Internal Ice-Layer Thickness for Incomplete Layer Traces}, 
      author={Zesheng Liu and Maryam Rahnemoonfar},
      year={2026},
      eprint={2604.20783},
      archivePrefix={arXiv},
      primaryClass={cs.LG},
      url={https://arxiv.org/abs/2604.20783}, 
}



\end{document}